\documentclass[default,sn-mathphys,Numbered,iicol]{sn-jnl}%

\usepackage{graphicx}%
\usepackage{multirow}%
\usepackage{amsmath,amssymb,amsfonts}%
\usepackage{amsthm}%
\usepackage{mathrsfs}%
\usepackage[title]{appendix}%
\usepackage{xcolor}%
\usepackage{textcomp}%
\usepackage{manyfoot}%
\usepackage{booktabs}%
\usepackage{algorithm}%
\usepackage{algorithmicx}%
\usepackage{algpseudocode}%
\usepackage{listings}%
\usepackage{epsfig}
\usepackage{url}            %
\usepackage{nicefrac}       %
\usepackage{microtype}      %
\usepackage{colortbl}
\usepackage{dsfont}
\usepackage{tabularx}
\usepackage{wrapfig}
\usepackage{diagbox}
\usepackage{pifont}%
\newcounter{abs}

\usepackage{mathtools}
\usepackage{caption}
\usepackage{xspace}

\usepackage{hyperref}
\usepackage{cleveref}

\makeatletter
\DeclareRobustCommand\onedot{\futurelet\@let@token\@onedot}
\def\@onedot{\ifx\@let@token.\else.\null\fi\xspace}

\def\eg{\emph{e.g}\onedot} 
\def\ie{\emph{i.e}\onedot} 
 
 \def\vs{\emph{vs}\onedot}
 
\def\iid{i.i.d\onedot}

\newcommand{\whj}[1]{\textcolor{black}{#1}}

\newcommand{\reb}[1]{\textcolor{black}{#1}}
\newcommand{\whjpreb}[1]{\textcolor{black}{#1}}

\makeatother

\begin{document}

\title{Dissecting Out-of-Distribution Detection and Open-Set Recognition: A Critical Analysis of Methods and Benchmarks}

\author[1]{ \sur{Hongjun Wang}}\email{hjwang@connect.hku.hk}

\author[2]{\sur{Sagar Vaze}}\email{sagar@robots.ox.ac.uk}

\author*[1]{\sur{Kai Han}}\email{kaihanx@hku.hk}

\affil[1]{\orgname{The University of Hong Kong}}
\affil[2]{\orgname{University of Oxford}}

\abstract{Detecting test-time distribution shift has emerged as a key capability for safely deployed machine learning models, with the question being tackled under various guises in recent years.
In this paper, we aim to provide a consolidated view of the two largest sub-fields within the community: out-of-distribution (OOD) detection and open-set recognition (OSR). 
In particular, we aim to provide rigorous empirical analysis of different methods across settings and provide actionable takeaways for practitioners and researchers.
Concretely, we make the following contributions:
(i) We perform rigorous cross-evaluation between state-of-the-art methods in the OOD detection and OSR settings and identify a strong correlation between the performances of methods for them;
(ii) We propose a new, large-scale benchmark setting which we suggest better disentangles the problem tackled by OOD detection and OSR, re-evaluating state-of-the-art OOD detection and OSR methods in this setting; 
(iii) We surprisingly find that the best performing method on standard benchmarks (Outlier Exposure) struggles when tested at scale, while scoring rules which are sensitive to the deep feature magnitude consistently show promise;
and (iv) We conduct empirical analysis to explain these phenomena and highlight directions for future research. 
Code: \url{https://github.com/Visual-AI/Dissect-OOD-OSR}
}


\maketitle

\section{Introduction}\label{sec:introduction}
Any practical machine learning model is likely to encounter test-time samples which differ substantially from its training set; \ie,  models are likely to encounter test-time \textit{distribution shift}. 
As such, \textit{detecting} distribution shift has emerged as a key research problem in the community~\cite{Scheirer_2013_TPAMI, hendrycks2016baseline, liu2020energy}.
Specifically, \textit{out-of-distribution} (OOD) detection~\cite{hendrycks2019oe,sun2021react} and \textit{open-set recognition} (OSR)~\cite{Chen_2020_ECCV,chen2021adversarial} have emerged as two rich sub-fields to tackle this task.
In fact, both tasks explicitly tackle the setting in which multi-way classifiers must detect if test samples are `unseen' with respect to their training set, with a variety of methods and benchmarks proposed within each field. 
OOD detection methods test on images which come from different \textit{datasets} to the training set, while OSR methods are evaluated on the ability to detect test images which come from different \textit{semantic categories} to the training set.
Research efforts in both of these fields largely occur independently, with little cross-pollination of ideas.
Though many prior works have recognized the similarity of the two sub-fields~\cite{vaze2022openset,tran2022plex,yang2024generalized, salehi2021unified}, there has been little benchmarking to understand the underlying similarities and differences between them.

In this study, \reb{we investigate the detection of distribution shifts, with a focus on exploring and analyzing OOD detection and OSR methods and benchmarks. Our aim is to gain a comprehensive understanding of the underlying similarities and differences between these two tasks.} We perform rigorous cross-evaluation between methods developed for OOD detection and OSR on current standard benchmarks, finding that methods which perform well for one are likely to perform well for the other (\Cref{sec:analysis_small}). 
We experiment both with methods which require specialized training strategies (\eg, Outlier Exposure~\cite{hendrycks2019oe} (OE) and ARPL~\cite{chen2021adversarial}) as well as different post-hoc scoring rules (\eg, MSP~\cite{hendrycks2016baseline}, MLS~\cite{vaze2022openset} and Energy~\cite{liu2020energy}).
We thoroughly evaluate all methods on both standard OOD detection and OSR benchmarks, after which we find that OE achieves almost saturating performance on the OOD detection task and also obtains state-of-the-art results on the OSR task. 
We further find that the scoring rules which are sensitive to the magnitude of the deep image embeddings (like MLS~\cite{vaze2022openset} and Energy Scoring~\cite{liu2020energy}) show \reb{the best} performance across tasks and datasets.

\begin{figure}[t]
    \centering
    \includegraphics[width=\linewidth]{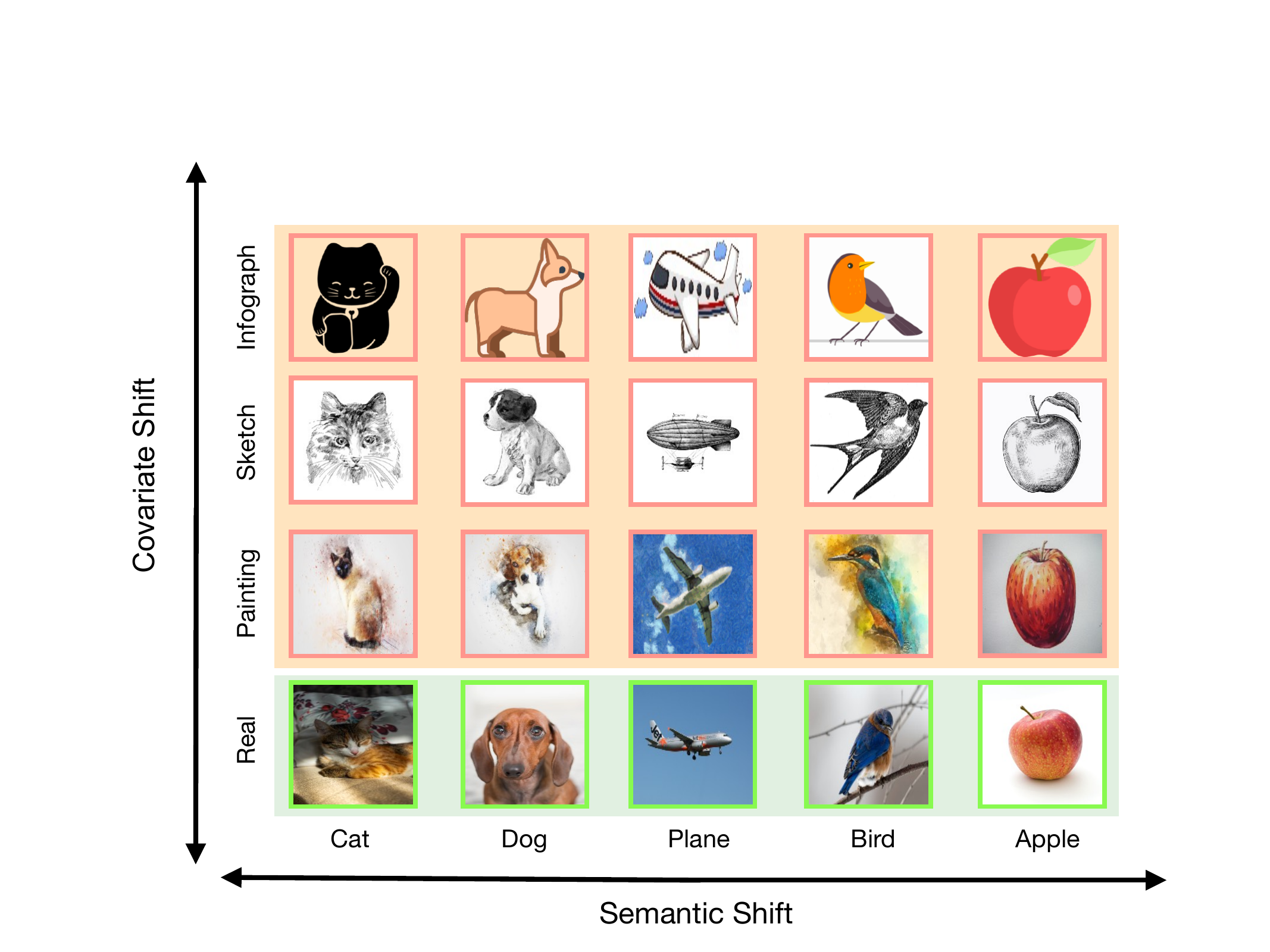}
    \caption{Semantic shift \vs covariate shift. We systematically perform cross-evaluation between SOTA methods for OSR and OOD detection and propose a large-scale benchmark setting in which we disentangle the tasks tackled in the two fields, proposing that they tackle \textit{semantic shift} (x-axis) and \textit{covariate shift} (y-axis) respectively.}
    \label{fig:teaser}
\end{figure}

Next, we propose a reconciling perspective on the tasks tackled by the two fields, and propose a new benchmark to assess this (\Cref{sec:analysis_new}). 
Specifically, we propose a new, large-scale benchmark setting, in which we disentangle different distribution shifts, namely \textit{semantic} shift and \textit{covariate} shift, that occur in OOD detection and OSR (see~\Cref{fig:teaser}). 
Though these concepts have been discussed before~\cite{hendrycks2021natural,tian2021exploring}, \whjpreb{the standard large-scale benchmarks in OOD detection have not adequately separated them. For example, semantic shift and covariate shift simultaneously occur when detecting OOD samples from Places using pre-trained ImageNet models.}
We propose a conceptual framework to understand them and further propose large-scale evaluation settings, including for pre-trained ImageNet models.
For example, to isolate \textit{semantic shift} on ImageNet, we leverage the recently introduced Semantic Shift Benchmark (SSB)~\cite{vaze2022openset}, in which the original ImageNet-1K~\cite{russakovsky2015imagenet} is regarded as `seen' closed-set data while `unseen' data is carefully drawn from the disjoint set of ImageNet-21K-P~\cite{ridnik2021imagenet}. For \textit{covariate shift}, we leverage ImageNet-C \cite{hendrycks2019robustness} and ImageNet-R \cite{Hendrycks2020TheMF} to demonstrate distribution shift with respect to the standard ImageNet dataset.
\reb{Furthermore}, to account for the tension between \reb{being} \textit{robust} to covariate shift (also known as OOD-generalisation~\cite{sagawa2019distributionally,ye2022ood}) and being able to \textit{detect the presence of it}, we further introduce a new metric `Outlier-Aware Accuracy' (OAA).

Finally, we examine SoTA OOD detection and OSR methods on our large-scale benchmark to validate whether the findings on the standard (small-scale) datasets still hold on our consolidated large-scale evaluation. 
Through large-scale analysis, we surprisingly find that OE struggles to scale to larger benchmarks, while the magnitude-aware scoring rules, especially MLS~\cite{vaze2022openset}, still show promise. 
We further provide empirical insights by analysing the representations extracted by different models under different distribution shifts. 
Our analysis suggest that the strong performance of OE on existing benchmarks is largely attributed to the fact that the auxiliary OOD data used for training has a distribution overlap with the OOD testing data (as measured by distances in feature space).
Meanwhile, we find that it is not straightforward to find auxiliary OOD data which reflects the range of possible distribution shifts with respect to large-scale datasets. 
We believe there are still many open questions to be answered in the shared space of OOD detection and OSR, and hope the findings in our work can serve as a platform for future investigation. 

\section{Related work}
\label{sec:related}

\textbf{Open-set recognition.} Previous work~\cite{scheirer2012toward} coins `open-set recognition', the objective of which is to identify unknown classes while classifying the known ones. OpenMax resorts to Activation Vector (AV) and models the distribution of AVs based on the Extreme Value Theorem (EVT). Recent works~\cite{ge2017generative,neal2018open,kong2021opengan} show that the generated data from synthetic distribution would be helpful to improve OSR. OSRCI~\cite{neal2018open} generates images belonging to the unknown classes but similar to the training data to train an open-set classifier. \cite{kong2021opengan} adversarially trains discriminator to distinguish closed from open-set images and introduces real open-set samples for model selection. Prototype-based methods~\cite{Chen_2020_ECCV,chen2021adversarial} (\ie, ARPL / ARPL+CS) adjust the boundaries of different classes and identify open-set images based on distances to the learned prototypes of known classes. \reb{MLS~\cite{vaze2022openset} uses maximum logit scores rather than softmax scores to maintain the magnitude information.}

\textbf{Out of Distribution Detection.} The goal of OOD detection is generally specified as identifying test-time samples coming from a `different distribution' from the training data. \cite{hendrycks2016baseline} formalizes the task of out-of-distribution detection and provides a paradigm to evaluate deep learning out-of-distribution detectors using the maximum softmax probability (MSP). A test sample with a large MSP score is detected as an in-distribution (ID) example rather than out-of-distribution (OOD) example. ODIN~\cite{Liang2018ODIN} and its learnable variant G-ODIN \cite{hsu2020generalized} add adversarial perturbations to both ID and OOD samples and employ temperature scaling strategy on the softmax output to separate them. \cite{liu2020energy} proposes the energy score derived from the logit outputs for OOD uncertainty estimation. \cite{sun2021react} rectifies the distribution of per-unit activations in the penultimate layer for ID and OOD data. \reb{GradNorm~\cite{huang2021importance} calculates gradients by backpropagating the KL divergence between the softmax output and a uniform distribution, assuming that the magnitude of gradients is higher for ID data than that for OOD data.
ASH~\cite{djurisic2023extremely} removes a large portion of the activations based on the \textit{p}th-percentile of the entire representation at a late layer. The remaining activations are utilized to calculate an energy score for OOD detection.
SHE~\cite{zhang2023out} quantifies the dissimilarity between the ID training samples from each category and the testing samples based on the features extracted from the penultimate layer of the model. This dissimilarity is then used as the score to judge whether a testing sample is OOD or not. 
} 
Outlier Exposure (OE)~\cite{hendrycks2019oe} and GradNorm~\cite{huang2021importance} both design a loss based on the KL divergence between the softmax output and a uniform probability distribution to encourage models to output a uniform softmax distribution on outliers. The former leverages real OOD data for training while the latter directly employs the vector norm of gradients to perform uncertainty estimation. 

\textbf{Relations between OOD detection and OSR.} Prior works discuss the separation between covariate and semantic distributional shift \cite{tian2021exploring,ahmed2020systematic,deecke2021transfer,yang2021semantically}. \cite{tian2021exploring} discusses separately detecting covariate and concept distributional shift on small-scale datasets (\ie, CIFAR-10/100).
However, similarly to~\cite{vaze2022openset}, we suggest that small-scale datasets with no explicit taxonomies (like CIFAR) are not well suited for defining semantic shift. As such, we aim to build a large-scale benchmark with a clear underlying taxonomy. 
\cite{hendrycks2021natural} introduces ImageNet-A (\ie, collections of natural adversarial examples) and ImageNet-O (\ie, samples of held-out classes from ImageNet-21K) for robustness evaluation and unseen classes recognition, while \cite{ahmed2020systematic} curates a set of artificial datasets to disentangle the evaluation of non-semantic distributional shift and semantic-shift. However, they focus more on achieving robustness to non-semantic distributional shift and do not develop cross-evaluation between state-of-the-art methods in the OOD detection and OSR settings. \cite{deecke2021transfer} treats both semantic and non-semantic tasks in an anomaly detection (AD) paradigm and applies popular AD methods to them on CIFAR10. Our work explicitly explores the relation between OSR and OOD detection tasks, and verify the effectiveness of respective popular methods in each field.
Two surveys \cite{yang2024generalized,salehi2021unified} summarize a number of approaches within the OOD detection and OSR settings, along with anomaly detection and Novelty Detection. \cite{kim2021unified} constructs a unified benchmark to verify existing OOD detection methods, delineating `far-OOD' and `near-OOD'. Meanwhile, \cite{xia2022augmenting} rethinks the importance of ID misclassifications in the OOD context and examines different approaches on selective classification in the presence of OOD datasets. 
In our work, we not only discuss the link between robustness and OOD detection, but also propose a new metric to reconcile the tasks. 
Concurrent work~\cite{yang2022openood} provides a codebase for representative methods within OSR and OOD detection. \whjpreb{In this paper, we categorize shift detection methods into two types: scoring rules (\eg, MSP, MLS, etc), which operate post-hoc on pre-trained networks, and specialized training, which modifies the networks' optimization procedures (\eg, ARPL/ARPL+CS, OE, etc).
}

\begin{table*}[t]
\caption{
\reb{Summary of representative OOD detection and OSR techniques. The works are categorized based on the task that they are developed for (\ie, OOD detection and OSR) and the methodology employed (\ie, scoring rules and training strategies).}
}\label{tab:stat}
\centering
\resizebox{0.8\linewidth}{!}{
\begin{tabular}{lccc}
\toprule
 & \begin{tabular}[c]{@{}c@{}}Developed for \\ OOD or OSR\end{tabular} & Scoring rule & Training strategy \\
\midrule
MSP~\cite{hendrycks2016baseline}    &   OOD  &        MSP           &     CE                    \\
MLS~\cite{vaze2022openset}      &  OSR   &    MLS               &     CE           \\
ODIN~\cite{liang2017enhancing}     &   OOD    &       MSP            &     CE                  \\
GODIN~\cite{hsu2020generalized}    &    OOD    &       MSP            &     CE                   \\
GradNorm~\cite{huang2021importance}      &    OOD  &          MSP         &     CE                 \\
SEM~\cite{yang2022fsOOD}      &    OOD   &           MSP        &     CE                \\
Energy~\cite{liu2020energy}   &    OOD   &         Energy          &     CE               \\
ReAct~\cite{sun2021react}    &    OOD   &     MSP/Energy/ODIN              &     CE                   \\
ASH~\cite{djurisic2023extremely}      &   OOD    &         MSP/Energy/ODIN/ReAct          &     CE                 \\
SHE~\cite{zhang2023out}      &   OOD    &         MSP/Energy/ODIN          &     CE                   \\
ARPL+CS~\cite{chen2021adversarial}  &  OSR   &   MSP           &          ARPL+CS           \\
OE~\cite{hendrycks2019oe}       &   OOD    &        MSP           &    OE        \\\bottomrule  
\end{tabular}
}
\end{table*}

\reb{\textbf{Auxiliary data in OOD detection.} Inspired by OE~\cite{hendrycks2019oe}, recent work~\cite{ming2022poem,chen2021atom,wang2023out,wang2024learning} leverage auxiliary data in some form to enhance the model's ability to detect OOD data. This could be through posterior sampling~\cite{ming2022poem}, adversarial training~\cite{chen2021atom}, augmenting distributions~\cite{wang2023out} or model perturbation~\cite{wang2024learning}. POEM~\cite{ming2022poem} focuses on posterior sampling to learn a decision boundary between ID and OOD data. ATOM~\cite{chen2021atom} introduces an adversarial training method with informative outlier mining, which is specifically designed to improve the robustness against adversarial attacks, in which the adversarial data is considered as a special type of OOD data.
DAL~\cite{wang2023out} addresses the distribution discrepancy between auxiliary and unseen real OOD data, by training predictors over the worst OOD data in a Wasserstein ball. 
DOE~\cite{wang2024learning} leverages implicit data transformation through the embedding features' perturbation to minimize a distribution discrepancy measurement called worst OOD regret, aiming to enhance the model's robustness to distribution shifts. 
Our work offers unique insights into the selection of auxiliary data to optimize OOD detection performance. 
By uncovering the relationship between the auxiliary data and the model's OOD detection performance, 
our work has the potential to inform strategies for auxiliary data selection and manipulation toward more reliable OOD detection solutions.}

\textbf{Key similarities and distinctions with prior work.}
While several papers~\cite{deecke2021transfer,yang2022openood} have jointly considered methods in the OOD detection and OSR tasks, few works have clearly distinguished the academic and practical differences (or similarities) between them.
In this work, we not only provide \reb{empirical analysis} but also propose a \textit{conceptual framework} and \textit{large-scale benchmark} to better reconcile these problems.

\section{\reb{Cross-benchmarking of OOD detection and OSR methods}}
\label{sec:analysis_small}
\reb{Despite the growing popularity of OOD detection and OSR studies, 
these two tasks have largely evolved independently and in isolation from each other, as shown in~\Cref{tab:stat}.
Indeed, methods designed for OSR can be seamlessly adopted to address the OOD detection problem, and vice versa. 
Recent generalized OOD detection frameworks~\cite{deecke2021transfer,yang2022openood} unify 
tasks relevant to OOD detection and OSR. 
However, there is still a lack of cross-evaluation between methods developed for OOD detection and OSR on current standard benchmarks.
Given the strong inherent connections between OOD detection and OSR, a comprehensive cross-benchmarking comparison is crucial to shed light on the future development of the broader distribution shift detection problem.
}

As a starting point to reconcile OOD detection and OSR, in this section we perform cross-evaluation of methods from both sub-fields. 

\subsection{Experimental setup}

\textbf{Problem setting.}
\whj{
Let $X\in\mathcal{R}^D$ denote an input sample and $C\in\mathcal{R}$ denote the label of interest. 
Test-time \textit{distribution shift} occurs when the testing joint distribution is not equal to the training joint distribution, \ie, $P_{test}(X,C)\neq P_{train}(X,C)$. This shift can be further divided into two types: \textit{covariate shift} and \textit{semantic shift}. \textit{Covariate shift} occurs when $P_{test}(C|X) = P_{train}(C|X)$ but $P_{test}(X)\neq P_{train}(X)$. \textit{Semantic shift} occurs when $P_{test}(C|X) \neq P_{train}(C|X)$ but $P_{test}(X) = P_{train}(X)$. In OOD detection and OSR for multi-class classification, the label space contains multiple semantic categories $\left\{c_1, \cdots, c_L\right\}$, where $L$ is the total number of categories in the testing data. 
The model needs to identify the distribution from which test-time samples originate and conduct classification based on the posterior probability, represented as $p(C=c_i\mid X)$.
}

\textbf{Methods.}
We distinguish two categories of shift detection methods: \textit{scoring rules} (which operate post-hoc on top of pre-trained networks); and \textit{specialized training} (which change the optimization procedure of the networks).

For \textit{scoring rules}, we compare the maximum softmax probability (MSP)~\cite{hendrycks2016baseline}, the Maximum Logit Score (MLS)~\cite{vaze2022openset}, ODIN~\cite{Liang2018ODIN}, GODIN~\cite{hsu2020generalized}, Energy scoring~\cite{liu2020energy}, GradNorm~\cite{huang2021importance} and SEM~\cite{yang2022fsOOD}.
We further experiment with ReAct~\cite{sun2021react}, an activation pruning technique which can be employed in conjunction with any scoring rule.
While MLS was developed for OSR~\cite{vaze2022openset}, the other scoring rules were developed for OOD detection.
For now, we note that MLS, Energy and GradNorm are all sensitive to the \textit{magnitude of the feature norm} of the network, while the others are not. We refer to the former scoring rules as `magnitude aware'.

\begin{table*}[htb]
\caption{Evaluation on small-scale OOD detection and OSR benchmarks with various methods, using CIFAR10 as ID. The results are averaged from five independent runs. We report the in-distribution accuracy as `ID' and denote intractable results as `-', resulting from unaffordable computational cost. Bold values represent
the best results, while underlined values represent the second best results. Different methods have their optimal scope but MLS and Energy demonstrate their stability and models trained with OE dominate on almost all OOD datasets. }\label{tab:small_bench}
\centering
\resizebox{\linewidth}{!}{
\begin{tabular}{c|lccccccccccccc}
\multicolumn{15}{c}{\large{(a) Evaluation based on ResNet-18 trained with the CE loss.} }                                                                                                                                                                                                                     \\
\multicolumn{15}{c}{}                                                                                                                                                                                                                                                                       \\ \toprule
\multirow{2}{*}{\begin{tabular}[c]{@{}l@{}}Training \\ Method\end{tabular}} & \multicolumn{1}{l|}{\multirow{2}{*}{\begin{tabular}[c]{@{}l@{}}Scoring Rule\end{tabular}}} & \multicolumn{7}{c|}{OOD detection benchmarks}                                                                                                                                               & \multicolumn{5}{c|}{OSR benchmarks}                                                                                                                                                                                                                                                            & \multirow{2}{*}{Overall} \\ \cmidrule(rl){3-14} 
\multicolumn{1}{l|}{}                                 & \multicolumn{1}{l|}{}                              & SVHN           & Textures       & LSUN           & LSUN-R         & iSUN           & Places365      & \multicolumn{1}{c|}{\begin{tabular}[c]{@{}c@{}}AVG\\ ID=95.45\end{tabular}} & \begin{tabular}[c]{@{}c@{}}CIFAR10\\ ID=97.13\end{tabular} & \begin{tabular}[c]{@{}c@{}}CIFAR+10\\ ID=96.6\end{tabular}  & \begin{tabular}[c]{@{}c@{}}CIFAR+50\\ ID=96.8\end{tabular}  & \begin{tabular}[c]{@{}c@{}}TinyImageNet\\ ID=83.4\end{tabular}  & \multicolumn{1}{c|}{AVG}            &                          \\ \midrule
\multicolumn{1}{c|}{\multirow{11}{*}{\begin{tabular}[c]{@{}l@{}}\rotatebox{90}{CE}\end{tabular}}}              & \multicolumn{1}{l|}{MSP}                           & 93.65          & 91.35          & 95.49          & 94.88          & 94.33          & 90.77          & \multicolumn{1}{c|}{93.41}                                                  & 91.78                                                      & 93.81                                                       & 90.20                                                       & 79.82                                                           & \multicolumn{1}{c|}{88.90}          & 91.61                    \\
\multicolumn{1}{c|}{}                                 & \multicolumn{1}{l|}{MLS}                           & 94.49          & 91.54          & 96.94          & 96.13          & 95.52          & 91.64          & \multicolumn{1}{c|}{94.38}                                                  & \underline{92.54}                                             & \underline{95.62}                                              & \underline{91.81}                                              & 81.31                                                           & \multicolumn{1}{c|}{\underline{90.32}} & \underline{92.53}           \\
\multicolumn{1}{c|}{}                                 & \multicolumn{1}{l|}{ODIN}                          & 92.23          & 83.76          & 94.96          & 96.16          & 95.31          & 90.88          & \multicolumn{1}{c|}{90.88}                                                  & 89.77                                                      & 81.37                                                       & 80.22                                                       & 80.96                                                           & \multicolumn{1}{c|}{83.08}          & 88.56                    \\
\multicolumn{1}{c|}{}                                 & \multicolumn{1}{l|}{GODIN}                         & \textbf{97.60} & \textbf{96.21} & \textbf{99.59} & \textbf{97.81} & \textbf{97.74} & \textbf{94.33} & \multicolumn{1}{c|}{\textbf{97.21}}                                         & 90.22                                                      & 91.17                                                       & 87.38                                                       & 76.05                                                           & \multicolumn{1}{c|}{86.21}          & 92.21                    \\
\multicolumn{1}{c|}{}                                 & \multicolumn{1}{l|}{SEM}                           & 75.65          & 72.02          & 75.18          & 70.93          & 72.52          & 76.14          & \multicolumn{1}{c|}{73.74}                                                  & 40.21                                                      & 43.87                                                       & 42.70                                                       & -                                                               & \multicolumn{1}{c|}{42.26}          & 61.15                    \\
\multicolumn{1}{c|}{}                                 & \multicolumn{1}{l|}{Energy}                        & \underline{94.64}          & \underline{91.64}          & \underline{97.14}          & \underline{96.29}          & \underline{95.68}          & \underline{91.78}          & \multicolumn{1}{c|}{\underline{94.53}}                                                  & 92.52                                             & \textbf{95.68}                                              & \textbf{91.86}                                              & 81.28                                                           & \multicolumn{1}{c|}{\textbf{90.34}} & \textbf{92.62}           \\
\multicolumn{1}{c|}{}                                 & \multicolumn{1}{l|}{MLS+ReAct}                     & 92.56          & 89.97          & 95.39          & 95.78          & 95.17          & 90.69          & \multicolumn{1}{c|}{93.26}                                                  & \textbf{92.57}                                             & 94.92                                                       & 90.88                                                       & \underline{81.65}                                                  & \multicolumn{1}{c|}{90.01}          & 91.78                    \\
\multicolumn{1}{c|}{}                                 & \multicolumn{1}{l|}{ODIN+ReAct}                    & 91.29          & 83.50          & 94.70          & 96.05          & 95.19          & 82.55          & \multicolumn{1}{c|}{90.55}                                                  & 86.65                                                      & 87.76                                                       & 88.40                                                       & 81.30                                                           & \multicolumn{1}{c|}{86.03}          & 88.74                    \\
\multicolumn{1}{c|}{}                                 & \multicolumn{1}{l|}{Energy+ReAct}                  & 92.68          & 90.05          & 95.67          & 96.03          & 95.42          & 90.89          & \multicolumn{1}{c|}{93.46}                                                  & 92.58                                                      & 95.02                                                       & 90.99                                                       & \textbf{81.67}                                                  & \multicolumn{1}{c|}{90.07}          & 91.92                    \\
\multicolumn{1}{c|}{}                                 & \multicolumn{1}{l|}{MLS+ASH}                    & 95.50          & 88.87          & 90.06          & 92.84          & 85.84          & 82.44          & \multicolumn{1}{c|}{89.26}                                                  & 89.19                                                      & 90.15                                                       & 82.11                                                       & 78.76                                                           & \multicolumn{1}{c|}{85.05}          & 87.58                    \\
\multicolumn{1}{c|}{}                                 & \multicolumn{1}{l|}{MLS+SHE}                    & 86.30          & 76.40          & 84.72          & 81.12          & 80.56          & 81.39          & \multicolumn{1}{c|}{81.75}                                                  & 79.19                                                      & 74.35                                                       & 78.02                                                       & 78.78                                                           & \multicolumn{1}{c|}{77.59}          & 80.09                    \\\bottomrule
\multicolumn{15}{c}{}                                                                                                                                                                                                                                                                                                                                                                                                                                                                                                                                                                                                      \\
\multicolumn{15}{c}{\large{(b) Evaluation based on ResNet-18 trained with the ARPL+CS loss.} }                                                                                                                                                                                                                                                                                                                                                                  \\
\multicolumn{15}{l}{}                                                                                                                                                                                                                                                                       \\ \toprule
\multirow{2}{*}{\begin{tabular}[c]{@{}l@{}}Training \\ Method\end{tabular}} & \multicolumn{1}{l|}{\multirow{2}{*}{Scoring Rule}} & \multicolumn{7}{c|}{OOD detection benchmarks}                                                                                                                                               & \multicolumn{5}{c|}{OSR benchmarks}                                                                                                                                                                                                                                                            & \multirow{2}{*}{Overall} \\ \cmidrule(rl){3-14}
\multicolumn{1}{l|}{}                                 & \multicolumn{1}{l|}{}                              & SVHN           & Textures       & LSUN           & LSUN-R         & iSUN           & Places365      & \multicolumn{1}{c|}{\begin{tabular}[c]{@{}c@{}}AVG\\ ID=91.02\end{tabular}} & \begin{tabular}[c]{@{}c@{}}CIFAR10\\ ID=96.96\end{tabular} & \begin{tabular}[c]{@{}c@{}}CIFAR+10\\ ID=96.77\end{tabular} & \begin{tabular}[c]{@{}c@{}}CIFAR+50\\ ID=96.69\end{tabular} & \begin{tabular}[c]{@{}c@{}}TinyImageNet\\ ID=86.91\end{tabular} & \multicolumn{1}{c|}{AVG}            &                          \\ \midrule
\multicolumn{1}{c|}{\multirow{11}{*}{\begin{tabular}[c]{@{}l@{}}\rotatebox{90}{ARPL+CS}\end{tabular}}}         & \multicolumn{1}{l|}{MSP}                           & 93.41          & 91.64          & 94.29          & 94.02          & 94.28          & 90.77          & \multicolumn{1}{c|}{93.07}                                                  & 92.53                                                      & 95.71                                                       & 94.03                                                       & \underline{82.80}                                                           & \multicolumn{1}{c|}{91.27}          & 92.41                    \\
\multicolumn{1}{c|}{}                                 & \multicolumn{1}{l|}{MLS}                           & \underline{96.36} & 90.20 & \underline{96.59} & \underline{96.95} & \underline{96.88} & \underline{93.29} & \multicolumn{1}{c|}{95.05}                                         & \underline{93.16}                                             & 96.58                                                       & 94.67                                                       & \textbf{84.79}                                                  & \multicolumn{1}{c|}{\textbf{92.30}}  & \textbf{93.95}           \\
\multicolumn{1}{c|}{}                                 & \multicolumn{1}{l|}{ODIN}                          & 75.92          & 71.64          & 86.25          & 95.14          & 95.19          & 75.97          & \multicolumn{1}{c|}{83.35}                                                  & 58.04                                                      & 74.80                                                       & 71.52                                                       & 63.13                                                           & \multicolumn{1}{c|}{66.87}          & 76.76                    \\
\multicolumn{1}{c|}{}                                 & \multicolumn{1}{l|}{GODIN}                         & 95.78          & 89.61          & 95.41          & 96.88          & 96.17          & 92.59          & \multicolumn{1}{c|}{94.41}                                                  & 91.99                                                      & 95.73                                                       & 93.76                                                       & 81.25                                                           & \multicolumn{1}{c|}{90.68}          & 92.92                    \\
\multicolumn{1}{c|}{}                                 & \multicolumn{1}{l|}{SEM}                           & 76.42          & 74.26          & 84.45          & 76.08          & 77.73          & 71.23          & \multicolumn{1}{c|}{76.70}                                                  & 35.01                                                      & 38.27                                                       & 44.15                                                       & -                                                               & \multicolumn{1}{c|}{39.14}          & 64.18                    \\
\multicolumn{1}{c|}{}                                 & \multicolumn{1}{l|}{Energy}                        & \textbf{96.52} & 90.11          & \textbf{96.76} & \textbf{97.16} & \textbf{97.07} & \textbf{93.45} & \multicolumn{1}{c|}{\underline{95.18}}                                         & \textbf{93.22}                                             & \textbf{96.74}                                              & \textbf{94.82}                                              & 82.10                                                           & \multicolumn{1}{c|}{\underline{91.72}}          & \underline{93.80}           \\
\multicolumn{1}{c|}{}                                 & \multicolumn{1}{l|}{MLS+ReAct}                     & 95.87          & \textbf{92.37} & 96.37          & 96.34          & 96.30          & 92.97          & \multicolumn{1}{c|}{95.04}                                                  & 92.70                                                      & 96.42                                                       & 94.53                                                       & 82.05                                                           & \multicolumn{1}{c|}{91.43}          & 93.59                    \\
\multicolumn{1}{c|}{}                                 & \multicolumn{1}{l|}{ODIN+ReAct}                    & 71.87          & 73.36          & 83.19          & 92.34          & 92.36          & 69.10          & \multicolumn{1}{c|}{80.37}                                                  & 55.71                                                      & 62.88                                                       & 61.85                                                       & 54.29                                                           & \multicolumn{1}{c|}{58.68}          & 71.70                    \\
\multicolumn{1}{c|}{}                                 & \multicolumn{1}{l|}{Energy+ReAct}                  & 96.06          & \underline{92.35}          & \underline{96.59} & 96.58          & 96.53          & 93.17          & \multicolumn{1}{c|}{\textbf{95.21}}                                         & 92.80                                                      & \underline{96.61}                                                       & \underline{94.70}                                                       & 82.14                                                           & \multicolumn{1}{c|}{91.56}          & 93.75          \\ 
\multicolumn{1}{c|}{}                                 & \multicolumn{1}{l|}{MLS+ASH}                    & 94.85          & 91.57          & 91.14          & 96.43          & 88.35          & 89.11          & \multicolumn{1}{c|}{91.91}                                                  & 91.89                                                      & 93.26                                                       & 91.81                                                       & 79.15                                                           & \multicolumn{1}{c|}{89.03}          & 90.76                    \\
\multicolumn{1}{c|}{}                                 & \multicolumn{1}{l|}{MLS+SHE}                    & 83.20          & 81.54          & 84.36          & 88.18          & 83.50          & 82.13          & \multicolumn{1}{c|}{83.82}                                                  & 77.10                                                      & 74.25                                                       & 74.85                                                       & 75.44                                                           & \multicolumn{1}{c|}{75.41}          & 80.46                    \\\bottomrule
\multicolumn{15}{c}{}                                                                                                                                                                                                                                                                                                                                                                                                                                                                                                                                                                                                      \\
\multicolumn{15}{c}{\large{(c) Evaluation based on ResNet-18 trained with the OE loss.}  }                                                                                                                                                                                                                                                                                                                                                                  \\
\multicolumn{15}{l}{}                                                                                                                                                                                                                                                                       \\ \toprule
\multirow{2}{*}{\begin{tabular}[c]{@{}l@{}}Training \\ Method\end{tabular}} & \multicolumn{1}{l|}{\multirow{2}{*}{Scoring Rule}} & \multicolumn{7}{c|}{OOD detection benchmarks}                                                                                                                                               & \multicolumn{5}{c|}{OSR benchmarks}                                                                                                                                                                                                                                                            & \multirow{2}{*}{Overall} \\ \cmidrule(rl){3-14} 
\multicolumn{1}{l|}{}                                 & \multicolumn{1}{l|}{}                              & SVHN           & Textures       & LSUN           & LSUN-R         & iSUN           & Places365      & \multicolumn{1}{c|}{\begin{tabular}[c]{@{}c@{}}AVG\\ ID=94.16\end{tabular}} & \begin{tabular}[c]{@{}c@{}}CIFAR10\\ ID=97.8\end{tabular}  & \begin{tabular}[c]{@{}c@{}}CIFAR+10\\ ID=98.3\end{tabular}  & \begin{tabular}[c]{@{}c@{}}CIFAR+50\\ ID=97.92\end{tabular} & \begin{tabular}[c]{@{}c@{}}TinyImageNet\\ ID=83.4\end{tabular}  & \multicolumn{1}{c|}{AVG}            &                          \\ \midrule
\multicolumn{1}{l|}{}                                 & \multicolumn{1}{l|}{MSP}                           & 99.21          & 98.81          & 99.02          & 98.52          & 98.55          & 97.29          & \multicolumn{1}{c|}{98.57}                                                  & 96.29                                                      & 99.29                                              & 98.70                                              & 78.67                                                           & \multicolumn{1}{c|}{93.24}          & \underline{96.44}           \\
\multicolumn{1}{c|}{\multirow{11}{*}{\begin{tabular}[c]{@{}l@{}}\rotatebox{90}{OE}\end{tabular}}}              & \multicolumn{1}{l|}{MLS}                           & 99.21          & \underline{98.82}          & 99.02          & 98.53          & 98.57          & \underline{97.32}          & \multicolumn{1}{c|}{\underline{98.58}}                                                  & 96.28                                                      & \underline{99.32}                                              & \underline{98.72}                                              & \textbf{80.19}                                                  & \multicolumn{1}{c|}{\textbf{93.63}} & \textbf{96.60}           \\
\multicolumn{1}{c|}{}                                 & \multicolumn{1}{l|}{ODIN}                          & \underline{99.43}          & 98.73          & \underline{99.14}          & \underline{98.78}          & \underline{98.75}          & 96.41          & \multicolumn{1}{c|}{98.54}                                                  & \underline{96.29}                                                      & 95.27                                                       & 94.30                                                       & 79.97                                                           & \multicolumn{1}{c|}{91.46}          & 95.71                    \\
\multicolumn{1}{c|}{}                                 & \multicolumn{1}{l|}{GODIN}                         & 97.25          & 95.17          & 89.05          & 83.42          & 84.63          & 89.51          & \multicolumn{1}{c|}{89.84}                                                  & 93.64                                                      & 92.01                                                       & 91.63                                                       & 78.21                                                           & \multicolumn{1}{c|}{88.87}          & 89.45                    \\
\multicolumn{1}{c|}{}                                 & \multicolumn{1}{l|}{SEM}                           & 98.13          & 97.04          & 98.77          & 97.01          & 97.16          & 94.86          & \multicolumn{1}{c|}{97.16}                                                  & 30.19                                                      & 33.73                                                       & 33.91                                                       & -                                                               & \multicolumn{1}{c|}{32.61}          & 73.69                    \\
\multicolumn{1}{c|}{}                                 & \multicolumn{1}{l|}{Energy}                        & 99.20          & 98.78          & 99.02          & 98.55          & 98.58          & 97.31          & \multicolumn{1}{c|}{98.57}                                                  & 93.12                                                      & \textbf{99.33}                                              & \textbf{98.74}                                              & \underline{80.16}                                                  & \multicolumn{1}{c|}{\underline{92.84}} & 96.28           \\
\multicolumn{1}{c|}{}                                 & \multicolumn{1}{l|}{GradNorm}                      & \textbf{99.95} & \textbf{99.71} & \textbf{99.83} & \textbf{99.46} & \textbf{99.42} & \textbf{97.93} & \multicolumn{1}{c|}{\textbf{99.38}}                                         & \textbf{96.57}                                             & 99.26                                              & 98.51                                              & 60.56                                                           & \multicolumn{1}{c|}{88.73}          & 95.12                    \\
\multicolumn{1}{c|}{}                                 & \multicolumn{1}{l|}{MLS+ReAct}                     & 95.18          & 92.22          & 79.46          & 83.34          & 83.68          & 87.46          & \multicolumn{1}{c|}{86.89}                                                  & 95.43                                                      & 98.73                                                       & 97.93                                                       & 79.92                                                           & \multicolumn{1}{c|}{93.00}          & 89.34                    \\
\multicolumn{1}{c|}{}                                 & \multicolumn{1}{l|}{ODIN+ReAct}                    & 84.16          & 82.92          & 64.00          & 73.90          & 75.45          & 71.65          & \multicolumn{1}{c|}{75.35}                                                  & 87.52                                                      & 87.78                                                       & 85.62                                                       & 79.47                                                           & \multicolumn{1}{c|}{85.10}          & 79.25                    \\
\multicolumn{1}{c|}{}                                 & \multicolumn{1}{l|}{Energy+ReAct}                  & 94.41          & 91.36          & 73.88          & 80.03          & 81.16          & 86.19          & \multicolumn{1}{c|}{84.51}                                                  & 95.43                                                      & 98.74                                                       & 78.67                                                       & 79.84                                                           & \multicolumn{1}{c|}{88.17}          & 85.97                    \\ 
\multicolumn{1}{c|}{}                                 & \multicolumn{1}{l|}{MLS+ASH}                    & 99.10          & 98.55          & 98.88          & 98.52          & 98.53          & 97.31          & \multicolumn{1}{c|}{98.49}                                                  & 95.40                                                      & 89.21                                                       & 91.54                                                       & 75.30                                                           & \multicolumn{1}{c|}{87.86}          & 94.24                    \\
\multicolumn{1}{c|}{}                                 & \multicolumn{1}{l|}{MLS+SHE}                    & 97.79          & 93.68          & 94.16          & 89.34          & 88.77          & 90.26          & \multicolumn{1}{c|}{92.33}                                                  & 82.01                                                      & 76.45                                                       & 87.05                                                       & 70.20                                                           & \multicolumn{1}{c|}{78.93}          & 86.97                    \\\bottomrule
\end{tabular}
}
\label{tab:small_scale_main}
\end{table*}

For \textit{specialized training}, we first experiment with the standard cross-entropy (CE) loss. 
We also use ARPL + CS \cite{chen2021adversarial} from the OSR literature.  
This method learns a set of `reciprocal points' which are trained to be far away from all training category embeddings. 
We note that the reciprocal points can be treated as a linear classification layer, allowing us to use any of the scoring rules mentioned above on top of this representation.
Finally, we train models with Outlier Exposure (OE)~\cite{hendrycks2019oe} from the OOD detection literature, where real outlier examples are used during training as examples of OOD detection.
In this case, the model is encouraged to predict a uniform softmax output.

\begin{figure*}[tb]
    \centering
    \includegraphics[width=0.8\linewidth]{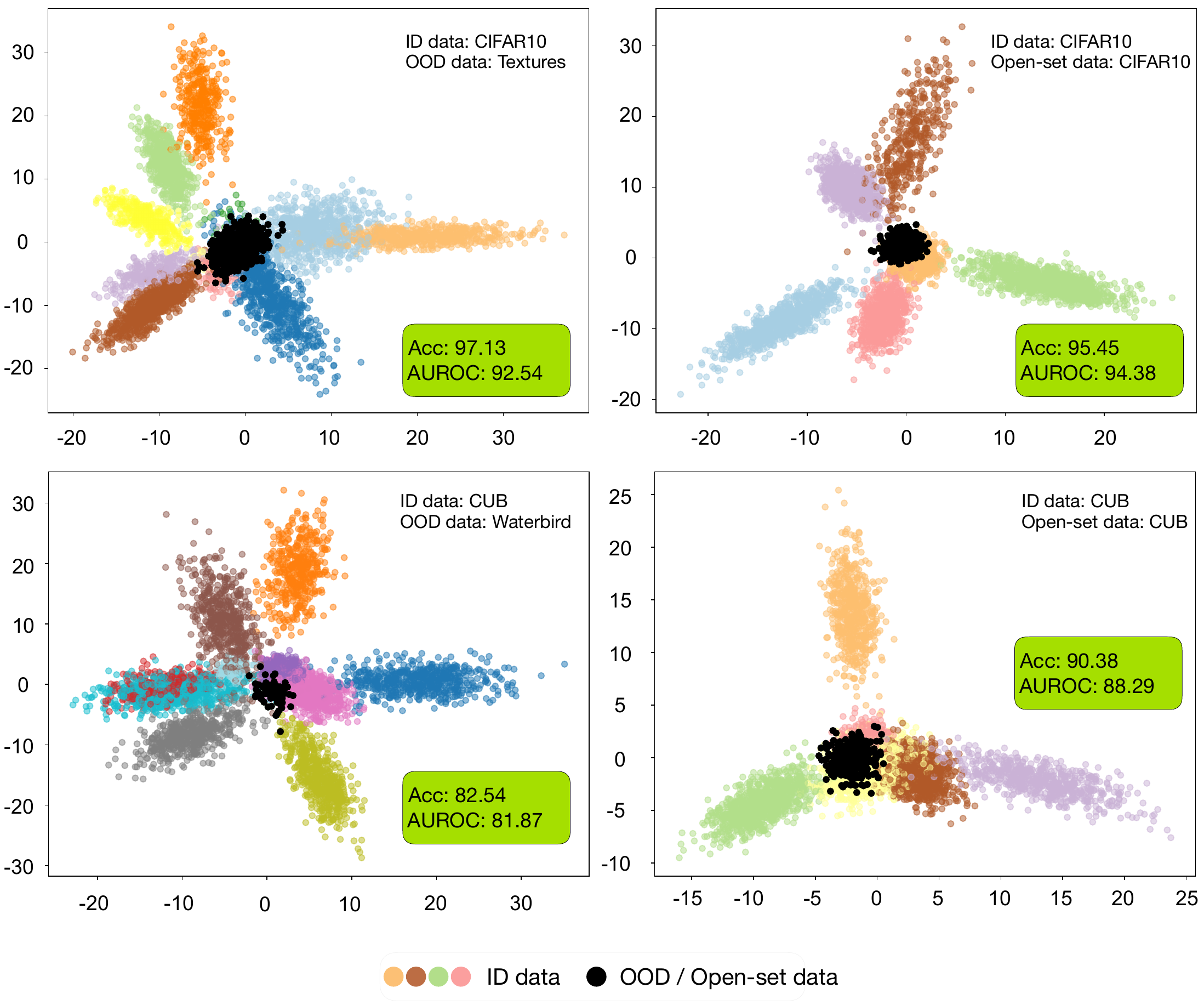}
    \caption{Visualization of feature projections for images from ID and open-set / OOD datasets. We project the features into a two-dimensional space using an additional linear layer with an output dimension of two after the penultimate layer. \whjpreb{We conduct OOD detection and OSR experiments using ResNet-18 on CIFAR-10 (first row) and ResNet-50 on CUB (second row) datasets. 
    For CIFAR-10, the OOD experiment uses the full CIFAR-10 dataset as ID data and Textures as OOD data, while the OSR experiment utilizes the first six classes in CIFAR-10 as ID data and the remaining four as open-set data. For CUB, the OOD experiment employs the full CUB dataset as ID and Waterbird as OOD data, while the OSR experiment uses six classes in CUB as ID data and four CUB classes as open-set data. These classes are randomly selected from the ID and open-set splits introduced in SSB. Notably, these visualizations reveal that the feature magnitudes of ID data exceed those of OOD or OSR data.}}
    \label{fig:reb_osr_ood}
\end{figure*}
\textbf{Datasets.}
For the OOD detection setting, we treat CIFAR10~\cite{krizhevsky2009learning} as in-distribution data and train models on it. 
As OOD data, we use six common datasets: SVHN \cite{cimpoi2014describing}, Textures \cite{ovadia2019can}, LSUN-Crop \cite{yu2015lsun}, LSUN-Resize \cite{yu2015lsun}, iSUN \cite{xu2015turkergaze} and Places365 \cite{zhou2017places}, all of which have mutually exclusive classes on CIFAR10.
\reb{In the supplementary, we also provide experiments using CIFAR-100 as the ID training data (see Tables S1-S3 in Section S2) and using different training configurations (see Tables S4-S7 in Section S3). The supplementary experiments yield consistent results, reinforcing the main findings to be discussed in this section.
}

For the OSR benchmark, following the standard protocols in \cite{Neal_2018_ECCV}, we set up four sub-tasks containing CIFAR10, CIFAR+10, CIFAR+50 and TinyImageNet \cite{le2015tiny}.
In all cases, models are trained on a subset of categories with remaining used as `unseen' at test time.
The CIFAR+N settings involve training on four classes from CIFAR10 and evaluating on $N$ classes from CIFAR-100.
Note that, for a given method, benchmarking on OOD detection involves training a single model and evaluating on multiple downstream datasets. In contrast, OSR benchmarks involve training a different model for each evaluation.

\textbf{Training configurations.}
We train the ResNet18 from scratch on all benchmarks. For CIFAR10, we always set the initial learning rate to 0.1 and apply the cosine annealing schedule, using SGD with the momentum of 0.9. The weight decay factor is set to $5e^{-4}$. The number of total training epochs is 200 and the batch size is 128.
For CIFAR-100, we also set the initial learning rate to 0.1, which is then divided by 5 at 60th, 120th, 160th epochs. The model is trained for 200 epochs with a batch size of 128, a weight decay of $5e^{-4}$, and Nesterov momentum of 0.9, following~\cite{devries2017improved}. 
\reb{Additional results using other network architectures and training setups can be found in~\Cref{apd:H} in Appendix.}

\textbf{Metrics.}
Following standard practise in both OOD and OSR tasks, we use the Area Under the Receiver Operating characteristic Curve (AUROC) as an evaluation metric throughout this paper as we find that other metrics were correlated strongly with the AUROC. \reb{For results on other metrics, please refer to the supplementary (Section S5).}

\subsection{Quantitative results}
\label{subsec:qr}

\whjpreb{In~\Cref{tab:small_bench}, we benchmark OOD detection and OSR tasks across nine common datasets, with different training strategies and scoring rules. The results are averaged from five independent runs. Although there is not always one clear winner regarding methodology, we have three main observations.}

\textbf{Firstly, MLS~\cite{vaze2022openset} and Energy~\cite{liu2020energy} tend to perform best across OOD and OSR datasets. }
We hypothesize that this is because both are sensitive to the magnitude of the feature vector before the networks' classification layer. 
To verify our conjecture, we investigate the magnitude of features by projecting the features of both ID and OOD/open-set samples into a two-dimensional space in~\Cref{fig:reb_osr_ood}. \whjpreb{We experiment on generic and fine-grained datasets, namely, CIFAR-10 and CUB.} This projection is achieved by training a linear layer with an output dimension of two, after the penultimate layer of the model. The feature magnitude of ID data is larger than that of open-set/OOD data.
This is consistent with the finding in~\cite{vaze2022openset} that `unfamiliar' examples tend to have lower feature magnitude than ID samples, providing a strong signal for distribution shift detection.

\textbf{Secondly, we observe that Outlier Exposure~\cite{hendrycks2019oe} provides excellent performance on the OOD detection benchmarks, often nearly saturating performance.} More results can be found in \Cref{apd:A} in Appendix.
It also often boosts OSR performance, though to a lesser degree, a phenomenon which we explore next in \Cref{sec:analysis_new}.

\begin{figure}[!ht]
\centering
\includegraphics[width=\linewidth]{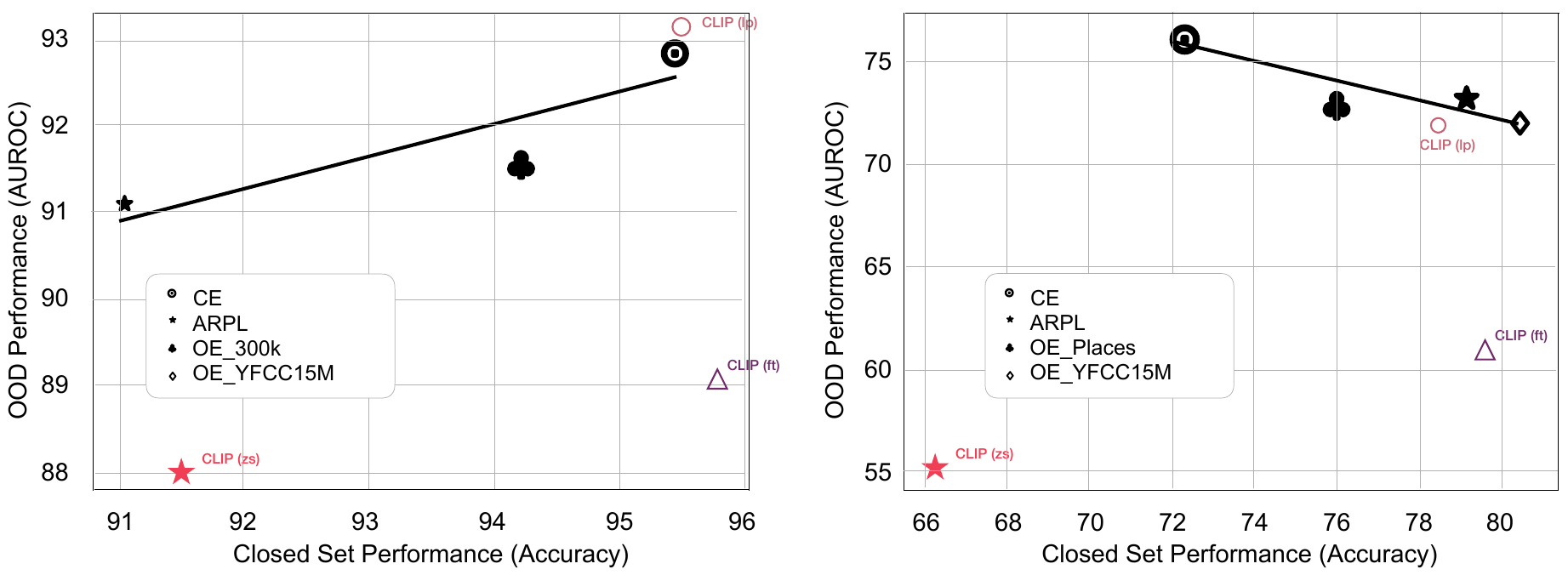}
\caption{\whjpreb{OSR performance \vs OOD detection performance of different training methods averaged across various scoring rules. 
CLIP variants are included here for reference and are not used to fit the correlation.}}\label{fig:clip_train_cor}
\end{figure}

\whjpreb{\textbf{Thirdly, for small-scale datasets, OOD detection accuracy is positively related to ID accuracy, while an inverse correlation is observed for large-scale datasets. }In~\Cref{fig:clip_train_cor}, we further include the results using the recent vision-language model, CLIP~\cite{Radford2021Learning}, for reference, which are not included for fitting the lines.
We experiment with three variants, namely, zero-shot (zs), finetuning (ft), and linear probing (lp). We find that only the linear probing CLIP falls into the correlation fitted for other methods, while the zero-shot and finetuning counterparts are not well aligned with the trend.}

\begin{figure}[!ht]
\centering
\includegraphics[width=\linewidth]{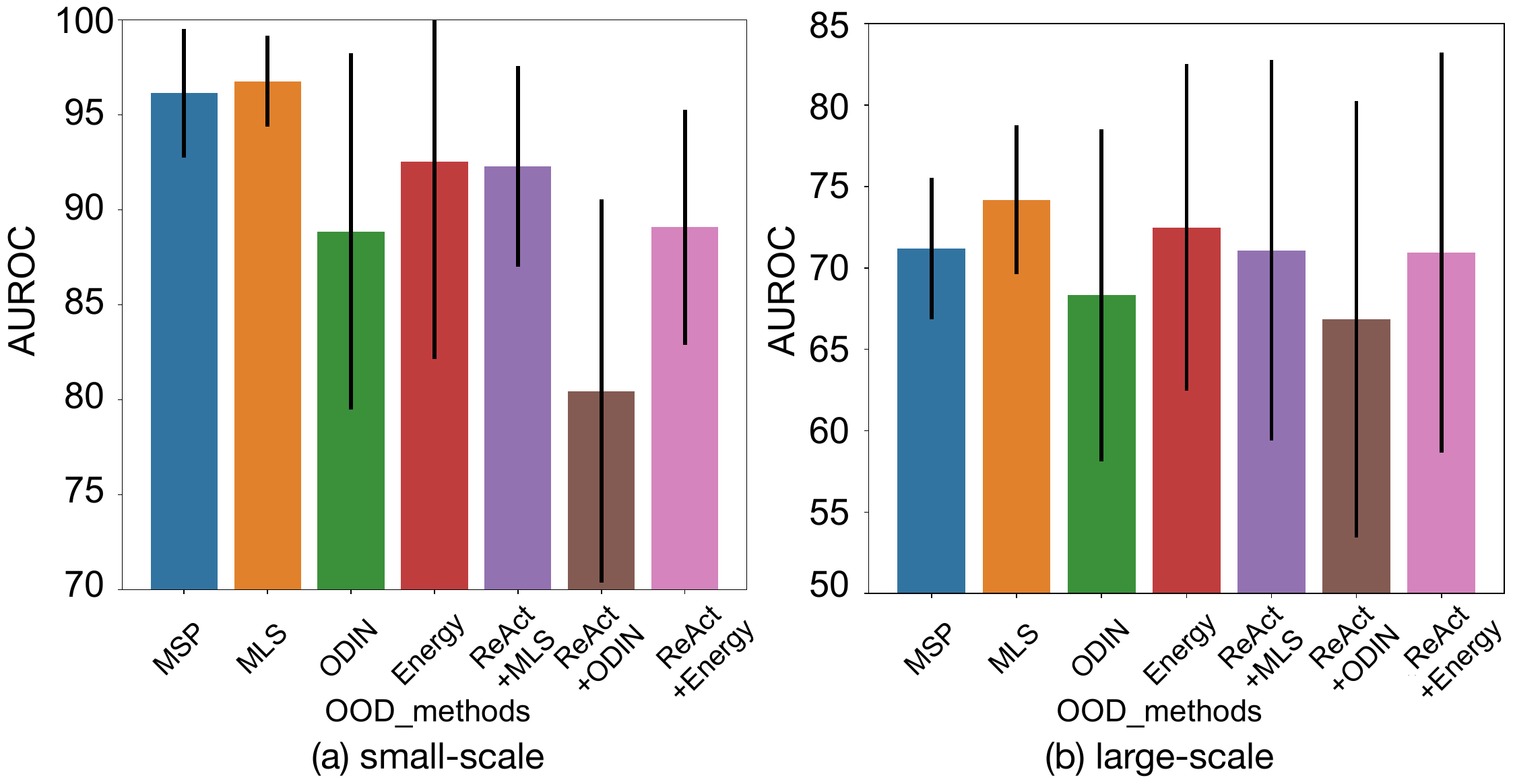}
\caption{\whj{OOD detection performance of various scoring rules averaged across different models.
Magnitude-aware scoring rules, particularly MLS, are the most efficient and stable techniques.}}\label{fig:clip_cor}
\end{figure}

\whjpreb{\textbf{Additional analysis.} (1) \textit{Mixup hurts magnitude-aware methods' performance.} To investigate the impact of different training setups, we also adopt Mixup~\cite{zhang2017mixup} to our training procedure. As shown in~\Cref{tab:arch3}, magnitude-aware techniques (\ie, MLS and Energy) demonstrate stability. Interestingly, we also find that the MSP method achieves the best performance compared to other methods, but it is still inferior to magnitude-aware methods' performance in~\Cref{tab:small_scale_main}(a). It appears that the Mixup mechanism, which distributes confidence to both involved categories, decreases the maximum magnitude value. Therefore, we highlight that the training setup (\eg, Mixup) that neglects magnitude information occurs poorer performance in both OOD detection and OSR tasks.
(2) \textit{Sensitivity of methods to hyper-parameters.} 
In~\Cref{fig:clip_cor}, we present the results of different scoring rules averaged across different training methods. For methods that are sensitive to the selection of hyperparameters (\eg, the perturbation magnitude of ODIN, thresholds of ReAct), we average the results among five different selections (instead of using five different random seeds like others). Therefore, in~\Cref{fig:clip_cor}, the result for each scoring rule is the average of 15 independent runs (\ie, 3 training methods $\times$ 5 runs). We find that magnitude-aware scoring rules (\ie, MLS and Energy) offer obvious advantages for evaluating model performance. Considering the error bars of MLS and Energy, MLS is the optimal choice for all the scenarios. Furthermore, we observe that ODIN and ReAct, two techniques that are not magnitude-aware, exhibit instability. 
This instability can be attributed to the reliance on a carefully tuned noise value for stochastic predictions for ODIN and threshold value to truncate activation for ReAct.
\begin{table*}[!h]
\centering
\caption{\whjpreb{Evaluation on small-scale OOD detection and OSR benchmarks with various scoring rules on ResNet-18, using CIFAR10 as ID training data. The results are averaged from five independent runs. We adopt Mixup~\cite{zhang2017mixup} to the training procedure and report the in-distribution accuracy as `ID'. Bold values represent the best results, while underlined values represent the second best results. Magnitude-aware techniques (\ie, MLS and Energy) demonstrate stability.}}\label{tab:arch3}
\resizebox{\linewidth}{!}{
\begin{tabular}{c|lccccccccccccc}
\toprule
\multirow{2}{*}{\begin{tabular}[c]{@{}l@{}}Training \\ Method\end{tabular}} & \multicolumn{1}{l|}{\multirow{2}{*}{\begin{tabular}[c]{@{}l@{}}Scoring Rule\end{tabular}}} & \multicolumn{7}{c|}{OOD detection benchmarks}                                                                                                                                               & \multicolumn{5}{c|}{OSR benchmarks}                                                                                                                                                                                                                                                            & \multirow{2}{*}{Overall} \\ \cmidrule(rl){3-14} 
\multicolumn{1}{l|}{}                                 & \multicolumn{1}{l|}{}                              & SVHN           & Textures       & LSUN           & LSUN-R         & iSUN           & Places365      & \multicolumn{1}{c|}{\begin{tabular}[c]{@{}c@{}}AVG\\ ID=95.81\end{tabular}} & \begin{tabular}[c]{@{}c@{}}CIFAR10\\ ID=97.40\end{tabular} & \begin{tabular}[c]{@{}c@{}}CIFAR+10\\ ID=96.82\end{tabular}  & \begin{tabular}[c]{@{}c@{}}CIFAR+50\\ ID=96.91\end{tabular}  & \begin{tabular}[c]{@{}c@{}}TinyImageNet\\ ID=83.75\end{tabular}  & \multicolumn{1}{c|}{AVG}            &                          \\ \midrule
\multicolumn{1}{c|}{\multirow{9}{*}{\begin{tabular}[c]{@{}l@{}}\rotatebox{90}{CE+Mixup}\end{tabular}}}              & \multicolumn{1}{l|}{MSP}                           & 87.63          & \underline{81.48}          & \textbf{95.01}         & 88.22          & \underline{88.58}          & \textbf{86.02}          & \multicolumn{1}{c|}{\textbf{87.82}}                                                  & 92.01                                                      & 93.96                                                       & 90.44                                                       & 80.02                                                           & \multicolumn{1}{c|}{89.11}          & \textbf{88.34}                    \\
\multicolumn{1}{c|}{}                                 & \multicolumn{1}{l|}{MLS}                           & 86.83          & 79.62          & \underline{94.95}          & 88.05          & 87.99          & 81.99          & \multicolumn{1}{c|}{86.57}                                                  & \underline{92.88}                                             & \textbf{95.99}                                              & \underline{92.16}                                              & \textbf{81.55}                                                           & \multicolumn{1}{c|}{\textbf{90.65}} & \underline{88.17}           \\
\multicolumn{1}{c|}{}                                 & \multicolumn{1}{l|}{ODIN}                          & 89.43          & 75.56          & 87.31          & 55.94          & 61.36          & 83.63          & \multicolumn{1}{c|}{75.54}                                                  & 85.27                                                      & 79.24                                                       & 83.51                                                       & 78.81                                                           & \multicolumn{1}{c|}{81.71}          & 78.01                    \\
\multicolumn{1}{c|}{}                                 & \multicolumn{1}{l|}{GODIN}                         & \textbf{94.28} & \textbf{89.71} & 91.52 & \underline{88.50} & 87.45 & 81.28 & \multicolumn{1}{c|}{88.79}                                         & 90.13                                                      & 91.01                                                       & 86.88                                                       & 72.48                                                           & \multicolumn{1}{c|}{85.13}          & 87.33                    \\                              & \multicolumn{1}{l|}{Energy}                        & 78.87          & 76.46          & 93.21          & 88.09          & 85.69          & 78.53          & \multicolumn{1}{c|}{83.48}                                                  & 92.79                                             & \underline{95.92}                                              & 92.10                                              & \underline{81.53}                                                           & \multicolumn{1}{c|}{\underline{90.59}} & 86.32           \\
\multicolumn{1}{c|}{}                                 & \multicolumn{1}{l|}{MLS+ReAct}                     & \underline{91.47}          & 77.25          & 93.89          & \textbf{88.73}          & \textbf{88.84}         & \underline{84.32}          & \multicolumn{1}{c|}{\underline{87.42}}                                                  & 90.33                                             & 95.02                                                       & \textbf{92.18}                                                       & 79.42                                                  & \multicolumn{1}{c|}{89.24}          & 88.15                    \\
\multicolumn{1}{c|}{}                                 & \multicolumn{1}{l|}{ODIN+ReAct}                    & 72.19          & 57.63          & 69.15          &     54.50      & 54.62          & 72.75          & \multicolumn{1}{c|}{63.47}                                                  & 83.30                                                      & 81.02                                                       & 85.82                                                       & 77.21                                                           & \multicolumn{1}{c|}{81.84}          & 70.82                    \\
\multicolumn{1}{c|}{}                                 & \multicolumn{1}{l|}{Energy+ReAct}                  & 89.80          & 75.02          & 93.62          & 87.11          & 87.40          & 81.76          & \multicolumn{1}{c|}{85.79}                                                  & \textbf{93.00}                                                      & 91.69                                                       & 86.45                                                       & 81.28                                                  & \multicolumn{1}{c|}{88.11}          & 86.66                    \\ \bottomrule
\end{tabular}
}
\end{table*}
We can also find from~\Cref{tab:small_scale_main} that ReAct, which has been shown to be effective in the literature, does not seem to bring performance gain in well-trained models with a high in-distribution accuracy. 
Here, we follow the techniques from~\cite{vaze2022openset} to obtain the highest ID accuracy possible. 
It appears that when the classifier is strong enough, it is difficult for ReAct to bring extra improvement.
Besides, ReAct is sensitive to the choice of the activation pruning percentile. The optimal percentile values are different for different open-set/OOD datasets (see Section S4 in the supplementary).
For identifying out-of-distribution inputs, we recommend more stable and deterministic magnitude-aware scoring rules.}

\subsection{Qualitative analysis}
\label{sec:representations}

\begin{figure*}[!ht]
\centering
\includegraphics[width=\linewidth]{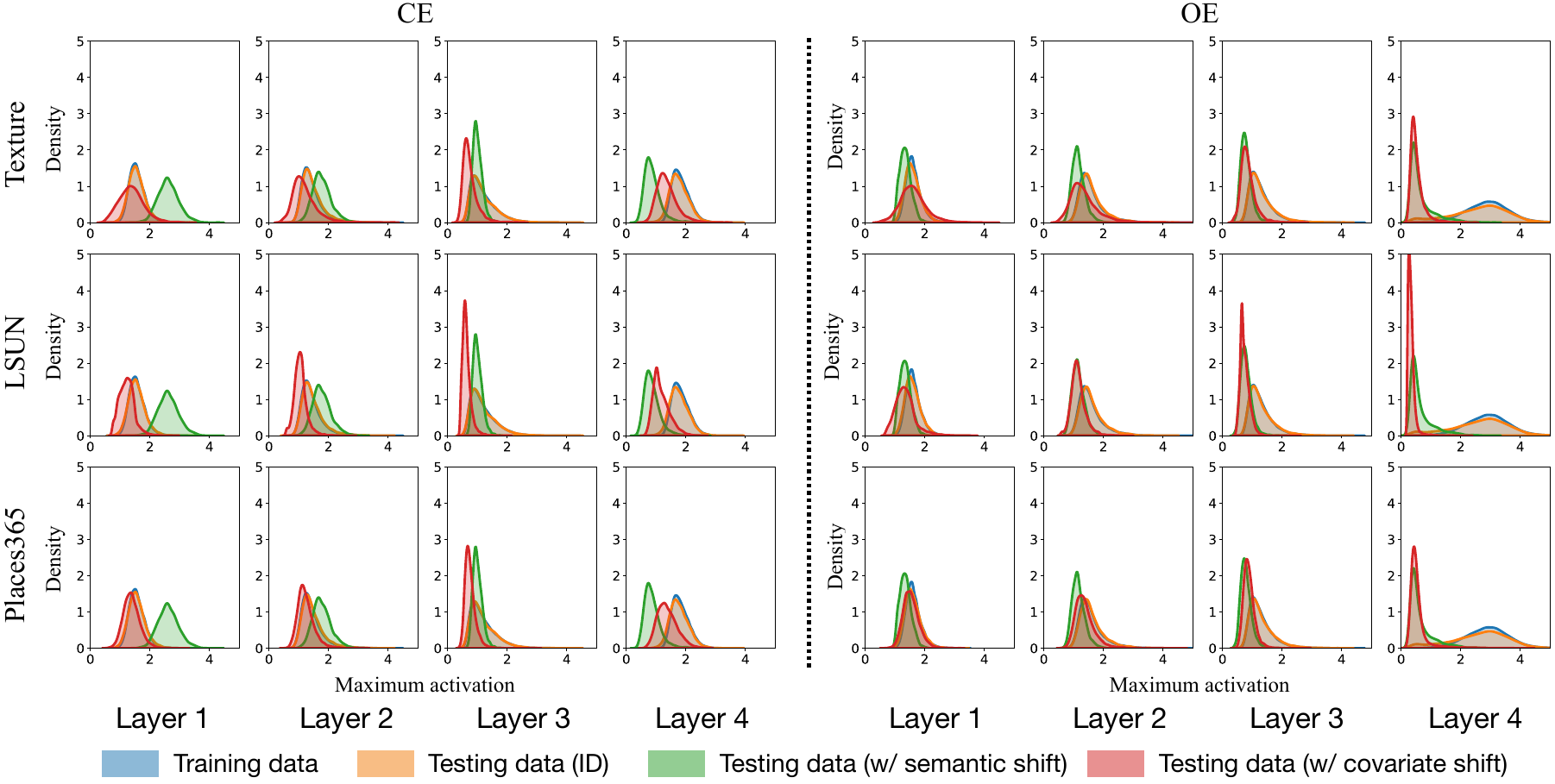}
\caption{Histogram of activations for ResNet-18 pretrained on a subset of CIFAR10 with four training classes and evaluated on: training and ID testing data; open-set data (disjoint six classes in CIFAR10) and OOD data (from Textures, LSUN and Places365). Specifically, each subplot shows the maximum activation (along channel, width and height dimension) at the outputs from \texttt{layer\_1} to \texttt{layer\_4} of ResNet-18, displayed from left to right in the figures. 
The behavior of OE is different from CE, whose activation maps become more separable in the deeper rather than the shallower layers. \reb{See~\Cref{apd:B} in Appendix for results on more datasets.}}\label{fig:second}
\end{figure*}

In this section, we qualitatively interrogate the learned representations of Cross-Entropy and Outlier Exposure networks in order to explain the stark performance boost of OE on existing OOD detection benchmarks. 
Specifically, we use the value of the maximally activated neuron at various layers to analyze how the networks respond to distribution shifts.
We pass every sample through the network, and plot the histogram of maximum activations at every layer in~\Cref{fig:second} \reb{(see~\Cref{fig:reb_arpl} for the analogous results by training with the ARPL+CS method)}.

This is inspired by~\cite{vaze2022openset}, who show the `maximum logit score' (MLS, the maximum activation at a network's output layer) can achieve SOTA for OSR. 
Furthermore,~\cite{dietterich2022openset} propose that networks respond to a `lack of familiarity' under distribution shift by failing to light in-distribution activation pathways.
We investigate how activations at various stages of a deep network vary under different `unseen' datasets. \Cref{fig:second} shows histograms of the maximum activations at the outputs from \texttt{layer\_1} to \texttt{layer\_4} of ResNet-18~\cite{he2016deep} trained on CIFAR10 when evaluated on data with different shifts
(here we use `layer' to refer to ResNet block).

For open-set data, we find that early layer activations are largely the same as for the ID test data.
It is only later in the network that the activation patterns begin to differ. 
This is intuitive as the low-level textures and statistics of the open-set data do not vary too much from the training images. 
Furthermore, it has long been known that early filters in CNNs tend to focus on textural details such as edges~\cite{Krizhevsky2012alexnet}.
In contrast, we find that some OOD datasets, such as SVHN, induce very different activations in the early layers.
Our explanation for this phenomenon is analogous: SVHN contains very different image statistics and low-level features to the training dataset of CIFAR10, and hence induces different activations in early layers. 
Most interestingly, however, some datasets which show markedly different early layer activations actually display \textit{more similar} activations at later layers (like SVHN, see~\Cref{fig:layers_app}).

Meanwhile, OE displays show substantially different intermediate activations. Interestingly, the maximum activation in early layers look very similar to the ID testing data, but tend to be less so later on in the network. It is clear that activations in later layers are more discriminative after using OE loss when compared with using CE loss.

\section{Disentangling distribution shifts}
\label{sec:analysis_new}

Having analysed methodologies for detecting distribution shift across the OOD detection and OSR settings, we turn our attention to the benchmarks. 
While it is clear that OSR specifically aims to detect unseen categories, there is no specification of the type of distribution shift which OOD detection benchmarks aim to capture, or how they would relate to a real-world scenario.
In this section, we propose a lens through which to consolidate types of distribution shift.
Specifically, we propose that `distribution shift' can be parameterised along two broad, orthogonal, axes: 
\textit{semantic} shift and \textit{covariate} shift.
Pure semantic shift is when new categories are encountered, and is the explicit focus of OSR, while covariate shift refers to the setting when the semantics of test images remain constant, but other features change.

Formally, similarly to~\cite{wiles2022fine}, we consider a latent variable model of the data generation process, with latent $z$:
\begin{equation}
    \medmuskip=5mu
    \thickmuskip=0.1mu
    \renewcommand\arraystretch{1.5}
    z \sim p(z) \quad\quad y^i \sim p(y^i|z) \quad i \in \{1...L\} \quad\quad \mathbf{x} \sim p(\mathbf{x}|z)
\end{equation}

Here, $\mathbf{x}$ is an image and $y^i$ represents an image attribute.
The set of attributes could include traditional features such as `color' or `texture', or refer to more abstract features such as `beak shape' of a bird. 
We define a set of semantic attributes, $Y_S$, such that the category label of an image is a function of these attributes.
Furthermore, we define covariate attributes, $Y_C$, which can be freely varied without the category label changing.
In this framing, given marginal training distributions $p_{train}(Y_S)$ and $p_{train}(Y_C)$, detecting semantic shift is the task of flagging when $p_{test}(Y_S) \neq p_{train}(Y_S)$.
Analogously, we wish to flag covariate shift if $p_{test}(Y_C) \neq p_{train}(Y_C)$.

To motivate this setting, consider the perceptual system in an autonomous vehicle, which has been trained to recognize \textit{cars} during the \textit{day}. 
A \textit{semantic shift} detector is necessary for when the system encounters a new category, \eg, to flag that \textit{bicycle} is an unknown concept.
Meanwhile, a \textit{covariate shift} detector is necessary for when the system is deployed at \textit{night-time}, where the categories may be familiar, but the performance of the system could be expected to degrade.

\subsection{Datasets} 
\begin{table*}[!ht]
\caption{Results of OOD detection and OSR benchmarks on large-scale datasets, using ResNet-50 model trained with the OE loss compared with CE and ARPL baselines. The results are averaged from five independent runs. We separately introduce outlier data from different data sources including Places and YFCC15M to feed OE. 
}\label{tab:large_oe}
\centering
\resizebox{\linewidth}{!}{
\begin{tabular}{l|c|ccc|ccccccccc|c}
\toprule
\multirow{2}{*}{\begin{tabular}[c]{@{}l@{}}Training \\ Method\end{tabular}} & \multicolumn{1}{l|}{\multirow{2}{*}{Scoring Rule}} & \multicolumn{3}{c|}{Covariate Shift}             & \multicolumn{9}{c|}{Semantic Shift}                                                                                                                                                                                                                                                                                                                                  & \multirow{2}{*}{Overall} \\ \cmidrule(rl){3-14}
                                                                            & \multicolumn{1}{l|}{}                              & ImageNet-C     & ImageNet-R     & AVG            & \multicolumn{2}{c}{\begin{tabular}[c]{@{}c@{}}ImageNet-SSB\\ (Easy/Hard)\end{tabular}} & \multicolumn{2}{c}{\begin{tabular}[c]{@{}c@{}}CUB\\ (Easy/Hard)\end{tabular}} & \multicolumn{2}{c}{\begin{tabular}[c]{@{}c@{}}Scars\\ (Easy/Hard)\end{tabular}} & \multicolumn{2}{c}{\begin{tabular}[c]{@{}c@{}}FGVC\\ (Easy/Hard)\end{tabular}} & \multicolumn{1}{l|}{AVG} &                          \\ \midrule
CE                                                                          & \multirow{4}{*}{MLS}                               & \textbf{67.92} & \textbf{86.71} & \textbf{77.32} & 80.28                                      & 75.05                                     & \textbf{88.29}                        & \textbf{79.33}                        & 94.03                                  & 82.24                                  & \textbf{90.65}                         & \textbf{82.55}                        & \textbf{84.05}           & \textbf{82.71}           \\
ARPL+CS                                                                     &                                                    & 63.94          & 82.77          & 73.36          & 79.92                                      & 74.60                                     & 83.50                                 & 75.49                                 & \textbf{94.78}                         & \textbf{83.63}                         & 87.04                                  & 77.71                                 & 82.08                    & 80.34                    \\
OE (w/ Places)                    &                                                    & 61.77          & 80.53          & 71.15          & \textbf{82.42}                             & \textbf{75.58}                            & 79.16                                 & 73.83                                 & 91.02                                  & 78.69                                  & 88.38                                  & 79.19                                 & 80.81                    & 78.88                    \\
OE (w/ YFCC15M)                  &                                                    & 64.12          & 82.01          & 73.07          & 79.37                                      & 72.55                                     & 75.19                                 & 70.28                                 & 84.03                                  & 71.34                                  & 74.20                                  & 66.63                                 & 71.12                    & 71.51                    \\ \bottomrule
\end{tabular}
}
\end{table*}

\begin{table*}[!ht]
\caption{Evaluation on large-scale OOD detection and OSR benchmarks using ResNet-50 model trained with different losses and scoring rules. The results are averaged from five independent runs. Bold values represent
the best results, while underlined values represent the second best results. Models trained with the CE loss outperforms the ones with ARPL on both covariate shift and semantic shift.}\label{tab:large_bench}
\centering
\resizebox{\linewidth}{!}{
\begin{tabular}{clccccccccccc}
\multicolumn{13}{c}{(a) Evaluation based on ResNet-50 trained with the CE loss.}            \\                                                                                                                      \multicolumn{13}{c}{}                                                                                                                                              \\ \toprule
\multicolumn{1}{l|}{\multirow{2}{*}{\begin{tabular}[c]{@{}l@{}}Training \\ Method\end{tabular}}}        & \multicolumn{1}{l|}{\multirow{2}{*}{Scoring Rule}} & \multicolumn{5}{c|}{Covariate Shift}                                                                                                                                                                                                                      & \multicolumn{5}{c|}{Semantic Shift}                                                                                                                                                                          & \multirow{2}{*}{Overall} \\ \cmidrule(rl){3-12}
\multicolumn{1}{l|}{}                                                                                   & \multicolumn{1}{l|}{}                              & ImageNet-C & ImageNet-R & \multicolumn{2}{c}{\begin{tabular}[c]{@{}c@{}}Waterbird\\ (Easy/Hard)\end{tabular}} & \multicolumn{1}{c|}{AVG}            & \multicolumn{2}{c}{\begin{tabular}[c]{@{}c@{}}ImageNet-SSB\\ (Easy/Hard)\end{tabular}} & \multicolumn{2}{c}{\begin{tabular}[c]{@{}c@{}}CUB\\ (Easy/Hard)\end{tabular}} & \multicolumn{1}{c|}{AVG}            &                          \\ \midrule
\multicolumn{1}{c|}{\multirow{7}{*}{\begin{tabular}[c]{@{}c@{}}\rotatebox{90}{CE}\end{tabular}}}                     & \multicolumn{1}{l|}{MSP}                           & 64.63                                                         & 80.53                                                         & 81.65                                    & \textbf{75.33}                           & \multicolumn{1}{c|}{75.54}          & 80.16                                      & 75.01                                     & 88.11                                 & \textbf{79.43}                        & \multicolumn{1}{c|}{80.68}          & 78.11                    \\
\multicolumn{1}{c|}{}                                                                                   & \multicolumn{1}{l|}{MLS}                           & \underline{67.92}                                                   & \underline{86.71}                                                   & 81.87                                    & \underline{75.18}                              & \multicolumn{1}{c|}{\underline{77.92}}    & \textbf{80.28}                             & 75.05                                     & 88.29                                 & \underline{79.33}                           & \multicolumn{1}{c|}{\textbf{80.74}} & \underline{79.33}              \\
\multicolumn{1}{c|}{}                                                                                   & \multicolumn{1}{l|}{ODIN}                          & 63.69                                                         & 85.62                                                         & 79.51                                    & 71.54                                    & \multicolumn{1}{c|}{75.09}          & 74.56                                      & \underline{75.27}                               & 86.24                                 & 73.88                                 & \multicolumn{1}{c|}{77.49}          & 76.29                    \\
\multicolumn{1}{c|}{}                                                                                   & \multicolumn{1}{l|}{Energy}                        & \textbf{68.05}                                                & \textbf{87.04}                                                & \textbf{82.49}                           & 74.60                                    & \multicolumn{1}{c|}{\textbf{78.05}} & 79.76                                      & 74.96                                     & \textbf{88.81}                        & 79.06                                 & \multicolumn{1}{c|}{80.65}          & \textbf{79.35}           \\
\multicolumn{1}{c|}{}                                                                                   & \multicolumn{1}{l|}{MLS+ReAct}                     & 66.64                                                         & 84.82                                                         & 81.69                                    & 75.12                                    & \multicolumn{1}{c|}{77.07}          & \textbf{80.28}                             & 75.07                                     & 88.29                                 & \underline{79.33}                           & \multicolumn{1}{c|}{\textbf{80.74}} & 78.91                    \\
\multicolumn{1}{c|}{}                                                                                   & \multicolumn{1}{l|}{ODIN+ReAct}                    & 61.69                                                         & 83.25                                                         & 79.48                                    & 71.50                                    & \multicolumn{1}{c|}{73.98}          & 74.56                                      & \textbf{75.29}                            & 86.24                                 & 73.88                                 & \multicolumn{1}{c|}{77.49}          & 75.74                    \\
\multicolumn{1}{c|}{}                                                                                   & \multicolumn{1}{l|}{Energy+ReAct}                  & 66.88                                                         & 83.92                                                         & \underline{82.48}                              & 74.55                                    & \multicolumn{1}{c|}{76.96}          & 79.76                                      & 74.99                                     & \textbf{88.81}                        & 79.06                                 & \multicolumn{1}{c|}{80.66}          & 78.81                    \\ \bottomrule
\\
\multicolumn{13}{c}{\multirow{2}{*}{(b) Evaluation based on ResNet-50 trained with the ARPL loss.}}                                                                                                                                                                                                                                                                                                                                                                                                                                                                                                                                                                \\
\multicolumn{13}{c}{}                                                                                                                                                                                                                                                                                                                                                                                                                                                                                                                                                                                                                                              \\ \toprule
\multicolumn{1}{l|}{\multirow{2}{*}{\begin{tabular}[c]{@{}l@{}}Training \\ Method\end{tabular}}}        & \multicolumn{1}{l|}{\multirow{2}{*}{Scoring Rule}} & \multicolumn{5}{c|}{Covariate Shift}                                                                                                                                                                                                                      & \multicolumn{5}{c|}{Semantic Shift}                                                                                                                                                                          & \multirow{2}{*}{Overall} \\ \cmidrule(rl){3-12}
\multicolumn{1}{l|}{}                                                                                   & \multicolumn{1}{l|}{}                              & ImageNet-C & ImageNet-R & \multicolumn{2}{c}{\begin{tabular}[c]{@{}c@{}}Waterbird\\ (Easy/Hard)\end{tabular}} & \multicolumn{1}{c|}{AVG}            & \multicolumn{2}{c}{\begin{tabular}[c]{@{}c@{}}ImageNet-SSB\\ (Easy/Hard)\end{tabular}} & \multicolumn{2}{c}{\begin{tabular}[c]{@{}c@{}}CUB\\ (Easy/Hard)\end{tabular}} & \multicolumn{1}{c|}{AVG}            &                          \\ \midrule
\multicolumn{1}{c|}{\multirow{7}{*}{\begin{tabular}[c]{@{}c@{}}\rotatebox{90}{ARPL+CS}\end{tabular}}} & \multicolumn{1}{l|}{MSP}                           & 61.85                                                         & 78.68                                                         & 79.42                                    & \textbf{72.30}                           & \multicolumn{1}{c|}{73.06}          & 79.90                                      & \textbf{74.67}                            & 83.53                                 & \textbf{75.64}                        & \multicolumn{1}{c|}{\textbf{78.44}} & 75.75                    \\
\multicolumn{1}{c|}{}                                                                                   & \multicolumn{1}{l|}{MLS}                           & \underline{63.94}                                                   & \underline{82.77}                                                   & 79.48                                    & 72.09                                    & \multicolumn{1}{c|}{\underline{74.57}}    & \textbf{79.92}                             & \underline{74.60}                               & 83.50                                 & \underline{75.49}                           & \multicolumn{1}{c|}{\underline{78.38}}    & \underline{76.47}              \\
\multicolumn{1}{c|}{}                                                                                   & \multicolumn{1}{l|}{ODIN}                          & 61.88                                                         & 77.03                                                         & 73.76                                    & 69.26                                    & \multicolumn{1}{c|}{70.48}          & 68.72                                      & 71.23                                     & 73.87                                 & 69.77                                 & \multicolumn{1}{c|}{70.90}          & 70.69                    \\
\multicolumn{1}{c|}{}                                                                                   & \multicolumn{1}{l|}{Energy}                        & \textbf{64.13}                                                & \textbf{83.25}                                                & \textbf{79.64}                           & 71.86                                    & \multicolumn{1}{c|}{\textbf{74.72}} & 79.87                                      & 74.49                                     & \textbf{83.70}                        & 75.46                                 & \multicolumn{1}{c|}{\underline{78.38}}    & \textbf{76.55}           \\
\multicolumn{1}{c|}{}                                                                                   & \multicolumn{1}{l|}{MLS+ReAct}                     & 62.69                                                         & 80.69                                                         & 79.44                                    & 72.07                                    & \multicolumn{1}{c|}{73.72}          & \textbf{79.92}                             & \underline{74.60}                               & 83.44                                 & 75.43                                 & \multicolumn{1}{c|}{78.35}          & 76.04                    \\
\multicolumn{1}{c|}{}                                                                                   & \multicolumn{1}{l|}{ODIN+ReAct}                    & 62.23                                                         & 76.08                                                         & 73.75                                    & 69.23                                    & \multicolumn{1}{c|}{70.32}          & 68.72                                      & 71.23                                     & 67.42                                 & 63.91                                 & \multicolumn{1}{c|}{67.82}          & 69.07                    \\
\multicolumn{1}{c|}{}                                                                                   & \multicolumn{1}{l|}{Energy+ReAct}                  & 62.89                                                         & 81.17                                                         & \underline{79.60}                              & 71.83                                    & \multicolumn{1}{c|}{73.87}          & 79.87                                      & 74.49                                     & \textbf{83.70}                        & 75.41                                 & \multicolumn{1}{c|}{78.37}          & 76.12                    \\ \bottomrule
\end{tabular}
}
\end{table*}

As a starting point, we note that~\cite{vaze2022openset} introduced the Semantic Shift Benchmark (SSB), a distribution shift benchmark with isolates \textit{semantic shift}.
We mainly focus on ImageNet-SSB \cite{ILSVRC15} and CUB-SSB \cite{WahCUB_200_2011} datasets. 
`Seen' classes in ImageNet-SSB are the original ImageNet-1K classes, while `unseen' classes selected from the disjoint set of ImageNet-21K-P \cite{ridnik2021imagenet}. 
Meanwhile, CUB-SSB splits the 200 bird classes in CUB into `seen' and `unseen' categories.
Furthermore, the unseen categories are split into Easy and Hard classes by their attributes, and the splitting rule depends on semantic similarity of every pair of visual attributes in the unknown classes and the training classes. 
For all the above datasets, categories appearing in the training set would not be included in the evaluation set.

For \textit{covariate shift}, we propose ImageNet-C \cite{hendrycks2019robustness} and ImageNet-R \cite{Hendrycks2020TheMF} to demonstrate distribution shift with respect to the standard ImageNet dataset.
Both datasets contain images from a subset of the ImageNet-1K categories, but with different low-level image statistics.
ImageNet-C applies four main corruptions (\eg, noise, blur, weather, and digital) with varying intensities to the validation images of ImageNet-1K, while ImageNet-R collects various artistic renditions of foreground classes from the ImageNet-1K dataset.
We also choose Waterbirds \cite{sagawa2019distributionally} to test the model trained on the CUB-SSB `Seen' classes. 
Waterbirds inserts bird photographs from the CUB dataset into backgrounds picked from the Places dataset \cite{zhou2017places}, meaning it has the same semantic categories to CUB but in different contexts.

\begin{figure*}[!ht]
\centering
\includegraphics[width=\linewidth]{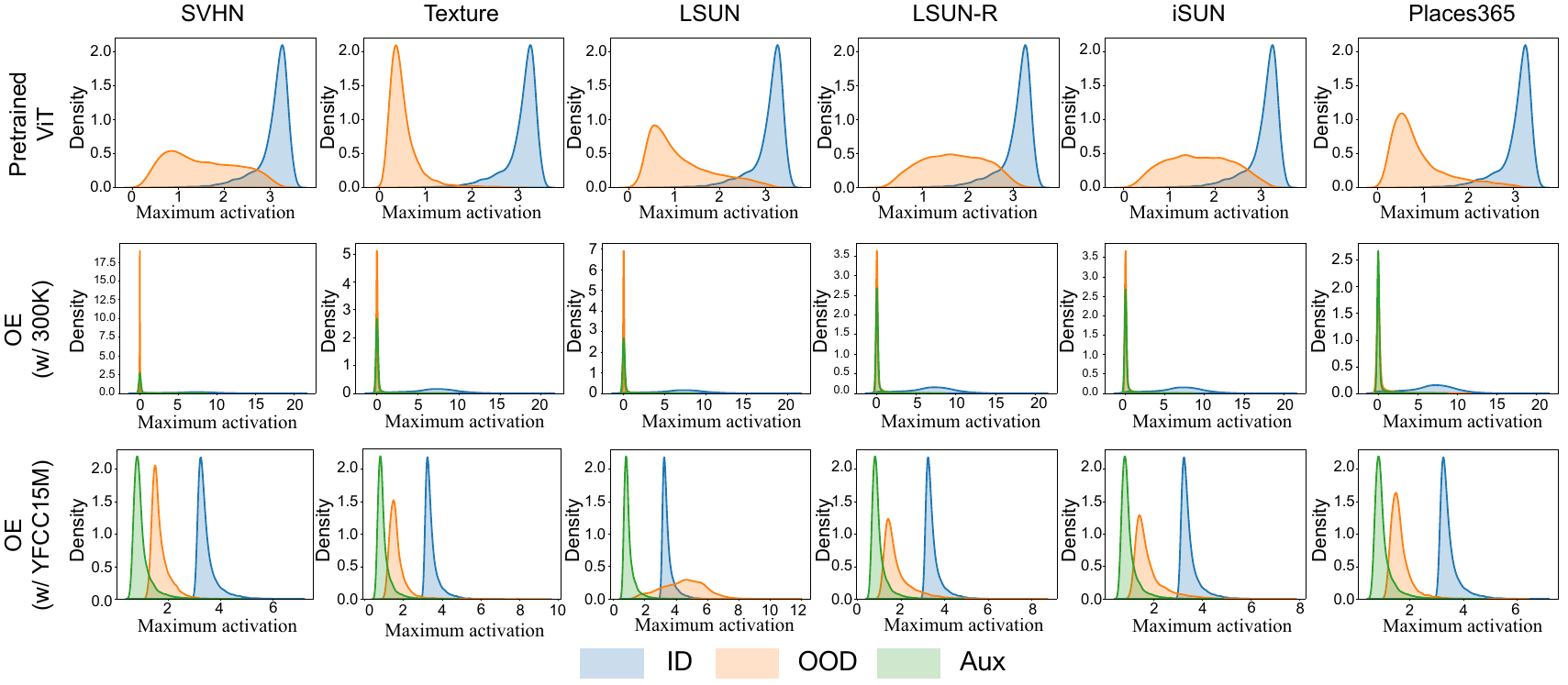}
\caption{The distribution of the maximum activation of the output feature from the last block is investigated for ID training data, OOD testing data, and auxiliary training data on small-scale datasets. When equipped with OE loss, the highly correlated auxiliary data (300K images) can greatly enhance the OOD detection performance. Therefore, careful selection of auxiliary data is crucial.}\label{fig:oe_plot}
\end{figure*}
\begin{figure*}[!ht]
\centering
\includegraphics[width=0.9\linewidth]{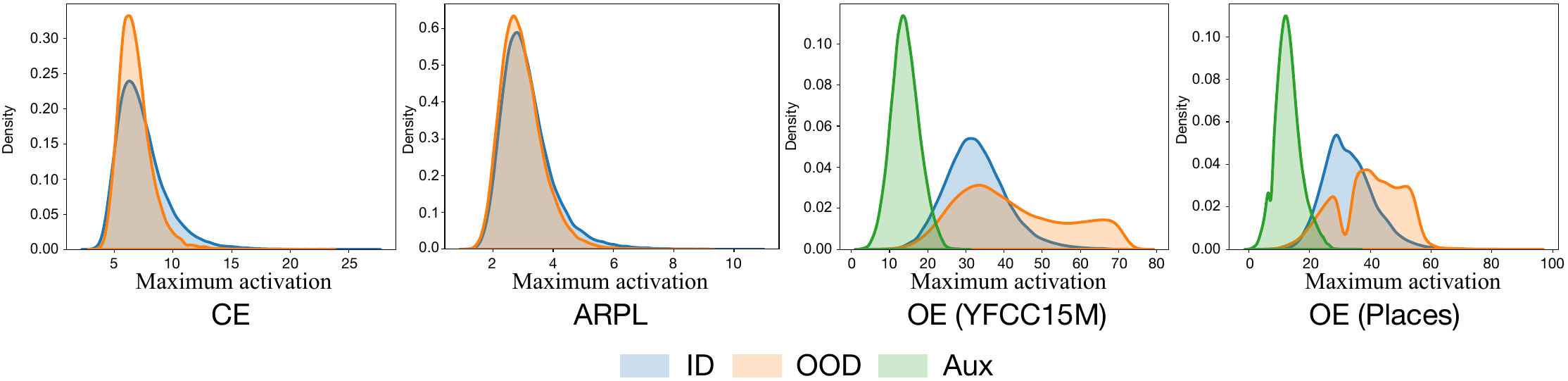}
\caption{
\whj{
Analysis on the distribution of the maximum activation of the output feature from the last layer, for ID training data, OOD testing data, and auxiliary training data on large-scale datasets.
Compared to the 300K images for small-scale datasets, there is less similarity between the auxiliary data (\ie, YFCC15M and Places) and the OOD data. 
This finding aligns with the results in Table~\ref{tab:large_oe}.}
}\label{fig:oe_plot_large}
\end{figure*}

\textbf{Discussion.}
We note that there is no uniquely optimal framing for discussing distribution shift, and here briefly discuss alternate proposals.
For instance,~\cite{zhao22robin} propose a fine-grained analysis of the shifts, where the test time distribution is controlled for specific attributes such as shape and pose.
Also related,~\cite{tran2022plex} discuss that indications of `unfamiliarity' in a neural network could refer to many things, including confusing classes and sub-population shift.
We propose our simple framing as a way to fill the `negative space' left by the semantic shift detection task of OSR.
Furthermore, we suggest it is important to study distribution shift in this way, as classifiers are specifically optimized to differentiate between one set features ($Y_S$) while in fact being invariant to others ($Y_C$).
As such, we would expect models to react differently to changes their distributions.

Finally, we note that for the covariately shifted samples, we ideally wish to develop classifiers which are robust and can perform well despite the presence of distribution shift.
However, given that machine learning models performance degrades under distribution shift, we wish to be able to \textit{detect} when the shift is present. 
To measure whether there is a trade-off between robust models and those which can detect covariate shift, we introduce a new metric in \Cref{sec:outlier_aware_acc}.

\subsection{Quantitative analysis} 
In \Cref{tab:large_bench,tab:large_oe}, we evaluate a selection of previously discussed methods on our large-scale benchmark for both OOD detection and OSR.
Through this large-scale evaluation, we find that \textit{in terms of training methods, among CE, ARPL (+CS), and OE, there \reb{are} no clear winners across the board}. It is surprising that the best performer on the previous small scale benchmarks (see~\Cref{tab:small_scale_main}), OE, appears to struggle when scaled up (last two rows in~\Cref{tab:large_bench}). 
We analyse this contradiction in the next section.
In terms of scoring rules, we again find that \textit{the \reb{magnitude-aware} scoring rules} (MLS and Energy), consistently produce the best performance regardless of the methods and benchmarks (both standard small-scale ones and our large-scale ones). 
\begin{figure*}[!h]
\centering
\includegraphics[width=\linewidth]{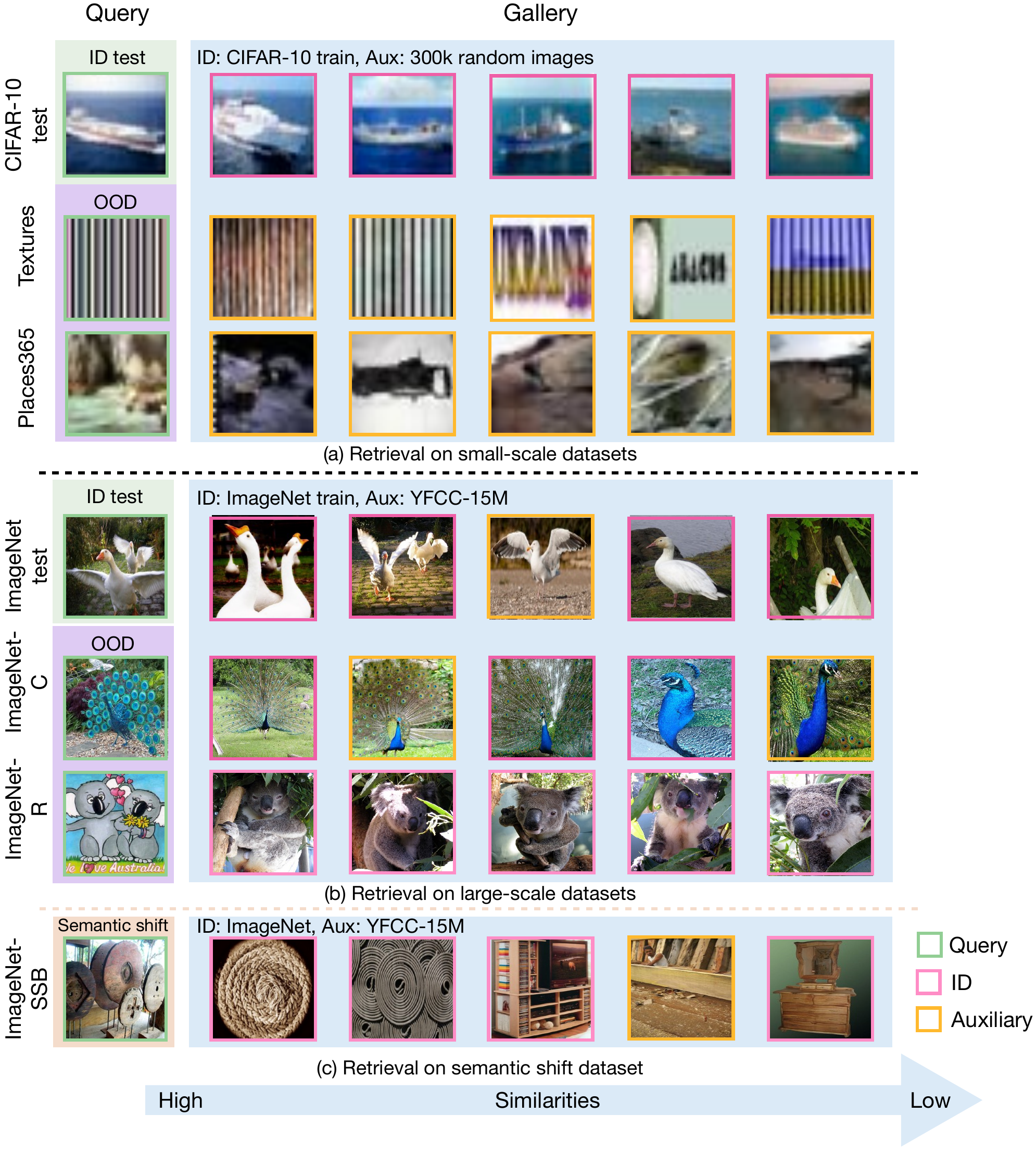}
\caption{\whj{Visualization of nearest neighbors of test samples retrieved from the union of ID and auxiliary training data. We search for the nearest neighbors of samples in both small-scale (\eg, Textures and Places365) and large-scale (\eg, ImageNet-C and ImageNet-R) OOD datasets. 
We also investigate the nearest neighbours for the sample with semantic shift using ImageNet-SSB. It is clear that the improvement in OOD detection performance with OE is closely tied to the similarity between OOD and auxiliary data.}
}\label{fig:oe_auxiliary}
\end{figure*}

\begin{figure*}[t]
\centering
\includegraphics[width=0.8\linewidth]{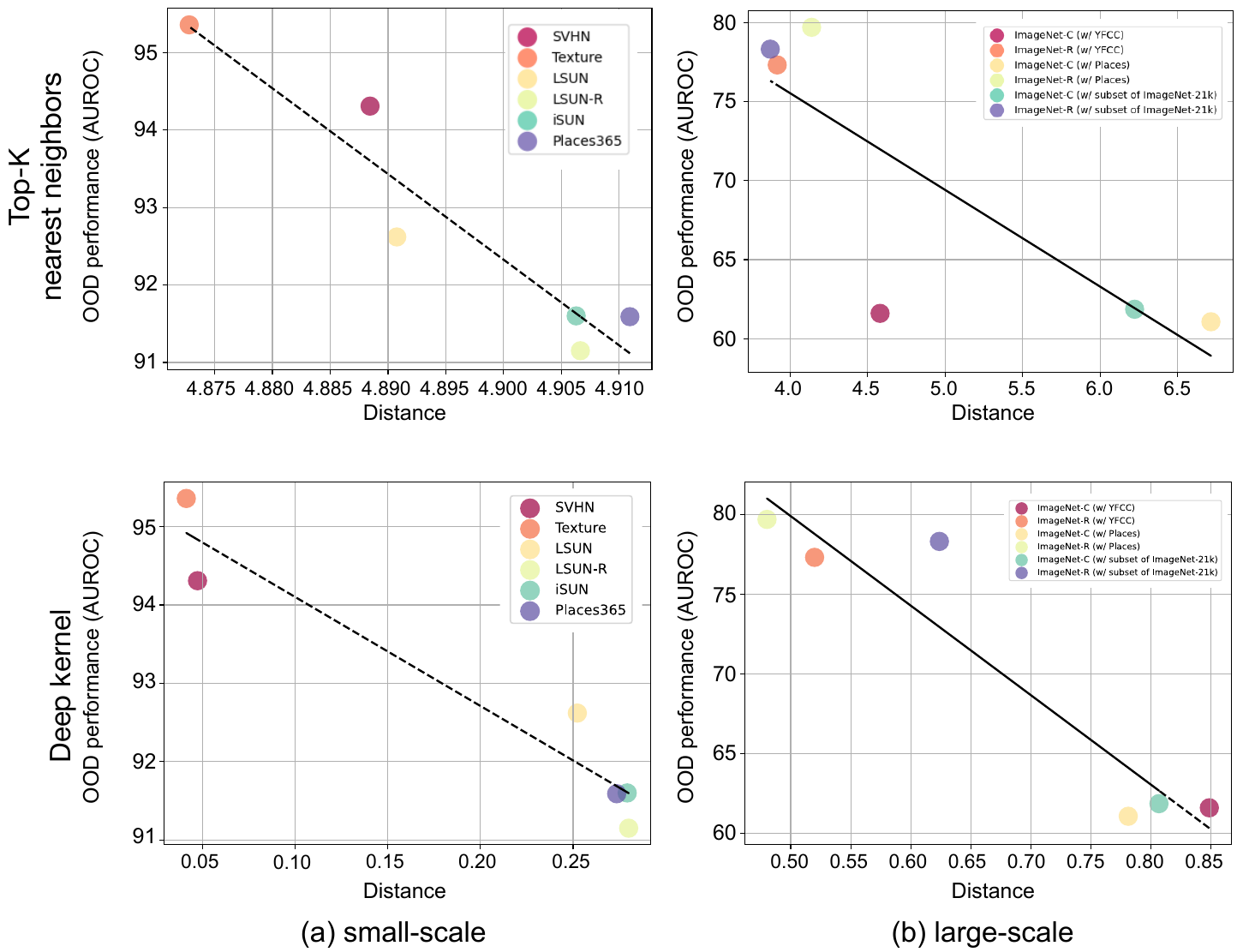}
\caption{\whj{OOD detection performance \vs OOD-AUX data distance of different auxiliary training data for both (a) small-scale and (b) large-scale OOD data. \reb{The OOD detection performance decreases as the distance between OOD data and auxiliary data increases,
for both measurements based on Top-K nearest neighbors and deep kernel distance.}
}}\label{fig:dist_vs_per}
\end{figure*}

\subsection{OE on large-scale datasets}
Here, we investigate why OE performs worse than other methods on a large-scale benchmark. 
One critical difference between OE and other methods is that OE uses auxiliary OOD data for training. 
Intuitively, if the distribution of the auxiliary OOD training data accurately reflects the distribution of the actual OOD testing data, we would expect better detection performance. Otherwise, incomplete or biased outlier data can negatively impact learning.
To analyze this further, we plot the distribution of maximum activation of the output feature (from the last layer) for samples from different data sources: ID data, OOD data, and auxiliary data. The results are shown in~\Cref{fig:oe_plot}. 
It is worth noting that the OOD detection performance strongly (negatively) correlates with the overlapped region of the ID and OOD curve. Additionally, when using 300K random images as auxiliary OOD data (as shown in the $2^{nd}$ row), there is a high correlation with actual OOD data, resulting in excellent performance (see~\Cref{tab:small_scale_main}). We also provide results on a large-scale dataset in~\Cref{fig:oe_plot_large} and further qualitative investigation in~\Cref{apd:D} in Appendix.

\whj{We further retrieve nearest neighbors for the given samples on both small-scale (\eg, Textures and Places365) and large-scale (\eg, ImageNet-C and ImageNet-R) benchmarks using models trained by OE (see~\Cref{fig:oe_auxiliary}). By retrieving the nearest neighbors from the union of ID data and auxiliary data, we observe that for small-scale scenarios, the retrieved nearest neighbors are found in the auxiliary data. However, such phenomenon does not occur consistently in the large-scale datasets. This observation aligns with the correlation between the distance from OOD data to auxiliary data and the OOD detection performance of models trained using OE in~\Cref{tab:large_oe}.}

\subsection{Dataset proximity \emph{vs}. OOD detection performance}
\label{subsec:clip_dist}

To verify that the dataset proximity between auxiliary data and OOD data correlates to the OOD detection performance,
we measure the correlation between OOD detection performance and dataset proximity. We quantify this proximity by calculating the distance between OOD data and auxiliary data. \reb{For a specific OOD dataset, we compute the distance via Top-K nearest neighbors and a deep kernel method~\cite{liu2020learning}, respectively. For Top-K nearest neighbors, we compute the average of all the distances between the normalized feature of each OOD sample and its Top-K nearest neighbors in the auxiliary dataset.}
It can be formulated by:
\begin{equation}
Dist_{nn}(\mathcal{D}^{ood}, \mathcal{D}^{aux})=\frac{\sum_{i=1}^{|\mathcal{D}^{ood}|}\sum_{k=1}^{K}d(Z_i^{ood},Z_k^{aux})}{K|\mathcal{D}^{ood}|},
\end{equation}
where $\mathcal{D}^{ood}$ and $\mathcal{D}^{aux}$ represent the OOD and auxiliary datasets. $Z^{ood}$ and $Z^{aux}$ are the $l_2$-normalized extracted feature from the pretrained model given OOD and auxiliary samples. $d(\cdot,\cdot)$ is the distance measure for nearest neighbors retrieval. 
\reb{We also calculate the deep kernel distance between the OOD and auxiliary datasets following~\cite{liu2020learning}: 
\begin{equation}
Dist_{dk}(\mathcal{D}^{ood}, \mathcal{D}^{aux})=\widehat{\mathrm{MMD}}_u^2\left(\mathcal{D}^{ood}, \mathcal{D}^{aux} ; \phi\right), 
\end{equation}
where $\widehat{\mathrm{MMD}}_u^2(\cdot,\cdot)$ is
the \textit{U}-static estimator
of the maximum mean discrepancy (MMD)~\cite{gretton2012kernel} based on \iid samples from $\mathcal{D}^{ood}$ and $\mathcal{D}^{aux}$, which is considered as an unbiased estimator having nearly minimal variance~\cite{gretton2009fast}. 
$\phi(x, y)=\left[(1-\epsilon) \kappa\left(Z^{ood}, Z^{aux}\right)+\epsilon\right] q(x, y)$, where $\kappa(\cdot, \cdot)$ and $q(\cdot, \cdot)$ are Gaussian kernels. $\epsilon$ is sampled from  $(0, 1)$. The intuition behind the deep kernel method is to effectively distinguish data distributions via representative deep features of samples from each distribution.} We compute the distance (denoted as OOD-AUX data distance) for both small-scale and large-scale OOD data in \Cref{fig:dist_vs_per}, which further validates that closer proximity between auxiliary data and OOD data leads to better performance by OE models.

\subsection{Outlier-aware accuracy}
\label{sec:outlier_aware_acc}

Finally, we introduce a new metric to reconcile the problems of \textit{detecting} covariate shift and being \textit{robust} to it.
\whj{Although AUROC is commonly used to compare different techniques for distinguishing out-of-distribution samples, it does not capture the model's ability to reliably classify testing samples in the presence of distribution shifts.} 
To analyze the relationship of performance between covariate shift and robustness, we introduce a novel measure, which we term \textit{Outlier-Aware Accuracy} (OAA). 
At a given threshold, and a given set of predictions (both ID \vs OOD predictions, and predictions within the closed-set categories), we compute the aggregate frequency of `correct' predictions. The definition of `correct' varies depending on the prediction. Specifically, as shown in~\Cref{fig:robustness_tab}, all instances predicted as ID samples should have accurate class predictions, and all OOD samples that are not already categorized should be detected. This is because we expect a good model to correctly classify all samples with any covariate shift and identify any remaining OOD samples.

The number of the counted instances is then divided by the total number of testing instances to produce the OAA, which is robust to non-semantic shift. This measure is computed under different thresholds based on the scoring rules and aggregated: 
\begin{figure*}[!t]
\centering
\includegraphics[width=\linewidth]{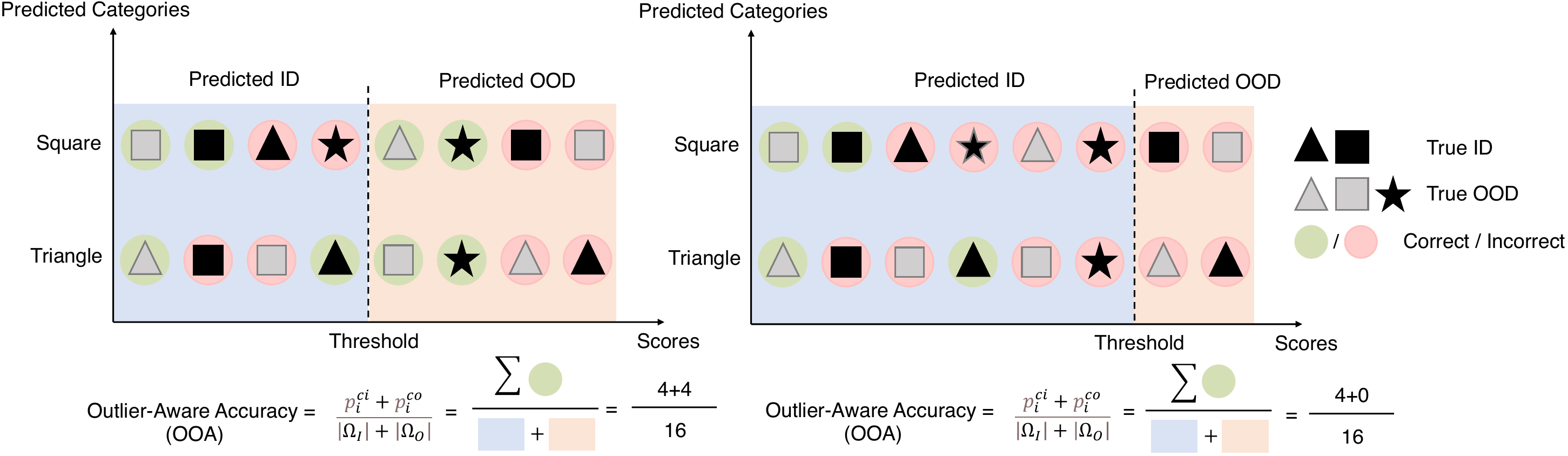}
\caption{Demonstration for Outlier-Aware Accuracy (OAA).
OAA measures the frequency of `correct' predictions made by the model at a given threshold. The `correct' predictions are defined as: (1) Among the testing samples predicted as ID, those whose semantic class labels are \textit{correctly} predicted, denoted as $p^{ci}_i$; (2) True OOD samples that have \textit{incorrect} semantic class predictions, denoted as $p^{co}_i$.
}\label{fig:robustness_tab}
\end{figure*}

\begin{table*}[!h]
\caption{We compute the mean OAA (mOAA) rate across all thresholds to compare different approaches.}\label{tab:metrics}
\centering
\resizebox{0.7\linewidth}{!}{
\begin{tabular}{l|ccccc}
\toprule
Methods & MSP                         & MLS                         & Energy                      & MLS+ReAct                   & Energy+ReAct \\ \midrule
CE      &  {0.661} &  {0.658} &  {0.655} &  {0.655} & 0.641       \\
ARPL    &  {\textbf{0.664}} &  {\textbf{0.660}} &  {\textbf{0.657}} &  {\textbf{0.659}} & \textbf{0.646}       \\ 
OE (w/ YFCC15M)    &  {0.623} &  {0.611} &  {0.611} &  {0.583} & 0.597       \\ \bottomrule
\end{tabular}
}
\end{table*}

\whj{\textbf{mOAA metric.} 
The OAA values across all thresholds can also be aggregated into a single value within $[0, 1]$ as an overall measure, mean OOA (mOAA). To compute the mOAA, we consider testing images including both ID data from $\mathcal{D}^{id}$ and OOD data from $\mathcal{D}^{ood}$. The mOAA is defined as follows:}
\begin{equation}
mOAA=\frac{1}{N}\sum_{i=1}^{N}OAA_i=\frac{1}{N}\sum_{i=1}^{N}\frac{p^{ci}_i+p^{co}_i}{|\mathcal{D}^{id}|+|\mathcal{D}^{ood}|},
\end{equation}
\whj{
where $p^{ci}_i$ denotes the correct predictions among the testing samples predicted as ID, $p^{co}_i$ denotes the correct predictions among the testing samples predicted as OOD (see~\Cref{fig:robustness_tab}), $|\cdot|$ represents the size of the testing set, and $N$ is the number of different thresholds. 
The mOAA score ranges from 0 to 1. A higher value indicates better performance, with a score of 1 representing perfect detection and recognition, while a score of 0 represents the worst performance in terms of separation and recognition.
}

\begin{figure}[!ht]
\centering
\includegraphics[width=\linewidth]{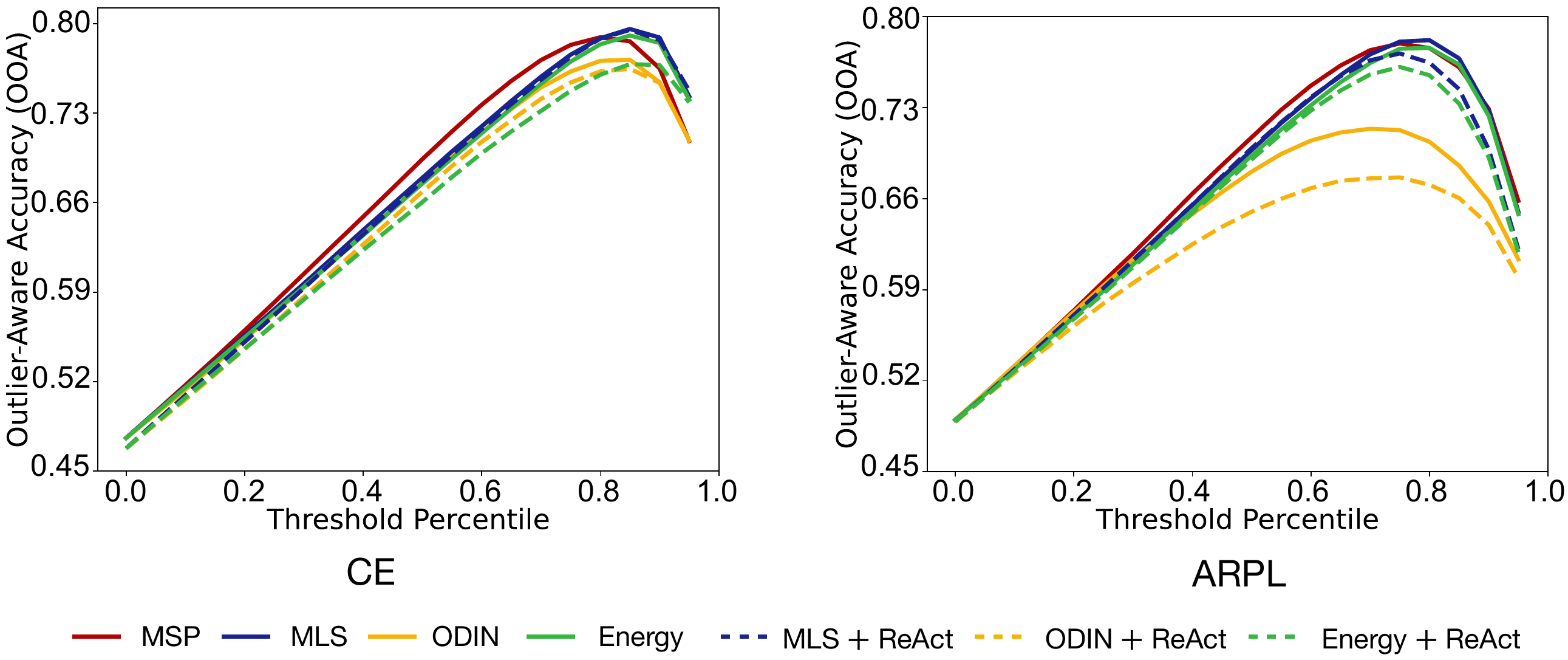}
\caption{Outlier-Aware Accuracy (OAA) of a classifier trained with CE/ARPL loss computed by different thresholds.}\label{fig:metric_plot}
\end{figure}

Based on the numerical results in~\Cref{fig:metric_plot} and \Cref{tab:metrics}, we have observed a turning point for the threshold that achieves the optimal balance between model robustness and OOD detection.

It is worth noting that this metric has a connection to AURC \cite{geifman2019bias}. While both AURC and OAA consider classifier performance, our proposed OAA specifically measures the `correct prediction rate', providing an interpretable value between 0 and 1. Therefore, we believe that this metric can be effectively used to study the tradeoff between OOD detection and generalization with greater precision.

\section{Summary of empirical phenomena}
\label{takeaway}
\whj{In the previous sections, we thoroughly evaluated methods for OOD detection and OSR in terms of scoring rules, training methods, and auxiliary data. To summarize these phenomena, we brieﬂy highlight the key observations as follows:}
(i) Magnitude-aware scoring rules (\ie, MLS and energy) offer obvious advantages for both OOD detection and OSR. Compared with non-magnitude-aware techniques (\ie, ODIN and ReAct), we recommend more stable and deterministic magnitude-aware scoring rules.
(ii) Outlier Exposure (OE) is by far the most effective training method for improving performance on OOD detection and OSR benchmarks when the auxiliary data is highly correlated to the actual OOD data.
However, while finding such auxiliary data is possible for small-scale benchmarks, it is highly non-trivial to find such data in (more realistic) large-scale settings.

\section{Conclusion}
\label{conclusion}
In this study, we explore Out-of-Distribution (OOD) detection and Open-set Recognition (OSR). We conducted a thorough cross-evaluation of methods for OOD detection and OSR. Additionally, we introduced a new benchmark setting that separates the distribution shift problem into \textit{covariate} shift and \textit{semantic} shift, proposing large-scale evaluation protocols for both settings. Our study revealed that the best performing method current OSR and OOD datasets (Outlier Exposure) does not generalize well to our challenging large-scale benchmark. We also discovered that magnitude-aware scoring rules are generally more reliable than others. Overall, our new benchmark can serve as an improved testbed for measuring progress in OSR and OOD detection while providing insights into these two problems. \reb{We hope that our thorough empirical investigation on the OOD detection and OSR methods and benchmarks can shed light for future study and applications on the broader data distribution shift detection problem.}

\section*{Acknowledgments}
This work is supported by Hong Kong Research
Grant Council - Early Career Scheme (Grant No. 27208022); National Natural Science Foundation of
China (Grant No. 62306251); a Facebook AI Research Scholarship; and HKU Seed Fund for Basic Research.

\clearpage

\bibliography{sn-bib}%
\clearpage

\begin{appendices}
\renewcommand\thefigure{A\arabic{figure}}
\renewcommand\thetable{A\arabic{table}}

\section{ }
{\large\textbf{More experimental results on benchmarks}}

\label{apd:A}

We additionally evaluate different methods on Scars-SSB and FGCV-Aircraft-SSB datasets from the Semantic-Shift Benchmark~\cite{vaze2022openset} to further investigate the performance of scoring rules against semantic shifts. From ~\Cref{tab:large_ce} to ~\Cref{tab:large_yfcc}, we can observe that magnitude-aware scoring rules perform well among different training methods.

\begin{table*}[!ht]
\caption{Results of OOD detection and OSR benchmarks on large-scale datasets, using ResNet-50 model trained with the CE loss. The results are averaged from five independent runs.}\label{tab:large_ce}
\centering
\resizebox{\linewidth}{!}{
\begin{tabular}{cl|ccc|ccccccccccc|c}
\toprule
\multicolumn{1}{l|}{\multirow{2}{*}{\begin{tabular}[c]{@{}l@{}}Training \\ Method\end{tabular}}}        & \multicolumn{1}{l|}{\multirow{2}{*}{Scoring Rule}} & \multicolumn{3}{c|}{Covariate Shift}                                                                                                           & \multicolumn{11}{c|}{Semantic Shift}                                                                                                                                                                                                                                                                                                                                                                                                             & \multirow{2}{*}{Overall} \\ \cmidrule(rl){3-16}
\multicolumn{1}{l|}{}                                                                                   & \multicolumn{1}{l|}{}    
                        & \begin{tabular}[c]{@{}c@{}}ImageNet-C\\ ID=63.05\end{tabular} & \begin{tabular}[c]{@{}c@{}}ImageNet-R\\ ID=76.13\end{tabular} & AVG            & \multicolumn{2}{c}{\begin{tabular}[c]{@{}c@{}}ImageNet-SSB\\ (Easy/Hard)\end{tabular}} & \multicolumn{2}{c}{\begin{tabular}[c]{@{}c@{}}CUB\\ (Easy/Hard)\end{tabular}} & \multicolumn{2}{c}{\begin{tabular}[c]{@{}c@{}}Waterbirds\\ (Easy/Hard)\end{tabular}} & \multicolumn{2}{c}{\begin{tabular}[c]{@{}c@{}}Scars\\ (Easy/Hard)\end{tabular}} & \multicolumn{2}{c}{\begin{tabular}[c]{@{}c@{}}FGVC\\ (Easy/Hard)\end{tabular}} & AVG            &                          \\ \midrule

\multicolumn{1}{c|}{\multirow{7}{*}{\begin{tabular}[c]{@{}c@{}}\rotatebox{90}{CE}\end{tabular}}}                     & \multicolumn{1}{l|}{MSP}                  & 64.63                                                         & 80.53                                                         & 72.58          & \underline{80.16}                             & \underline{75.01}                            & 88.11                                 & \textbf{79.43}                        & 81.65                                    & \textbf{75.33}                           & \textbf{94.15}                         & \textbf{82.34}                         & 90.63                         & 82.55                        & \underline{82.94} & 81.21          \\
\multicolumn{1}{c|}{} & \multicolumn{1}{l|}{MLS}                  & \underline{67.92}                                                & \underline{86.71}                                                & \underline{77.32} & \textbf{80.28}                             & \textbf{75.05}                            & 88.29                                 & \underline{79.33}                        & 81.87                                    & \underline{75.18}                                    & \underline{94.03}                         & \underline{82.24}                         & 90.65                         & 82.55                        & \textbf{82.95} & \underline{82.01} \\
\multicolumn{1}{c|}{} &\multicolumn{1}{l|}{ODIN}                 & 63.69                                                         & 85.62                                                         & 74.66          & 74.56                                      & 75.27                                     & 86.24                                 & 73.88                                 & 79.51                                    & 71.54                                    & 92.87                                  & 80.88                                  & \textbf{90.97}                         & 80.97                                 & 80.67 & 79.67          \\
\multicolumn{1}{c|}{} &\multicolumn{1}{l|}{Energy}               & \textbf{68.05}                                                & \textbf{87.04}                                                & \textbf{77.55} & 79.76                             & 74.96                            & \textbf{88.81}                        & 79.06                        & \textbf{82.49}                           & 74.60                                    & 93.92                         & \textbf{82.03}                         & 90.86                         & \textbf{82.82}                        & 82.93 & \textbf{82.03} \\
\multicolumn{1}{c|}{} &\multicolumn{1}{l|}{MLS+ReAct}            & 66.64                                                         & 84.82                                                         & 75.73          & 80.28                                      & 75.07                                     & 88.29                                 & 79.33                                 & 81.69                                    & 75.12                                    & 94.01                                  & 82.23                                  & 90.61                                  & 82.57                                 & 82.92 & 81.72          \\
\multicolumn{1}{c|}{} &\multicolumn{1}{l|}{ODIN+ReAct}           & 61.69                                                         & 83.25                                                         & 72.47          & 74.56                                      & 75.29                                     & 86.24                                 & 73.88                                 & 79.48                                    & 71.50                                    & 92.80                                  & 80.85                                  & \underline{90.88}                                  & 80.92                                 & 80.64 & 79.28          \\
\multicolumn{1}{c|}{} &\multicolumn{1}{l|}{Energy+ReAct}         & 66.88                                                         & 83.92                                                         & 75.40           & 79.76                                      & 74.99                                     & \textbf{88.81}                                 & 79.06                                 & \underline{82.48}                                    & 74.55                                    & 93.89                                  & 82.00                                  & 90.80                                  & \underline{82.79}                                 & 82.91 & 81.66     \\ \bottomrule    
\end{tabular}
}
\end{table*}

\begin{table*}[!ht]
\caption{Results of OOD detection and OSR benchmarks on large-scale datasets, using ResNet-50 model trained with the ARPL loss. The results are averaged from five independent runs.}\label{tab:large_arpl}
\centering
\resizebox{\linewidth}{!}{
\begin{tabular}{cl|ccc|ccccccccccc|c}
\toprule
\multicolumn{1}{l|}{\multirow{2}{*}{\begin{tabular}[c]{@{}l@{}}Training \\ Method\end{tabular}}}        & \multicolumn{1}{l|}{\multirow{2}{*}{Scoring Rule}} & \multicolumn{3}{c|}{Covariate Shift}                                                                                                           & \multicolumn{11}{c|}{Semantic Shift}                                                                                                                                                                                                                                                                                                                                                                                                             & \multirow{2}{*}{Overall} \\ \cmidrule(rl){3-15}
\multicolumn{1}{l|}{}                                                                                   & \multicolumn{1}{l|}{}    
                        & \begin{tabular}[c]{@{}c@{}}ImageNet-C\\ ID=63.05\end{tabular} & \begin{tabular}[c]{@{}c@{}}ImageNet-R\\ ID=76.13\end{tabular} & AVG            & \multicolumn{2}{c}{\begin{tabular}[c]{@{}c@{}}ImageNet-SSB\\ (Easy/Hard)\end{tabular}} & \multicolumn{2}{c}{\begin{tabular}[c]{@{}c@{}}CUB\\ (Easy/Hard)\end{tabular}} & \multicolumn{2}{c}{\begin{tabular}[c]{@{}c@{}}Waterbirds\\ (Easy/Hard)\end{tabular}} & \multicolumn{2}{c}{\begin{tabular}[c]{@{}c@{}}Scars\\ (Easy/Hard)\end{tabular}} & \multicolumn{2}{c}{\begin{tabular}[c]{@{}c@{}}FGVC\\ (Easy/Hard)\end{tabular}} & AVG            &                          \\ \midrule

\multicolumn{1}{c|}{\multirow{7}{*}{\begin{tabular}[c]{@{}c@{}}\rotatebox{90}{ARPL}\end{tabular}}}                     & \multicolumn{1}{l|}{MSP}                  & 61.85                                                         & 78.68                                                         & 70.27          & \underline{79.90}                             & \textbf{74.67}                            & 83.53                        & \textbf{75.64}                        & 79.42                                    & \textbf{72.30}                           & \textbf{94.83}                         & \textbf{83.96}                         & 86.81                         & \textbf{78.01}                        & \textbf{80.91} & 79.13          \\
\multicolumn{1}{c|}{} & \multicolumn{1}{l|}{MLS}                  & \underline{63.94}                                                & \underline{82.77}                                                & \underline{73.36} & \textbf{79.92}                             & \underline{74.60}                            & 83.50                        & \underline{75.49}                        & 79.48                                    & \underline{72.09}                                    & \underline{94.78}                         & 83.63                         & 87.04                         & 77.71                        & 80.82          & \underline{79.58} \\
\multicolumn{1}{c|}{} &\multicolumn{1}{l|}{ODIN}                 & 61.88                                                         & 77.03                                                         & 69.46          & 68.72                                      & 71.23                                     & 73.87                                 & 69.77                                 & 73.76                                    & 69.26                                    & 82.08                                  & 69.10                                  & 70.24                                  & 73.47                                 & 72.15          & 71.70          \\
\multicolumn{1}{c|}{} &\multicolumn{1}{l|}{Energy}               & \textbf{64.13}                                                & \textbf{83.25}                                                & \textbf{73.69} & 79.87                             & 74.49                            & \textbf{83.70}                        & 75.46                        & \textbf{79.64}                           & 71.86                           & 94.70                         & 83.56                         & \textbf{87.28}                         & \underline{77.74}                        & \underline{80.83}          & \textbf{79.64} \\
\multicolumn{1}{c|}{} &\multicolumn{1}{l|}{MLS+ReAct}            & 62.69                                                         & 80.69                                                         & 71.69          & 79.92                                      & 74.60                                     & 83.44                                 & 75.43                                 & 79.44                                    & 72.07                                    & 94.77                                  & \underline{83.66}                                  & 87.01                                  & 77.69                                 & 80.80          & 79.28          \\
\multicolumn{1}{c|}{} &\multicolumn{1}{l|}{ODIN+ReAct}           & 62.23                                                         & 76.08                                                         & 69.16          & 68.72                                      & 71.23                                     & 67.42                                 & 63.91                                 & 73.75                                    & 69.23                                    & 82.07                                  & 69.09                                  & 70.20                                  & 73.49                                 & 70.91          & 70.62          \\
\multicolumn{1}{c|}{} &\multicolumn{1}{l|}{Energy+ReAct}         & 62.89                                                         & 81.17                                                         & 72.03          & 79.87                                      & 74.49                                     & \textbf{83.70}                                 & 75.41                                 & \underline{79.60}                                    & 71.83                                    & 94.69                                  & 83.56                                  & \underline{87.27}                                  & 77.71                                 & 80.81          & 79.35 \\ \bottomrule    
\end{tabular}
}
\end{table*}

We also investigate the effect of OE using different auxiliary data (\eg, Places and YFCC-15M). As shown in \Cref{tab:large_places} and \Cref{tab:large_yfcc}, we can see that  OE performance heavily depends on the auxiliary training data. 

The degeneration of OE when applied to large-scale datasets drives us to think about the core contribution behind the OE method. To find out the reason, we apply OE to the small-scale datasets with different auxiliary data (\ie, YFCC-15M) in \Cref{tab:small_yfcc}. Compared with results using 300K random images, the one using YFCC-15M cannot exceed the performance of the OE baseline. This indicates that the selection of auxiliary data is essential to the OE method and the success of OE may come from the similarity of auxiliary data distribution and test-time outliers.
\begin{table*}[!ht]
\caption{Results of OOD detection and OSR benchmarks on large-scale datasets, using ResNet-50 model trained with the OE loss combined with auxiliary data from Places. The results are averaged from five independent runs.}\label{tab:large_places}
\centering
\resizebox{\linewidth}{!}{
\begin{tabular}{cl|ccccc|ccccc|c}
\toprule
\multicolumn{1}{l|}{\multirow{2}{*}{\begin{tabular}[c]{@{}l@{}}Training \\ Method\end{tabular}}}        & \multicolumn{1}{l|}{\multirow{2}{*}{Scoring Rule}} & \multicolumn{5}{c|}{Covariate Shift}                                                                                                   & \multicolumn{5}{c|}{Semantic Shift}                                                                                                                                                     & \multirow{2}{*}{Overall} \\ \cmidrule(rl){3-12} \multicolumn{1}{l|}{}                                                                                   & \multicolumn{1}{l|}{}    
                        
                        & ImageNet-C     & ImageNet-R     & \multicolumn{2}{c}{\begin{tabular}[c]{@{}c@{}}Waterbird\\ (Easy/Hard)\end{tabular}} & AVG            & \multicolumn{2}{c}{\begin{tabular}[c]{@{}c@{}}ImageNet-SSB\\ (Easy/Hard)\end{tabular}} & \multicolumn{2}{c}{\begin{tabular}[c]{@{}c@{}}CUB\\ (Easy/Hard)\end{tabular}} & AVG            &                          \\ \midrule
\multicolumn{1}{c|}{\multirow{7}{*}{\begin{tabular}[c]{@{}c@{}}\rotatebox{90}{OE}\end{tabular}}} & \multicolumn{1}{l|}{MSP}                  & 61.02          & 75.30          & \underline{79.11}                           & \textbf{73.88}                           & 72.33          & 82.20                                      & 73.45                                     & 75.91                                 & 69.18                                 & 75.19          & 73.76                    \\
\multicolumn{1}{c|}{} & \multicolumn{1}{l|}{MLS}                  & 61.77          & 80.53          & \textbf{79.31}                           & \textbf{73.88}                           & \textbf{73.87} & \underline{82.42}                                      & \underline{75.58}                            & \textbf{79.16}                        & \textbf{73.83}                        & \textbf{77.75} & \textbf{75.81}           \\
\multicolumn{1}{c|}{} & \multicolumn{1}{l|}{ODIN}                 & 57.74          & \textbf{82.31} & 71.28                                    & 69.30                                    & 70.16          & 81.75                                      & 70.87                                     & 73.71                                 & 66.05                                 & 73.10          & 71.63                    \\
\multicolumn{1}{c|}{} & \multicolumn{1}{l|}{Energy}               & \textbf{64.10} & \underline{81.11}          & 76.39                                    & 70.86                                    & \underline{73.12}          & \textbf{83.47}                             & \textbf{75.61}                            & \underline{78.56}                                 & \underline{73.01}                                 & \underline{77.66} & \underline{75.39}           \\
\multicolumn{1}{c|}{} & \multicolumn{1}{l|}{MLS+ReAct}            & \underline{62.39}          & 79.76          & 77.00                                    & 71.93                                    & 72.77          & 81.23                                      & 73.07                                     & 72.09                                 & 70.16                                 & 74.14          & 73.45                    \\
\multicolumn{1}{c|}{} & \multicolumn{1}{l|}{ODIN+ReAct}           & 58.28          & 77.94          & 69.74                                    & 70.06                                    & 69.01          & 80.27                                      & 70.54                                     & 74.40                                  & 68.00                                 & 73.30          & 71.15                    \\
\multicolumn{1}{c|}{} & \multicolumn{1}{l|}{Energy+ReAct}         & 62.26          & 80.91          & 75.32                                    & 69.71                                    & 72.05          & 82.10                                      & 73.79                                     & 77.30                                 & 70.74                                 & 75.98          & 74.02       \\ \bottomrule             
\end{tabular}
}
\end{table*}

\begin{table*}[!ht]
\caption{Results of OOD detection and OSR benchmarks on large-scale datasets, using ResNet-50 model trained with the OE loss combined with auxiliary data from YFCC-15M. The results are averaged from five independent runs.}\label{tab:large_yfcc}
\centering
\resizebox{\linewidth}{!}{
\begin{tabular}{cl|ccccc|ccccc|c}
\toprule
\multicolumn{1}{l|}{\multirow{2}{*}{\begin{tabular}[c]{@{}l@{}}Training \\ Method\end{tabular}}}        & \multicolumn{1}{l|}{\multirow{2}{*}{Scoring Rule}} & \multicolumn{5}{c|}{Covariate Shift}                                                                                                   & \multicolumn{5}{c|}{Semantic Shift}                                                                                                                                                     & \multirow{2}{*}{Overall} \\ \cmidrule(rl){3-12} \multicolumn{1}{l|}{}                                                                                   & \multicolumn{1}{l|}{}    
                        
                        & ImageNet-C     & ImageNet-R     & \multicolumn{2}{c}{\begin{tabular}[c]{@{}c@{}}Waterbird\\ (Easy/Hard)\end{tabular}} & AVG            & \multicolumn{2}{c}{\begin{tabular}[c]{@{}c@{}}ImageNet-SSB\\ (Easy/Hard)\end{tabular}} & \multicolumn{2}{c}{\begin{tabular}[c]{@{}c@{}}CUB\\ (Easy/Hard)\end{tabular}} & AVG            &                          \\ \midrule
\multicolumn{1}{c|}{\multirow{7}{*}{\begin{tabular}[c]{@{}c@{}}\rotatebox{90}{OE}\end{tabular}}} & \multicolumn{1}{l|}{MSP}                  & 59.02          & 70.01          & 73.67                                    & 68.71                                    & 67.85          & 68.44                                      & 71.60                                     & 71.11                                 & 65.27                                 & 69.11          & 68.48                    \\
\multicolumn{1}{c|}{} & \multicolumn{1}{l|}{MLS}                  & \underline{64.12} & \textbf{82.01} & \textbf{79.72}                           & \textbf{74.08}                           & \textbf{74.98} & \underline{79.37}                                      & 72.55                                     & 75.19                                 & \textbf{70.28}                        & \underline{74.35}          & \underline{74.67}           \\
\multicolumn{1}{c|}{} & \multicolumn{1}{l|}{ODIN}                 & 60.80          & 74.36          & 66.80                                    & 68.94                                    & 67.73          & 72.01                                      & 66.87                                     & 71.48                                 & 67.91                                 & 69.57          & 68.65                    \\
\multicolumn{1}{c|}{} & \multicolumn{1}{l|}{Energy}               & \textbf{64.31} & \underline{81.50}          & 77.95                                    & 72.76                                    & \underline{74.13}          & \textbf{81.50}                             & \textbf{74.33}                            & \textbf{77.78}                        & \underline{70.09}                        & \textbf{75.93} & \textbf{75.03}           \\
\multicolumn{1}{c|}{} & \multicolumn{1}{l|}{MLS+ReAct}            & 60.15          & 79.42          & 76.99                                    & \underline{73.58}                                    & 72.54          & 72.30                                      & 72.74                                     & 73.46                                 & 69.67                                 & 72.04          & 72.29                    \\
\multicolumn{1}{c|}{} & \multicolumn{1}{l|}{ODIN+ReAct}           & 60.98          & 74.05          & 66.10                                    & 68.89                                    & 67.51          & 64.67                                      & 61.79                                     & 72.26                                 & 67.76                                 & 66.62          & 67.06                    \\
\multicolumn{1}{c|}{} & \multicolumn{1}{l|}{Energy+ReAct}         & 61.87          & 79.79          & \underline{78.83}                                    & 71.98                                    & 73.12          & 71.24                                      & \underline{73.26}                                     & \underline{75.78}                                 & 69.85                                 & 72.53          & 72.83       \\ \bottomrule                 
\end{tabular}
}
\end{table*}

\begin{table*}[!ht]
\caption{Results of OOD detection and OSR benchmarks on small-scale datasets, using ResNet-18 model trained with the OE loss combined with auxiliary data from YFCC-15M. The results are averaged from five independent runs.}\label{tab:small_yfcc}
\centering
\resizebox{\linewidth}{!}{
\begin{tabular}{c|lccccccc|ccccc|c}
\toprule
\multirow{2}{*}{\begin{tabular}[c]{@{}l@{}}Training \\ Method\end{tabular}} & \multicolumn{1}{l|}{\multirow{2}{*}{\begin{tabular}[c]{@{}l@{}}Scoring Rule \end{tabular}}} & \multicolumn{7}{c|}{OOD detection benchmarks}                                                                                                                                               & \multicolumn{5}{c|}{OSR benchmarks}                                                                                                                                                                                                                                                            & \multirow{2}{*}{Overall} \\ \cmidrule(rl){3-14} 
\multicolumn{1}{l|}{}                                 & \multicolumn{1}{l|}{}                              & SVHN           & Textures       & LSUN           & LSUN-R         & iSUN           & Places365 & \begin{tabular}[c]{@{}c@{}}AVG\\ ID=95.47\end{tabular} & \begin{tabular}[c]{@{}c@{}}CIFAR10\\ ID=97.52\end{tabular} & \begin{tabular}[c]{@{}c@{}}CIFAR+10\\ ID=97.76\end{tabular} & \begin{tabular}[c]{@{}c@{}}CIFAR+50\\ ID=97.29\end{tabular} & \begin{tabular}[c]{@{}c@{}}TinyImageNet\\ ID=87.30\end{tabular}  & \multicolumn{1}{c|}{AVG}            &                          \\ \midrule
\multicolumn{1}{c|}{\multirow{7}{*}{\begin{tabular}[c]{@{}c@{}}\rotatebox{90}{OE}\end{tabular}}}             & \multicolumn{1}{l|}{MSP}                           & 98.96 & \textbf{99.50}    & \underline{98.17} & 94.44  & 94.50 & \textbf{99.61}     & \underline{97.53}                                                   & \underline{90.17}                                                       & 91.21                                                        & 88.17                                                        & 80.54                                                           & 87.52 & 93.53                    \\
\multicolumn{1}{c|}{}                                 & \multicolumn{1}{l|}{MLS}                           & \underline{98.97} & \textbf{99.50}    & \textbf{98.19} & 94.46  & 94.55 & \textbf{99.61}     & \textbf{97.55}                                                   & \textbf{90.36}                                                       & \underline{93.47}                                                        & 89.25                                                        & \underline{81.44}                                                           & \underline{88.63} & \underline{93.98}           \\
\multicolumn{1}{c|}{}                                 & \multicolumn{1}{l|}{ODIN}                          & \textbf{99.02} & 97.40    & 97.12 & 87.21  & 87.84 & 97.18     & 94.30                                                   & 87.93                                                       & 79.37                                                        & 79.47                                                        & \textbf{81.58}                                                           & 82.09 & 89.41                   \\
\multicolumn{1}{c|}{}                                 & \multicolumn{1}{l|}{Energy}                        & 98.93 & 99.48    & 98.12 & 94.19  & 94.33 & \textbf{99.61}     & 97.44                                                   & 89.90                                                       & \textbf{94.91}                                                        & \underline{90.20}                                                        & 81.43                                                           & \textbf{89.11} & \textbf{94.11}           \\
\multicolumn{1}{c|}{}                                 & \multicolumn{1}{l|}{MLS+ReAct}                     & 98.86 & 99.49    & 97.76 & \textbf{94.69}  & \textbf{94.78} & 99.60     & \underline{97.53}                                                   & 89.77                                                       & 91.99                                                        & 89.92                                                        & 81.32                                                           & 88.25 & 93.82                    \\
\multicolumn{1}{c|}{}                                 & \multicolumn{1}{l|}{ODIN+ReAct}                    & 98.89 & 96.66    & 95.48 & 82.50  & 83.60 & 96.37     & 92.25                                                   & 83.31                                                       & 84.32                                                        & 82.37                                                        & 81.43                                                           & 82.86 & 88.49                    \\
\multicolumn{1}{c|}{}                                 & \multicolumn{1}{l|}{Energy+ReAct}                  & 98.83 & 99.47    & 97.70 & \underline{94.51}  & \underline{94.66} & \textbf{99.61}     & 97.46                                                   & 89.45                                                       & 93.00                                                        & \textbf{90.21}                                                        & 81.40                                                           & 88.52 & 93.88                   \\ \bottomrule
\end{tabular}
}
\end{table*}

\section{ }

{\large\textbf{Activations of OOD and open-set data at different layers}}
\label{apd:B}

\begin{figure*}[h]
\centering
\includegraphics[width=0.9\linewidth]{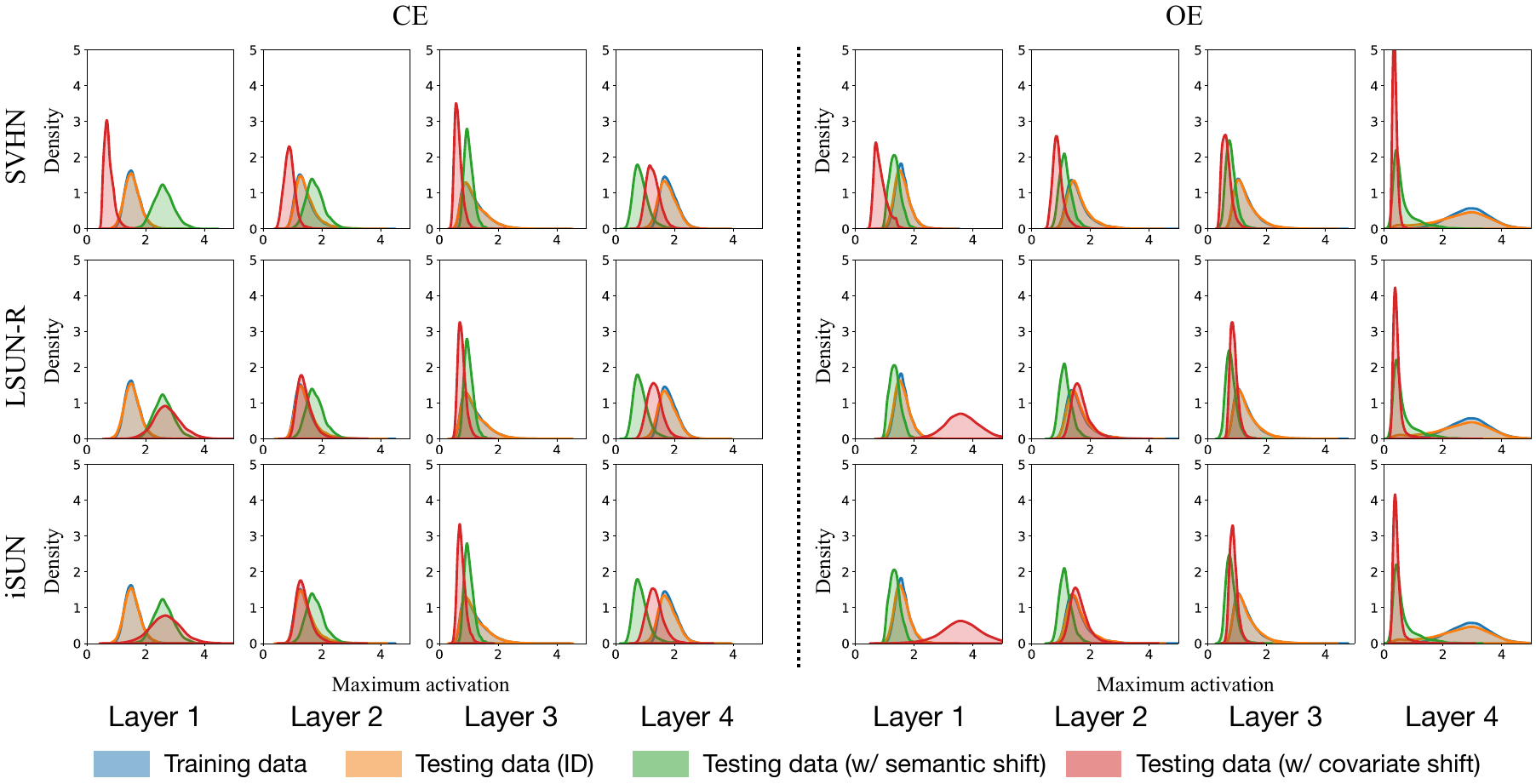}
\caption{Distribution of pre-unit activations for ResNet-18 pretrained on CIFAR10, evaluated on training and ID testing data; open-set data (from CIFAR-100) and OOD data (from SVHN, LSUN, iSUN). Specifically, each subplot shows the maximum activation (along the channel, width and height dimension) at the outputs from \texttt{layer\_1} to \texttt{layer\_4} of the ResNet-18 trained on CIFAR10, displayed from left to right in the figures.}\label{fig:layers_app}
\end{figure*}

We provide the maximum activations for intermediate layers of ResNet-18 trained on CIFAR10 when evaluated on in-distribution data and data with different shifts. Activations in later layers are more discriminative between ID and OOD/open-set samples. After using the OE loss, we can easily notice that the OOD samples are more separable than the model trained with CE loss in~\Cref{fig:second}. We also show the maximum activation of the remaining datasets in~\Cref{fig:layers_app}. 

\begin{figure*}[h]
    \centering
    \includegraphics[width=0.9\linewidth]{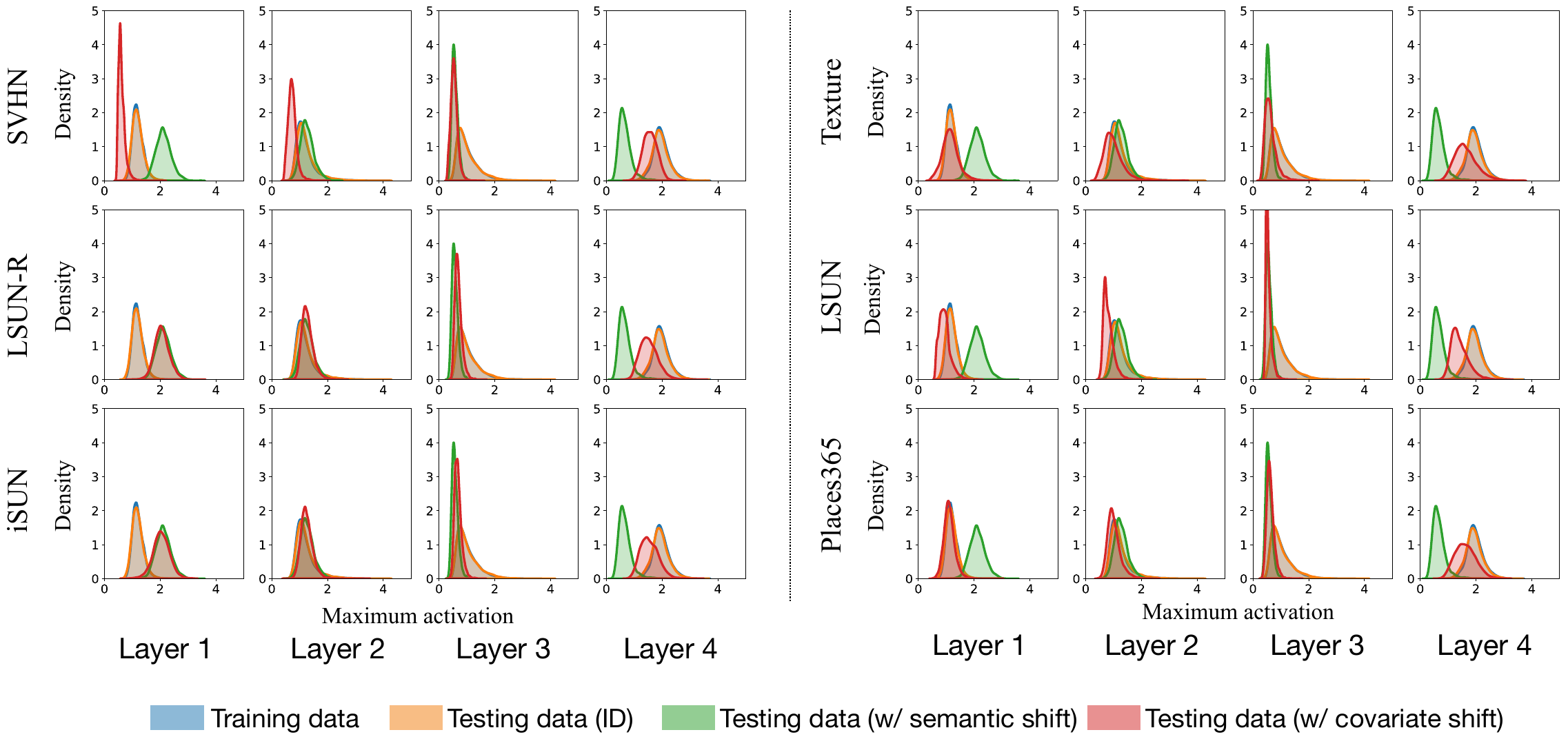}
    \caption{\reb{Histogram of activations for ResNet-18 pretrained using ARPL+CS on a subset of CIFAR10 (with four training classes) and evaluated on: training and testing ID data; open-set data (disjoint six classes in CIFAR10) and OOD data. Specifically, each subplot shows the maximum activation (along channel, width and height dimension) at the outputs from layer 1 to layer 4 of the ResNet-18, displayed from left to right in the figures. Activation maps become notably separable in the last layer between ID and open-set data.}}
    \label{fig:reb_arpl}
\end{figure*}

\reb{We also visualize the histogram of maximum activations of the model trained using ARPL+CS at every layer in~\Cref{fig:reb_arpl}. Our findings align with the observations in Section 3.3 in the main paper, indicating that the early layer activations closely resemble those of the ID test data while the activation patterns begin to differ in the deeper layers. Notably, the ARPL+CS method demonstrates superior separation compared to CE, but it lags behind OE, as illustrated in~\Cref{fig:second} and~\Cref{fig:layers_app}. These findings align with results in~\Cref{tab:small_scale_main}.}

\section{ }
\label{apd:D}
{\large\textbf{Correlation of OOD and auxiliary training data}}

\begin{figure*}[h]
\centering
\includegraphics[width=\linewidth]{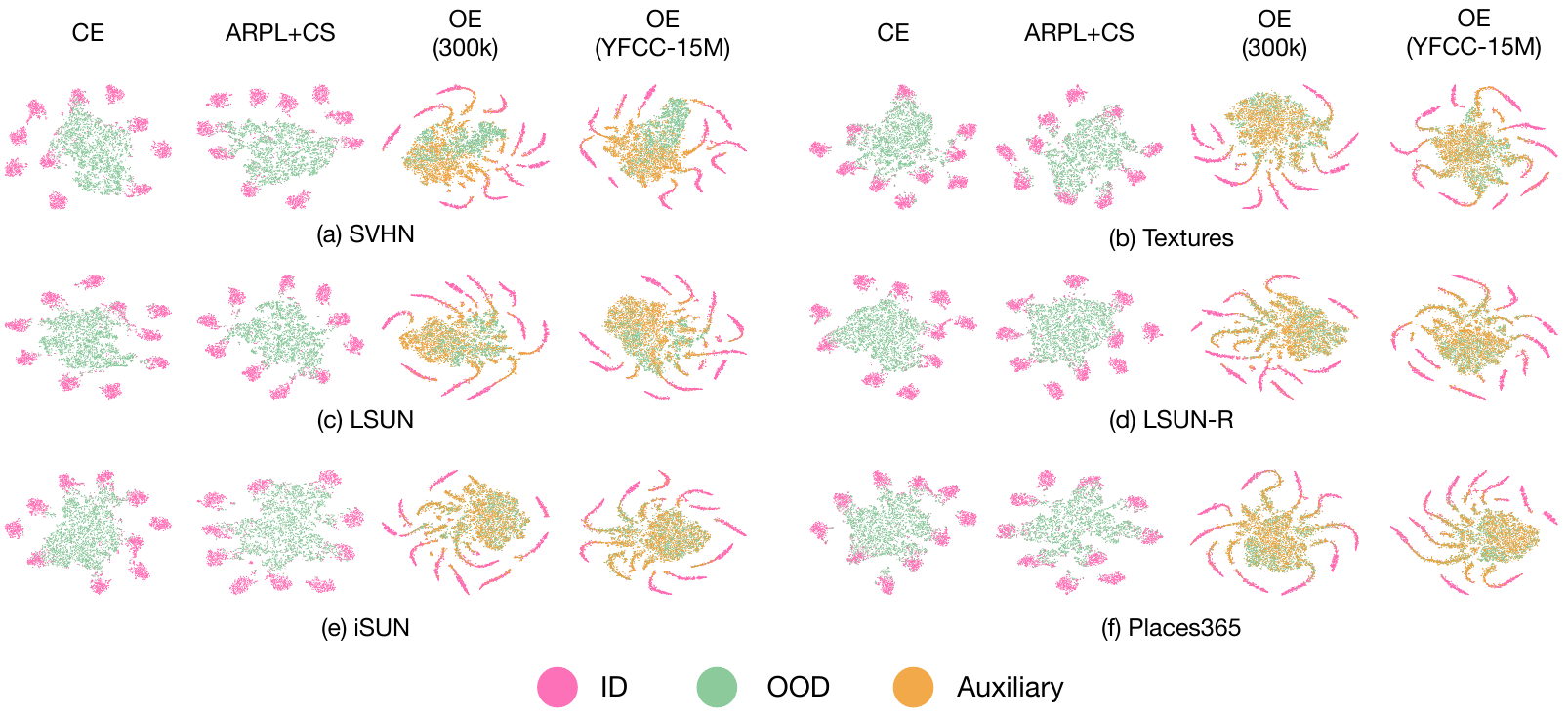}
\caption{t-SNE visualization of representations extracted by models with CE/ARPL+CS/OE loss. Each point denotes a sample and its color denotes which distribution it comes from. The pink/green/brown dots stand for ID/OOD/auxiliary data respectively. 
Together with quantitative results shown in~\Cref{tab:small_bench},
we can observe that the performance boost can be achieved only when the auxiliary data distribution has sufficient overlap with the test-time OOD data distribution (\eg, Texture and Places365).
}\label{fig:tsne}
\end{figure*}

\begin{figure*}[h]
\centering
\includegraphics[width=0.5\linewidth]{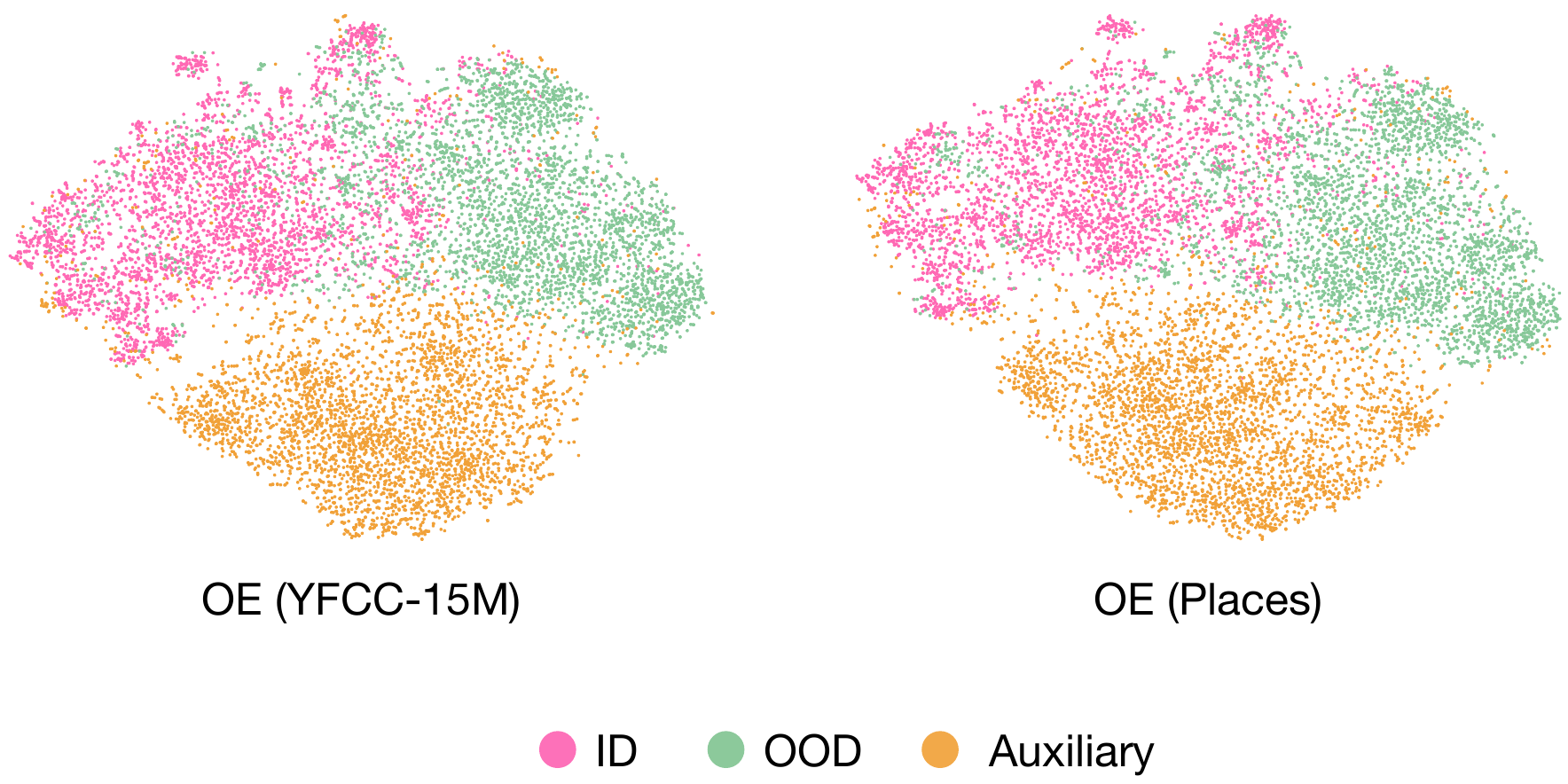}
\caption{t-SNE visualization of representations extracted by models with OE loss using different auxiliary data (\eg, Places and YFCC-15M), tested on ImageNet-R. Each point denotes a sample and its color denotes which distribution it comes from. The pink/green/brown dots stand for ID/OOD/auxiliary data respectively. 
Together with quantitative results shown in~\Cref{tab:large_bench}, the dispersion of OOD and auxiliary data may lead to the unsatisfying performance boost on large-scale datasets.
}\label{fig:tsne_large}
\end{figure*}

In~\Cref{fig:tsne}, we also visualize t-SNE projections of representations for various datasets: ID data (CIFAR10), auxiliary training OOD data (300K~\cite{hendrycks2019oe} \vs YFCC15M), and different test-time OOD datasets. As seen in \Cref{fig:tsne} and~\Cref{fig:tsne_large}, using 300K images generally leads to better overlap with test-time OOD data. Consequently, OE trained with 300K as auxiliary ODD achieves superior performance compared to its counterpart trained with YFCC15M (\Cref{tab:small_bench} \vs \Cref{tab:small_yfcc} in~\Cref{apd:A}) because it shows a better overlap with the test-time OOD data.

\section{ }

{\large\textbf{Different architectures and training setups}}
\label{apd:H}

Apart from ResNet, we also conduct experiments using DenseNet121 on small-scale datasets and using DinoViT-S/8 on large-scale datasets. As shown in~\Cref{tab:arch1} and~\Cref{tab:arch2}, magnitude-aware approaches still perform others on several datasets.

\begin{table*}[!ht]
\caption{\reb{Evaluation on small-scale OOD detection and OSR benchmarks with various methods, using CIFAR10 as ID. The results are averaged from five independent runs. We report the in-distribution accuracy as `ID' and denote intractable results as `-', resulting from unaffordable computational cost. Bold values represent
the best results, while underlined values represent the second best results. Different methods have their optimal scope but MLS and Energy demonstrate stability and models trained with OE dominate on almost all OOD datasets.}}\label{tab:arch1}
\centering
\resizebox{\linewidth}{!}{
\begin{tabular}{c|lccccccccccccc}
\multicolumn{15}{c}{\large{(a) Evaluation based on DenseNet-121 trained with the CE loss.} }                                                                                                                                                                                                                     \\
\multicolumn{15}{c}{}                                                                                                                                                                                                                                                                       \\ \toprule
\multirow{2}{*}{\begin{tabular}[c]{@{}l@{}}Training \\ Method\end{tabular}} & \multicolumn{1}{l|}{\multirow{2}{*}{\begin{tabular}[c]{@{}l@{}}Scoring Rule\end{tabular}}} & \multicolumn{7}{c|}{OOD detection benchmarks}                                                                                                                                               & \multicolumn{5}{c|}{OSR benchmarks}                                                                                                                                                                                                                                                            & \multirow{2}{*}{Overall} \\ \cmidrule(rl){3-14} 
\multicolumn{1}{l|}{}                                 & \multicolumn{1}{l|}{}                              & SVHN           & Textures       & LSUN           & LSUN-R         & iSUN           & Places365      & \multicolumn{1}{c|}{\begin{tabular}[c]{@{}c@{}}AVG\\ ID=95.43\end{tabular}} & \begin{tabular}[c]{@{}c@{}}CIFAR10\\ ID=97.04\end{tabular} & \begin{tabular}[c]{@{}c@{}}CIFAR+10\\ ID=96.73\end{tabular}  & \begin{tabular}[c]{@{}c@{}}CIFAR+50\\ ID=96.91\end{tabular}  & \begin{tabular}[c]{@{}c@{}}TinyImageNet\\ ID=83.52\end{tabular}  & \multicolumn{1}{c|}{AVG}            &                          \\ \midrule
\multicolumn{1}{c|}{\multirow{11}{*}{\begin{tabular}[c]{@{}l@{}}\rotatebox{90}{CE}\end{tabular}}}              & \multicolumn{1}{l|}{MSP}                           & 95.17          & 91.56          & 93.39          & 94.61          & 94.40          & 89.39          & \multicolumn{1}{c|}{93.09}                                                  & 91.63                                                      & 93.77                                                       & 90.47                                                       & 79.76                                                           & \multicolumn{1}{c|}{88.91}          & 91.42                    \\
\multicolumn{1}{c|}{}                                 & \multicolumn{1}{l|}{MLS}                           & \underline{97.05}          & \underline{91.85}          & \underline{94.53}          & 96.11          & \underline{95.84}          & \underline{89.46}          & \multicolumn{1}{c|}{\underline{94.14}}                                                  & \underline{92.63}                                             & \underline{95.70}                                              & \underline{91.88}                                              & \underline{81.52}                                                           & \multicolumn{1}{c|}{\textbf{93.59}} & \textbf{93.92}           \\
\multicolumn{1}{c|}{}                                 & \multicolumn{1}{l|}{ODIN}                          & 95.01          & 84.09          & 86.57          & \textbf{96.31}          & 95.70          & 77.41          & \multicolumn{1}{c|}{89.18}                                                  & 87.62                                                      & 79.42                                                       & 83.57                                                       & 78.92                                                           & \multicolumn{1}{c|}{82.38}          & 86.46                    \\
\multicolumn{1}{c|}{}                                 & \multicolumn{1}{l|}{GODIN}                         & 96.31 & 90.85 & 94.04 & 96.23 & 95.37 & 88.99 & \multicolumn{1}{c|}{93.63}                                         & 90.30                                                      & 91.34                                                       & 87.76                                                       & 76.23                                                           & \multicolumn{1}{c|}{86.41}          & 90.74                    \\
\multicolumn{1}{c|}{}                                 & \multicolumn{1}{l|}{SEM}                           & 75.65          & 72.02          & 75.18          & 70.93          & 72.52          & 76.14          & \multicolumn{1}{c|}{73.74}                                                  & 51.26                                                      & 47.60                                                       & 45.36                                                       & -                                                               & \multicolumn{1}{c|}{48.07}          & 58.67                    \\
\multicolumn{1}{c|}{}                                 & \multicolumn{1}{l|}{Energy}                        & \textbf{97.25}          & \textbf{91.97}          & \textbf{94.68}          & \underline{96.28}          & \textbf{96.02}          & \textbf{89.57}          & \multicolumn{1}{c|}{\textbf{94.30}}                                                  & \textbf{92.73}                                             & \textbf{95.89}                                              & \textbf{92.01}                                              & 81.04                                                           & \multicolumn{1}{c|}{\textbf{90.42}} & \underline{92.75}           \\
\multicolumn{1}{c|}{}                                 & \multicolumn{1}{l|}{MLS+ReAct}                     & 78.39          & 78.31          & 89.42          & 91.55          & 91.20         & 85.23          & \multicolumn{1}{c|}{85.68}                                                  & 92.40                                            & 94.36                                                       & 90.28                                                       & 81.04                                                  & \multicolumn{1}{c|}{89.52}          & 87.22                    \\
\multicolumn{1}{c|}{}                                 & \multicolumn{1}{l|}{ODIN+ReAct}                    & 83.47          & 78.74          & 89.23          & 94.05          & 93.79          & 83.64          & \multicolumn{1}{c|}{87.15}                                                  & 83.27                                                      & 84.25                                                       & 80.62                                                       & 78.27                                                           & \multicolumn{1}{c|}{82.27}          & 85.20                    \\
\multicolumn{1}{c|}{}                                 & \multicolumn{1}{l|}{Energy+ReAct}                  & 85.55          & 82.68          & 87.81          & 90.92          & 89.61          & 77.07          & \multicolumn{1}{c|}{85.61}                                                  & \underline{92.63}                                                      & 95.28                                                       & 91.03                                                       & \textbf{81.53}                                                  & \multicolumn{1}{c|}{90.12}          & 87.41                    \\ 
\multicolumn{1}{c|}{}                                 & \multicolumn{1}{l|}{MLS+ASH}                    & 97.01          & 91.40          & 93.72          & 95.56          & 90.39          & 86.30          & \multicolumn{1}{c|}{92.40}                                                  & 92.21                                                      & 94.35                                                       & 90.56                                                       & 80.88                                                           & \multicolumn{1}{c|}{89.50}          & 91.24                    \\
\multicolumn{1}{c|}{}                                 & \multicolumn{1}{l|}{MLS+SHE}                    & 89.80          & 75.02          & 91.62          & 87.11          & 87.40          & 81.76          & \multicolumn{1}{c|}{85.45}                                                  & 78.80                                                      & 81.92                                                       & 77.27                                                       & 79.92                                                           & \multicolumn{1}{c|}{79.48}          & 83.06                    \\\bottomrule
\multicolumn{15}{c}{}                                                                                                                                                                                                                                                                                                                                                                                                                                                                                                                                                                                                      \\
\multicolumn{15}{c}{\large{(b) Evaluation based on DenseNet-121 trained with the ARPL+CS loss.} }                                                                                                                                                                                                                                                                                                                                                                  \\
\multicolumn{15}{l}{}                                                                                                                                                                                                                                                                       \\ \toprule
\multirow{2}{*}{\begin{tabular}[c]{@{}l@{}}Training \\ Method\end{tabular}} & \multicolumn{1}{l|}{\multirow{2}{*}{Scoring Rule}} & \multicolumn{7}{c|}{OOD detection benchmarks}                                                                                                                                               & \multicolumn{5}{c|}{OSR benchmarks}                                                                                                                                                                                                                                                            & \multirow{2}{*}{Overall} \\ \cmidrule(rl){3-14}
\multicolumn{1}{l|}{}                                 & \multicolumn{1}{l|}{}                              & SVHN           & Textures       & LSUN           & LSUN-R         & iSUN           & Places365      & \multicolumn{1}{c|}{\begin{tabular}[c]{@{}c@{}}AVG\\ ID=92.11\end{tabular}} & \begin{tabular}[c]{@{}c@{}}CIFAR10\\ ID=97.00\end{tabular} & \begin{tabular}[c]{@{}c@{}}CIFAR+10\\ ID=96.78\end{tabular} & \begin{tabular}[c]{@{}c@{}}CIFAR+50\\ ID=96.43\end{tabular} & \begin{tabular}[c]{@{}c@{}}TinyImageNet\\ ID=86.88\end{tabular} & \multicolumn{1}{c|}{AVG}            &                          \\ \midrule
\multicolumn{1}{c|}{\multirow{11}{*}{\begin{tabular}[c]{@{}l@{}}\rotatebox{90}{ARPL+CS}\end{tabular}}}         & \multicolumn{1}{l|}{MSP}                           & 94.94          & 91.81          & 95.12          & 94.25          & 94.39          & 89.57          & \multicolumn{1}{c|}{93.25}                                                  & 92.60                                                      & 95.77                                                       & 94.15                                                       & \underline{82.71}                                                           & \multicolumn{1}{c|}{91.31}          & 92.47                    \\
\multicolumn{1}{c|}{}                                 & \multicolumn{1}{l|}{MLS}                           & \textbf{98.80} & 90.24 & 94.48 & \underline{96.88} & \underline{97.01} & \textbf{95.41} & \multicolumn{1}{c|}{\underline{95.05}}                                         & \underline{93.28}                                            & 96.65                                                       & 94.80                                                       & \textbf{84.91}                                                  & \multicolumn{1}{c|}{\textbf{92.41}}  & \underline{93.99}           \\
\multicolumn{1}{c|}{}                                 & \multicolumn{1}{l|}{ODIN}                          & 78.02          & 73.46          & 82.30          & 93.26          & 91.40          & 73.77          & \multicolumn{1}{c|}{82.04}                                                  & 67.81                                                      & 80.82                                                       & 73.40                                                       & 72.52                                                           & \multicolumn{1}{c|}{73.64}          & 78.68                    \\
\multicolumn{1}{c|}{}                                 & \multicolumn{1}{l|}{GODIN}                         & 94.83          & 90.86          & \underline{96.01}          & 96.28          & 95.88          & 92.12          & \multicolumn{1}{c|}{94.33}                                                  & 91.11                                                      & 95.32                                                       & 93.12                                                       & 80.89                                                           & \multicolumn{1}{c|}{90.11}          & 92.64                    \\
\multicolumn{1}{c|}{}                                 & \multicolumn{1}{l|}{SEM}                           & 78.30          & 76.35          & 85.52          & 76.88          & 78.03          & 71.93          & \multicolumn{1}{c|}{77.84}                                                  & 41.00                                                      & 40.35                                                       & 42.61                                                       & -                                                               & \multicolumn{1}{c|}{41.32}          & 65.67                    \\
\multicolumn{1}{c|}{}                                 & \multicolumn{1}{l|}{Energy}                        & \underline{97.82} & \underline{92.05}          & \textbf{97.24} & \textbf{97.85} & \textbf{97.72} & \underline{94.12} & \multicolumn{1}{c|}{\textbf{96.15}}                                         & \textbf{93.63}                                             & \textbf{96.87}                                              & \textbf{94.99}                                              & 82.12                                                           & \multicolumn{1}{c|}{\underline{91.90}}          & \textbf{94.45}           \\
\multicolumn{1}{c|}{}                                 & \multicolumn{1}{l|}{MLS+ReAct}                     & 85.27          & \textbf{94.50} & 82.44          & 87.50          & 82.98          & 89.43          & \multicolumn{1}{c|}{87.02}                                                  & 92.12                                                      & 95.33                                                       & 92.91                                                       & 80.50                                                           & \multicolumn{1}{c|}{90.22}          & 88.97                    \\
\multicolumn{1}{c|}{}                                 & \multicolumn{1}{l|}{ODIN+ReAct}                    & 83.29          & 79.08          & 85.12          & 88.62          & 83.47          & 82.05          & \multicolumn{1}{c|}{83.61}                                                  & 68.20                                                      & 73.12                                                       & 70.24                                                       & 55.10                                                           & \multicolumn{1}{c|}{66.67}          & 76.83                    \\
\multicolumn{1}{c|}{}                                 & \multicolumn{1}{l|}{Energy+ReAct}                  & 92.37          & 89.45          & 91.88 & 87.07          & 94.13          & 91.52          & \multicolumn{1}{c|}{91.07}                                         & 92.96                                                      & \underline{96.78}                                                       & \underline{94.91}                                                       & 82.04                                                           & \multicolumn{1}{c|}{91.67}          & 91.27          \\ 
\multicolumn{1}{c|}{}                                 & \multicolumn{1}{l|}{MLS+ASH}                    & 96.75          & 92.01          & 92.87          & 97.58          & 90.79          & 91.80          & \multicolumn{1}{c|}{93.63}                                                  & 93.19                                                      & 94.15                                                       & 93.62                                                       & 81.28                                                           & \multicolumn{1}{c|}{90.56}          & 92.40                    \\
\multicolumn{1}{c|}{}                                 & \multicolumn{1}{l|}{MLS+SHE}                    & 85.75          & 79.92          & 82.80          & 83.80          & 81.27          & 86.27          & \multicolumn{1}{c|}{83.30}                                                  & 77.52                                                      & 74.02                                                       & 75.25                                                       & 76.53                                                           & \multicolumn{1}{c|}{75.83}          & 80.31                    \\\bottomrule
\multicolumn{15}{c}{}                                                                                                                                                                                                                                                                                                                                                                                                                                                                                                                                                                                                      \\
\multicolumn{15}{c}{\large{(c) Evaluation based on DenseNet-121 trained with the OE loss.}  }                                                                                                                                                                                                                                                                                                                                                                  \\
\multicolumn{15}{l}{}                                                                                                                                                                                                                                                                       \\ \toprule
\multirow{2}{*}{\begin{tabular}[c]{@{}l@{}}Training \\ Method\end{tabular}} & \multicolumn{1}{l|}{\multirow{2}{*}{Scoring Rule}} & \multicolumn{7}{c|}{OOD detection benchmarks}                                                                                                                                               & \multicolumn{5}{c|}{OSR benchmarks}                                                                                                                                                                                                                                                            & \multirow{2}{*}{Overall} \\ \cmidrule(rl){3-14} 
\multicolumn{1}{l|}{}                                 & \multicolumn{1}{l|}{}                              & SVHN           & Textures       & LSUN           & LSUN-R         & iSUN           & Places365      & \multicolumn{1}{c|}{\begin{tabular}[c]{@{}c@{}}AVG\\ ID=94.82\end{tabular}} & \begin{tabular}[c]{@{}c@{}}CIFAR10\\ ID=97.83\end{tabular}  & \begin{tabular}[c]{@{}c@{}}CIFAR+10\\ ID=98.40\end{tabular}  & \begin{tabular}[c]{@{}c@{}}CIFAR+50\\ ID=97.95\end{tabular} & \begin{tabular}[c]{@{}c@{}}TinyImageNet\\ ID=83.48\end{tabular}  & \multicolumn{1}{c|}{AVG}            &                          \\ \midrule
\multicolumn{1}{l|}{}                                 & \multicolumn{1}{l|}{MSP}                           & \underline{99.27}          & 98.86          & 99.38          & 98.54          & 98.58          & 98.01          & \multicolumn{1}{c|}{98.77}                                                  & \underline{96.42}                                                      & 99.32                                              & \textbf{98.88}                                              & 78.90                                                           & \multicolumn{1}{c|}{\underline{93.38}}          & \underline{96.61}           \\
\multicolumn{1}{c|}{\multirow{11}{*}{\begin{tabular}[c]{@{}l@{}}\rotatebox{90}{OE}\end{tabular}}}              & \multicolumn{1}{l|}{MLS}                           & \underline{99.27}          & \underline{98.87}          & 99.38          & 98.54          & 98.58          & 98.02          & \multicolumn{1}{c|}{\underline{98.78}}                                                  & 96.40                                                      & \textbf{99.36}                                              & \underline{98.83}                                              & \textbf{80.37}                                                  & \multicolumn{1}{c|}{\textbf{93.74}} & \textbf{96.76}           \\
\multicolumn{1}{c|}{}                                 & \multicolumn{1}{l|}{ODIN}                          & 98.84          & 97.95          & 99.13          & \underline{99.00}          & \underline{98.99}          & 96.45          & \multicolumn{1}{c|}{98.39}                                                  & 94.47                                                      & 93.61                                                       & 94.55                                                       & 76.30                                                           & \multicolumn{1}{c|}{89.73}          & 94.93                    \\
\multicolumn{1}{c|}{}                                 & \multicolumn{1}{l|}{GODIN}                         & 97.01          & 95.11          & 88.98          & 83.36          & 84.57          & 89.39          & \multicolumn{1}{c|}{89.74}                                                  & 93.77                                                      & 92.63                                                       & 92.31                                                       & 78.20                                                           & \multicolumn{1}{c|}{89.23}          & 89.54                    \\
\multicolumn{1}{c|}{}                                 & \multicolumn{1}{l|}{SEM}                           & 97.43          & 96.74          & 98.62          & 96.99          & 97.11          & 94.80          & \multicolumn{1}{c|}{96.95}                                                  & 41.25                                                      & 47.11                                                       & 42.75                                                       & -                                                               & \multicolumn{1}{c|}{43.70}          & 79.20                    \\
\multicolumn{1}{c|}{}                                 & \multicolumn{1}{l|}{Energy}                        & 99.26          & \underline{98.87}          & \underline{99.41}          & 98.52          & 98.56          & \textbf{98.04}          & \multicolumn{1}{c|}{\underline{98.78}}                                                  & 93.32                                                      & \textbf{99.36}                                              & 98.81                                              & 80.21                                                  & \multicolumn{1}{c|}{92.93} & 96.4           \\
\multicolumn{1}{c|}{}                                 & \multicolumn{1}{l|}{Gradnorm}                      & \textbf{99.96} & \textbf{99.72} & \textbf{99.83} & \textbf{99.47} & \textbf{99.42} & \underline{97.94} & \multicolumn{1}{c|}{\textbf{99.39}}                                         & \textbf{96.63}                                             & 99.27                                              & 98.73                                              & 61.53                                                           & \multicolumn{1}{c|}{89.04}          & 95.25                    \\
\multicolumn{1}{c|}{}                                 & \multicolumn{1}{l|}{MLS+ReAct}                     & 85.25          & 87.12          & 89.13          & 88.12          & 88.74          & 83.28          & \multicolumn{1}{c|}{86.94}                                                  & 92.43                                                      & 95.56                                                       & 97.06                                                       & 76.63                                                           & \multicolumn{1}{c|}{90.42}          & 88.33                    \\
\multicolumn{1}{c|}{}                                 & \multicolumn{1}{l|}{ODIN+ReAct}                    & 84.27        & 85.71          & 86.13          & 88.12          & 85.35          & 88.07          & \multicolumn{1}{c|}{86.78}                                                  & 85.83                                                      & 84.21                                                       & 90.46                                                       & 75.25                                                           & \multicolumn{1}{c|}{83.94}          & 85.64                    \\
\multicolumn{1}{c|}{}                                 & \multicolumn{1}{l|}{Energy+ReAct}                  & 81.40          & 87.30          & 83.98          & 82.49          & 84.35          & 83.72          & \multicolumn{1}{c|}{83.87}                                                  & 95.89                                                      & 98.80                                                       & 82.73                                                       & \underline{80.31}                                                           & \multicolumn{1}{c|}{89.43}          & 86.09                    \\ 
\multicolumn{1}{c|}{}                                 & \multicolumn{1}{l|}{MLS+ASH}                    & 99.08          & 98.57          & 98.82          & 98.50          & 98.51          & 97.25          & \multicolumn{1}{c|}{98.46}                                                  & 95.62                                                      & 91.63                                                       & 93.89                                                       & 79.21                                                           & \multicolumn{1}{c|}{90.09}          & 95.11                    \\
\multicolumn{1}{c|}{}                                 & \multicolumn{1}{l|}{MLS+SHE}                    & 98.22          & 95.08          & 97.96          & 91.08          & 91.36          & 90.85          & \multicolumn{1}{c|}{94.09}                                                  & 83.19                                                      & 84.53                                                       & 82.75                                                       & 78.80                                                           & \multicolumn{1}{c|}{82.32}          & 89.38                    \\\bottomrule
\end{tabular}
}
\end{table*}

\begin{table*}[!t]
\caption{Results of various methods on OOD detection benchmarks using CIFAR10 as ID training data. The results are averaged from five independent runs. We train a DenseNet121 with the CE loss and report the in distribution accuracy as `ID'.}\label{tab:arch2}
\centering
\resizebox{\linewidth}{!}{
\begin{tabular}{l|l|ccc|ccc|c}
\toprule
\multirow{2}{*}{\begin{tabular}[c]{@{}l@{}}Training \\ Method\end{tabular}} & \multicolumn{1}{l|}{\multirow{2}{*}{\begin{tabular}[c]{@{}l@{}}Scoring Rule \end{tabular}}} & \multicolumn{3}{c|}{Covariate Shift}                                                                                                           & \multicolumn{3}{c|}{Semantic Shift}                                                             &                \\ \cmidrule(rl){3-8}  \multicolumn{1}{l|}{}                                 & \multicolumn{1}{l|}{}                        & \begin{tabular}[c]{@{}c@{}}ImageNet-C\\ ID=63.05\end{tabular} & \begin{tabular}[c]{@{}c@{}}ImageNet-R\\ ID=76.13\end{tabular} & AVG            & \multicolumn{2}{c|}{\begin{tabular}[c]{@{}c@{}}ImageNet-SSB\\ (Easy/Hard)\end{tabular}} & AVG   & Overall        \\ \midrule
\multicolumn{1}{l|}{\multirow{2}{*}{\begin{tabular}[c]{@{}c@{}}DenseNet121+CE\end{tabular}}} & \multicolumn{1}{c|}{MSP}                               & 64.63                                                         & 80.53                                                         & 72.58          & 80.16                            & \multicolumn{1}{c|}{75.01}                           & 77.59 & 75.08          \\
\multicolumn{1}{c|}{}                                 & \multicolumn{1}{c|}{MLS}                               & \textbf{67.92}                                                & 86.71                                                         & \textbf{77.32} & 80.28                            & \multicolumn{1}{c|}{75.05}                           & 77.67 & 77.49 \\ \midrule
\multicolumn{1}{l|}{\multirow{2}{*}{\begin{tabular}[c]{@{}c@{}}DenseNet121+ARPL\end{tabular}}} & \multicolumn{1}{c|}{MSP}                             & 61.85                                                         & 78.68                                                         & 70.27          & 74.56                            & \multicolumn{1}{c|}{75.27}                           & 74.92 & 72.59          \\
\multicolumn{1}{c|}{}                                 & \multicolumn{1}{c|}{MLS}                            & 63.94                                                         & 82.77                                                         & 73.36          & 79.76                            & \multicolumn{1}{c|}{74.96}                           & 77.36 & 75.36 \\ \midrule
\multicolumn{1}{l|}{\multirow{2}{*}{\begin{tabular}[c]{@{}c@{}}DinoViT-S/8\end{tabular}}} & \multicolumn{1}{c|}{MSP}                              & 59.23                                                         & 82.47                                                         & 70.85          & 80.28                            & \multicolumn{1}{c|}{75.07}                           & 77.68 & 74.26          \\
\multicolumn{1}{c|}{}                                 & \multicolumn{1}{c|}{MLS}                                   & 64.00                                                         & \textbf{89.22}                                                & 76.61          & \textbf{81.24}                   & \multicolumn{1}{c|}{\textbf{75.99}}                  & 78.62 & \textbf{77.61} \\ \bottomrule                      
\end{tabular}
}
\end{table*}

\end{appendices}

\end{document}


\pagestyle{empty} %

\title{Dissecting Out-of-Distribution Detection and Open-Set Recognition: A Critical Analysis of Methods and Benchmarks\\
--Supplementary Material--}
\author{Hongjun~Wang$^1$, Sagar~Vaze$^2$, Kai~Han$^1$\thanks{Corresponding author.} \\  [0.5em]
$^1$The University of Hong Kong
\qquad
$^2$University of Oxford}
\date{} %
\maketitle

\pagestyle{plain} %

\maketitle

{
\hypersetup{linkcolor=black}
\tableofcontents
}
\clearpage

\section{Bar charts for cross-benchmarking results}
In this section, we provide more experimental results for different benchmarks in different ways.
\label{supp:vis_cb}

\begin{figure}[!ht]
\centering
\includegraphics[width=\linewidth]{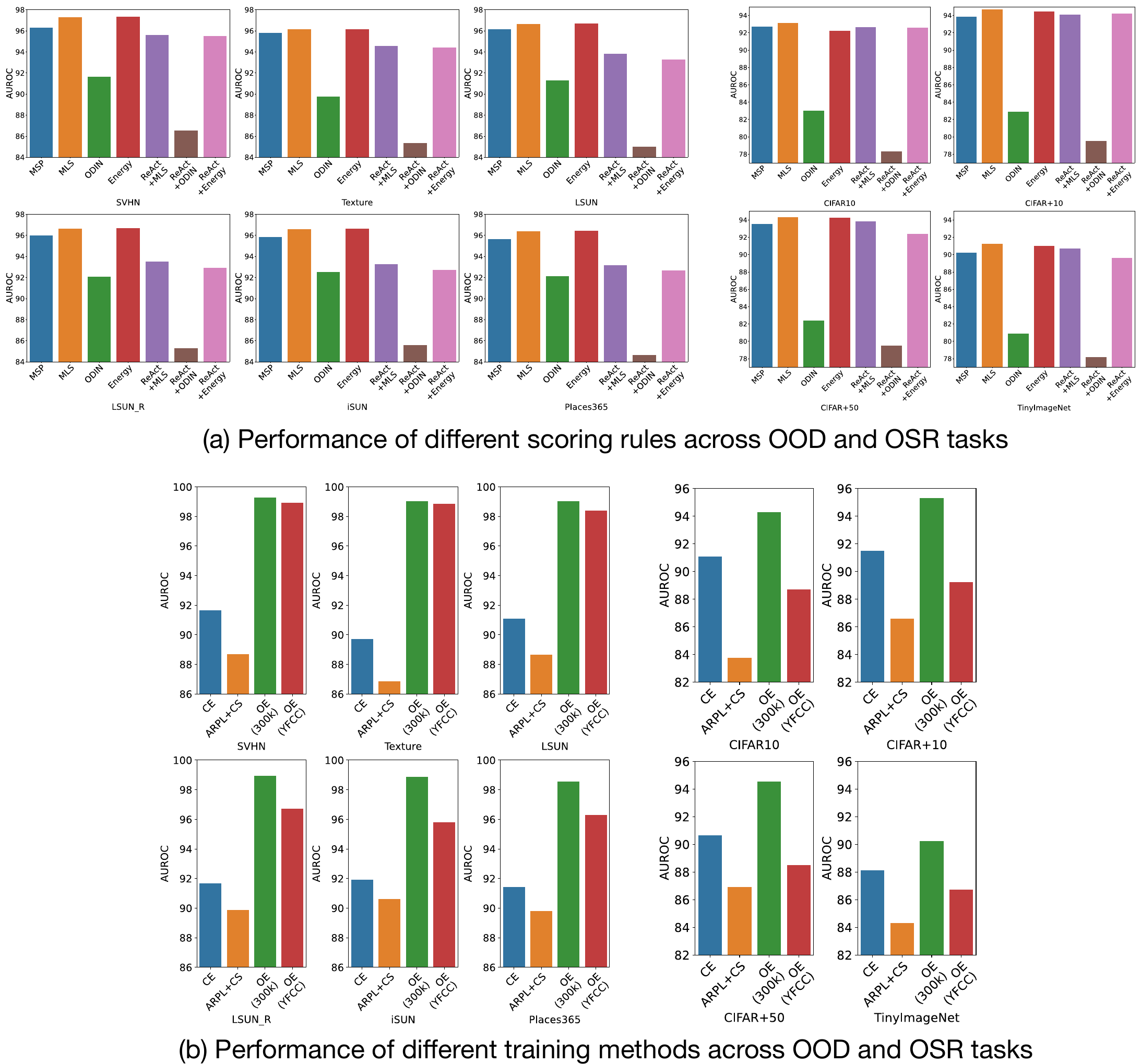}
\caption{Analysis of different scoring rules and training methods on standard benchmarks. (a) Among various scoring rules, MLS and Energy show their reliability across OOD and OSR datasets. (b) For different training methods, Outlier Exposure using different auxiliary data obtains an obvious performance boost compared with others on OOD detection and have slight gains on OSR.}\label{fig:bar}
\end{figure}

Apart from the numerical results in Table 2 in the main paper, we also present the bar charts for the comparison among different (1) scoring rules and (2) training methods in \Cref{fig:bar}. 

In \Cref{fig:bar} (a), we can observe that magnitude-aware methods (\ie, MLS and Energy) are stable among different training methods on both benchmarks while ODIN is encumbered by being combined with ARPL+CS method.

In \Cref{fig:bar} (b), we can see that the OE method with the help of 300K random auxiliary images outperforms others by a large margin. We also notice that the choice of auxiliary data is very important in OE performance since OE with YFCC-15M cannot achieve as excellent performance as the one with 300K random images.

\clearpage

\section{OOD detection using  CIFAR-100 as the ID training data}
\label{supp:cifar100}
To further verify our findings in Section 3.2 in the paper, we also perform experiments using CIFAR-100 as the ID training data. Results are shown in~\Cref{tab:ce_CIFAR-100}, \Cref{tab:arpl_CIFAR-100} and~\Cref{tab:oe_CIFAR-100}. While ODIN outperforms all the methods with the CE loss (due to superior performance on SVHN and LSUN), the magnitude-aware methods consistently perform well (especially when combined with post-processing techniques). This aligns with the findings observed on CIFAR-10.
\begin{table}[h]
\caption{Results of various methods on OOD detection benchmarks, using CIFAR-100 as ID. The results are averaged from five independent runs. We train ResNet-18 with the CE loss and report the in distribution accuracy as `ID'.}\label{tab:ce_CIFAR-100}
\centering
\resizebox{0.73\linewidth}{!}{
\begin{tabular}{c|l|ccccccc}
\toprule
\multirow{2}{*}{\begin{tabular}[c]{@{}l@{}}Training \\ Method\end{tabular}} & \multicolumn{1}{l|}{\multirow{2}{*}{\begin{tabular}[c]{@{}l@{}}Scoring Rule \end{tabular}}}& \multicolumn{7}{c}{OOD detection benchmarks}                                                                                                                                                \\ \cmidrule(rl){3-9} 
\multicolumn{1}{l|}{}                                 & \multicolumn{1}{l|}{} 
                        & SVHN           & Textures       & LSUN           & LSUN-R         & iSUN           & \multicolumn{1}{c|}{Places365}      & \begin{tabular}[c]{@{}c@{}}AVG\\ ID=78.69\end{tabular} \\ \midrule
\multicolumn{1}{c|}{\multirow{10}{*}{\begin{tabular}[c]{@{}c@{}}\rotatebox{90}{CE}\end{tabular}}}             & \multicolumn{1}{l|}{MSP}                  & 83.56          & 78.38          & 78.21          & 79.60          & 79.07          & \multicolumn{1}{c|}{73.56}          & 78.73                                                  \\
\multicolumn{1}{c|}{}                                 & \multicolumn{1}{l|}{MLS}                    & 85.21          & 79.33          & 77.75          & \underline{81.63} & \underline{81.12} & \multicolumn{1}{c|}{73.48}          & 79.75                                                  \\
\multicolumn{1}{c|}{}                                 & \multicolumn{1}{l|}{ODIN}                   & \textbf{97.47} & 77.05          & \textbf{90.49} & 77.59          & 79.38          & \multicolumn{1}{c|}{71.96}          & \textbf{82.32}                                         \\
\multicolumn{1}{c|}{}                                 & \multicolumn{1}{l|}{GODIN}                  & 52.70          & 58.40          & 59.64          & 76.34          & 76.59          & \multicolumn{1}{c|}{62.09}          & 64.29                                                  \\
\multicolumn{1}{c|}{}                                 & \multicolumn{1}{l|}{SEM}                    & 65.54          & 31.85          & 83.58          & 35.81          & 37.38          & \multicolumn{1}{c|}{51.01}          & 50.86                                                  \\
\multicolumn{1}{c|}{}                                 & \multicolumn{1}{l|}{Energy}                 & 85.71          & 79.50          & 77.12          & \textbf{82.19} & \textbf{81.72} & \multicolumn{1}{c|}{73.26}          & 79.92                                                  \\
\multicolumn{1}{c|}{}                                 & \multicolumn{1}{l|}{GradNorm}             & 65.73          & 62.82          & 61.13          & 64.41          & 64.60          & \multicolumn{1}{c|}{61.08}          & 63.30                                                  \\
\multicolumn{1}{c|}{}                                 & \multicolumn{1}{l|}{MLS+ReAct}            & 84.51          & \textbf{83.94} & 84.96          & 80.30          & 79.98          & \multicolumn{1}{c|}{\textbf{78.32}} & \underline{82.00}                                         \\
\multicolumn{1}{c|}{}                                 & \multicolumn{1}{l|}{ODIN+ReAct}           & \underline{96.39}          & 78.74          & \underline{89.62}          & 75.88          & 77.28          & \multicolumn{1}{c|}{71.78}          & 81.62                                                  \\
\multicolumn{1}{c|}{}                                 & \multicolumn{1}{l|}{Energy+ReAct}         & 84.36          & \underline{83.48} & 77.93          & 77.09          & 77.26          & \multicolumn{1}{c|}{\underline{75.55}}          & 79.28           \\ \bottomrule                                      
\end{tabular}
}
\end{table}

\begin{table}[h]
\caption{Results of various methods on OOD benchmarks, using CIFAR-100 as ID. The results are averaged from five independent runs. We train ResNet-18 with the ARPL+CS loss and report the in distribution accuracy as `ID'.}\label{tab:arpl_CIFAR-100}
\centering
\resizebox{0.73\linewidth}{!}{
\begin{tabular}{c|l|ccccccc}
\toprule
\multirow{2}{*}{\begin{tabular}[c]{@{}l@{}}Training \\ Method\end{tabular}} & \multicolumn{1}{l|}{\multirow{2}{*}{\begin{tabular}[c]{@{}l@{}}Scoring Rule \end{tabular}}}& \multicolumn{7}{c}{OOD detection benchmarks}                                                                                                                                                \\ \cmidrule(rl){3-9} 
\multicolumn{1}{l|}{}                                 & \multicolumn{1}{l|}{} 
                        & SVHN           & Textures       & LSUN           & LSUN-R         & iSUN           & \multicolumn{1}{c|}{Places365}      & \begin{tabular}[c]{@{}c@{}}AVG\\ ID=78.86\end{tabular} \\ \midrule
\multicolumn{1}{c|}{\multirow{10}{*}{\begin{tabular}[c]{@{}c@{}}\rotatebox{90}{ARPL+CS}\end{tabular}}}             & \multicolumn{1}{l|}{MSP}                  & 79.95          & 79.00          & 80.28          & 88.16          & 87.71          & \multicolumn{1}{c|}{74.69}          & 81.63                                                  \\
\multicolumn{1}{c|}{}                                 & \multicolumn{1}{l|}{MLS}                    & 81.27          & 79.73          & 79.93          & \underline{89.63}          & \underline{89.23}          & \multicolumn{1}{c|}{\underline{74.77}}          & 82.43                                                 \\
\multicolumn{1}{c|}{}                                 & \multicolumn{1}{l|}{ODIN}                   & \textbf{87.72} & 64.90          & 78.21          & 82.03          & 82.60          & \multicolumn{1}{c|}{65.12}          & 76.76                                         \\
\multicolumn{1}{c|}{}                                 & \multicolumn{1}{l|}{GODIN}                  & 73.81          & 62.55          & 59.30          & 66.42          & 75.00          & \multicolumn{1}{c|}{51.41}          & 64.75                                                  \\
\multicolumn{1}{c|}{}                                 & \multicolumn{1}{l|}{SEM}                    & 55.94          & 33.81          & \textbf{86.74} & 32.23          & 34.34          & \multicolumn{1}{c|}{57.19}          & 50.04                                                  \\
\multicolumn{1}{c|}{}                                 & \multicolumn{1}{l|}{Energy}                 & \underline{81.69}          & \underline{79.84}          & 78.91          & \textbf{90.57} & \textbf{90.21} & \multicolumn{1}{c|}{74.58}          & \underline{82.63}                                                  \\
\multicolumn{1}{c|}{}                                 & \multicolumn{1}{l|}{GradNorm}             & 50.79          & 52.91          & 49.03          & 54.68          & 56.20          & \multicolumn{1}{c|}{50.14}          & 52.29                                                   \\
\multicolumn{1}{c|}{}                                 & \multicolumn{1}{l|}{MLS+ReAct}            & 77.71          & \textbf{80.32} & \underline{85.93}          & 88.44          & 87.78          & \multicolumn{1}{c|}{\textbf{76.41}} & \textbf{82.77}                                         \\
\multicolumn{1}{c|}{}                                 & \multicolumn{1}{l|}{ODIN+ReAct}           & 20.02          & 33.07          & 20.25          & 36.67          & 36.74          & \multicolumn{1}{c|}{37.82}          & 30.76                                                  \\
\multicolumn{1}{c|}{}                                 & \multicolumn{1}{l|}{Energy+ReAct}         & 62.27          & 63.78          & 56.36          & 80.73          & 81.33          & \multicolumn{1}{c|}{60.05}          & 67.42           \\ \bottomrule                                      
\end{tabular}
}
\end{table}

\begin{table}[h]
\caption{Results of various methods on OOD detection benchmarks, using CIFAR-100 as ID. The results are averaged from five independent runs. We train ResNet-18 with the OE loss and report the in distribution accuracy as `ID'.}\label{tab:oe_CIFAR-100}
\centering
\resizebox{0.73\linewidth}{!}{
\begin{tabular}{c|l|ccccccc}
\toprule
\multirow{2}{*}{\begin{tabular}[c]{@{}l@{}}Training \\ Method\end{tabular}} & \multicolumn{1}{l|}{\multirow{2}{*}{\begin{tabular}[c]{@{}l@{}}Scoring Rule \end{tabular}}}& \multicolumn{7}{c}{OOD detection benchmarks}                                                                                                                                                \\ \cmidrule(rl){3-9} 
\multicolumn{1}{l|}{}                                 & \multicolumn{1}{l|}{} 
                        & SVHN           & Textures       & LSUN           & LSUN-R         & iSUN           & \multicolumn{1}{c|}{Places365}      & \begin{tabular}[c]{@{}c@{}}AVG\\ ID=77.16\end{tabular} \\ \midrule
\multicolumn{1}{c|}{\multirow{10}{*}{\begin{tabular}[c]{@{}c@{}}\rotatebox{90}{OE}\end{tabular}}}             & \multicolumn{1}{l|}{MSP}                  & 93.88          & 87.76          & 73.66          & 73.10          & 74.72          & \multicolumn{1}{c|}{\underline{74.49}}          & 79.60                                                   \\
\multicolumn{1}{c|}{}                                 & \multicolumn{1}{l|}{MLS}                    & 94.32          & 88.44          & 72.53          & 73.22          & 75.21          & \multicolumn{1}{c|}{74.10}          & 79.64                                                  \\
\multicolumn{1}{c|}{}                                 & \multicolumn{1}{l|}{ODIN}                   & \textbf{97.20} & 83.94          & \textbf{85.96} & 71.61          & 74.60          & \multicolumn{1}{c|}{71.90}          & 80.87                                         \\
\multicolumn{1}{c|}{}                                 & \multicolumn{1}{l|}{GODIN}                  & 74.18          & 83.35          & 67.85          & 74.71          & 77.22          & \multicolumn{1}{c|}{66.61}          & 73.99                                                  \\
\multicolumn{1}{c|}{}                                 & \multicolumn{1}{l|}{SEM}                    & 68.48          & 47.58          & 80.55          & 48.33          & 46.99          & \multicolumn{1}{c|}{49.15}          & 56.85                                                 \\
\multicolumn{1}{c|}{}                                 & \multicolumn{1}{l|}{Energy}                  & 94.26          & 88.47          & 71.19          & 72.56          & 74.67          & \multicolumn{1}{c|}{73.70}          & 79.14                                                  \\
\multicolumn{1}{c|}{}                                 & \multicolumn{1}{l|}{GradNorm}             & 86.54          & 79.75          & 53.73          & 55.55          & 57.59          & \multicolumn{1}{c|}{63.90}          & 66.18                                                  \\
\multicolumn{1}{c|}{}                                 & \multicolumn{1}{l|}{MLS+ReAct}            & 94.50          & \underline{89.04} & 77.72          & 80.31          & 80.88          & \multicolumn{1}{c|}{\textbf{76.73}} & \textbf{83.20}                                         \\
\multicolumn{1}{c|}{}                                 & \multicolumn{1}{l|}{ODIN+ReAct}           & \underline{96.76}          & 82.66          & \underline{85.65} & \textbf{76.91} & \textbf{78.48} & \multicolumn{1}{c|}{70.50}          & \underline{81.83}                                                   \\
\multicolumn{1}{c|}{}                                 & \multicolumn{1}{l|}{Energy+ReAct}         & 94.07          & \textbf{89.32} & 68.41          & \underline{75.63}          & \underline{77.25}          & \multicolumn{1}{c|}{73.94}          & 79.77          \\ \bottomrule                                      
\end{tabular}
}
\end{table}

\clearpage

\section{Influence of training configurations for OOD detection}
\label{supp:config}
As demonstrated by~\cite{vaze2022openset}, there exists a positive correlation between the closed-set accuracy and open-set performance. Likewise, we would like to verify the correlation between closed-set performance (Accuracy) and OOD detection performance (AUROC). Therefore, we train ResNet-18 with various configurations to investigate the relationship between closed-set accuracy and OOD detection performance~\cite{vaze2022openset}. For the OOD detection task, we use CIFAR10 as ID data and six commonly used datasets as OOD data: SVHN, Textures, LSUN, LSUN, iSUN, and Places365. For the OSR task, we experiment with CIFAR10, CIFAR+10, CIFAR+50, and TinyImageNet following the class split convention in the OSR literature.
(1) For the \emph{ReAct config}, the models are trained with a batch size of 128 for 100 epochs. The initial learning rate is 0.1 and decays by a factor of 10 at epochs 50, 75, and 90. 
(2) For the \emph{MLS config}, we train the models with a batch size of 128 for 600 epochs with the cosine annealed learning rate, restarting the learning rate to the initial value at epochs 200 and 400. Besides, we linearly increase the learning rate from 0 to the initial value at the beginning. The initial learning rate is 0.1 for CIFAR10/100 but 0.01 for TinyImageNet. 
Broadly speaking, we train a ResNet-18 on the ID data, with an SGD optimizer and cosine annealing schedule. 
We train ARPL + CS and OE largely based on the official public implementation.
For the auxiliary outlier dataset in the OE loss, we follow \cite{hendrycks2019oe} and use a subset of 80 Million Tiny Images \cite{torralba200880} with 300K images, removing all examples that appear in CIFAR10/100, Places or LSUN classes.
The results under different configurations are reported in~\Cref{tab:ce_react,tab:arpl_cs_mls,tab:arpl_cs_react,tab:oe_other}.
Comparing~\Cref{tab:arpl_cs_react} and~\Cref{tab:arpl_cs_mls}, we can see that the ID performance in~\Cref{tab:arpl_cs_react} is better than that in~\Cref{tab:arpl_cs_mls} (92.99 vs 91.02). However, the OOD detection performance in~\Cref{tab:arpl_cs_react} is inferior to that in~\Cref{tab:arpl_cs_mls}, indicating that the linear correlation for OSR revealed in~\cite{vaze2022openset} may not hold for the OOD detection task.
\begin{table*}[h]
\centering
\caption{Results on OOD detection and OSR benchmarks, using ResNet-18 trained with
the CE loss, adopting the ReAct config. The results are averaged from five independent runs.
}\label{tab:ce_react}
\resizebox{\linewidth}{!}{
\begin{tabular}{c|l|ccccccc|ccccc|c}
\toprule
\multirow{2}{*}{\begin{tabular}[c]{@{}l@{}}Training \\ Method\end{tabular}} & \multicolumn{1}{l|}{\multirow{2}{*}{\begin{tabular}[c]{@{}l@{}}Scoring Rule \end{tabular}}}& \multicolumn{7}{c|}{OOD detection benchmarks}                                                                                                                           & \multicolumn{5}{c|}{OSR benchmarks}                                                                                                                                                                                                                               &         \multirow{2}{*}{Overall}       \\ \cmidrule(rl){3-14} \multicolumn{1}{l|}{}                                 & \multicolumn{1}{l|}{} 
                        & SVHN           & Textures       & LSUN-C           & LSUN-R         & iSUN           & Places365      & \begin{tabular}[c]{@{}c@{}}AVG\\ ID=92.27\end{tabular} & \begin{tabular}[c]{@{}c@{}}CIFAR10\\ ID=97.13\end{tabular} & \begin{tabular}[c]{@{}c@{}}CIFAR+10\\ ID=96.6\end{tabular} & \begin{tabular}[c]{@{}c@{}}CIFAR+50\\ ID=96.8\end{tabular} & \begin{tabular}[c]{@{}c@{}}TinyImageNet\\ ID=83.4\end{tabular} & AVG  &        \\ \midrule
\multicolumn{1}{c|}{\multirow{7}{*}{\begin{tabular}[c]{@{}c@{}}\rotatebox{90}{CE}\end{tabular}}}             & \multicolumn{1}{l|}{MLS}                  & \underline{85.42} & \textbf{82.20} & 86.56          & 86.40          & 85.40          & 84.62          & 85.1                                                    & 92.54                                              & \underline{95.62}                                              & \underline{91.81}                                              & 81.31                                                           & \underline{90.32} & 87.19          \\
\multicolumn{1}{c|}{}                                 & \multicolumn{1}{l|}{ODIN}                  & 70.62          & 65.32          & 72.49          & 71.68          & 70.43          & 69.10          & 69.94                                                   & 59.77                                                       & 51.37                                                       & 50.22                                                       & 80.96                                                           & 60.58 & 66.20          \\
\multicolumn{1}{c|}{}                                 & \multicolumn{1}{l|}{Energy}               & \textbf{85.45} & \textbf{82.20} & 86.62          & 86.45          & 85.44          & 84.65          & 85.14                                                   & 92.52                                              & \textbf{95.68}                                              & \textbf{91.86}                                              & 81.28                                                           & \textbf{90.34} & 87.22          \\
\multicolumn{1}{c|}{}                                 & \multicolumn{1}{l|}{MLS+ReAct}            & 83.94          & 79.67          & \underline{98.18} & \underline{93.87} & \underline{92.31} & \underline{88.42} & \underline{89.40}                                          & \underline{92.57}                                              & 94.92                                                       & 90.88                                                       & \underline{81.65}                                                  & 90.01 & \textbf{89.64} \\
\multicolumn{1}{c|}{}                                 & \multicolumn{1}{l|}{ODIN+ReAct}           & 83.84          & 79.62          & \textbf{98.31} & \textbf{94.00} & \textbf{92.44} & \textbf{88.48} & \textbf{89.45}                                          & 56.65                                                       & 47.76                                                       & 48.40                                                        & 81.30                                                           & 58.53 & 77.08          \\
\multicolumn{1}{c|}{}                                 & \multicolumn{1}{l|}{Energy+ReAct}         & 83.08          & 76.63          & 95.06          & 94.17          & 92.18          & 85.46          & 87.76                                                   & \textbf{92.58}                                                       & 95.02                                                       & 90.99                                                       & \textbf{81.67}                                                  & 90.07 & \underline{88.14}     \\\bottomrule    
\end{tabular}
}
\end{table*}

\begin{table*}[h]
\centering
\caption{Results on OOD detection and OSR benchmarks, using ResNet-18 trained with the ARPL+CS loss, adopting the ReAct config. The results are averaged from five independent runs.}\label{tab:arpl_cs_react}
\resizebox{\linewidth}{!}{
\begin{tabular}{c|l|ccccccc|ccccc|c}
\toprule
\multirow{2}{*}{\begin{tabular}[c]{@{}l@{}}Training \\ Method\end{tabular}} & \multicolumn{1}{l|}{\multirow{2}{*}{\begin{tabular}[c]{@{}l@{}}Scoring Rule \end{tabular}}}& \multicolumn{7}{c|}{OOD detection benchmarks}                                                                                                                           & \multicolumn{5}{c|}{OSR benchmarks}                                                                                                                                                                                                                               &         \multirow{2}{*}{Overall}       \\ \cmidrule(rl){3-14} \multicolumn{1}{l|}{}                                 & \multicolumn{1}{l|}{} 
                        & SVHN           & Textures       & LSUN-C           & LSUN-R         & iSUN           & Places365      & \begin{tabular}[c]{@{}c@{}}AVG\\ ID=92.99\end{tabular} & \begin{tabular}[c]{@{}c@{}}CIFAR10\\ ID=97.13\end{tabular} & \begin{tabular}[c]{@{}c@{}}CIFAR+10\\ ID=97.75\end{tabular} & \begin{tabular}[c]{@{}c@{}}CIFAR+50\\ ID=97.83\end{tabular} & \begin{tabular}[c]{@{}c@{}}TinyImageNet\\ ID=85.1\end{tabular} & AVG  &        \\ \midrule
\multicolumn{1}{c|}{\multirow{7}{*}{\begin{tabular}[c]{@{}c@{}}\rotatebox{90}{ARPL+CS}\end{tabular}}}             & \multicolumn{1}{l|}{MLS}                  & 79.18          & 86.85          & \underline{95.59} & 95.00          & 94.47          & 89.98          & 90.18                                                   & \textbf{93.23}                                              & 97.93                                               & 96.27                                               & \textbf{82.50}                                                   & \textbf{92.48} & 91.10          \\
\multicolumn{1}{c|}{}                                 & \multicolumn{1}{l|}{ODIN}                  & 50.14          & 38.35          & 74.25          & 42.96          & 43.46          & 53.99          & 50.53                                                   & 92.57                                                       & 97.37                                                        & 95.23                                                        & 53.20                                                           & 84.59          & 64.15          \\
\multicolumn{1}{c|}{}                                 & \multicolumn{1}{l|}{Energy}               & 79.08          & 86.82          & \textbf{95.68} & 95.07          & 94.54          & 90.03          & 90.20                                                   & \underline{93.19}                                              & \underline{97.96}                                               & \textbf{96.30}                                               & \underline{80.45}                                                           & \underline{91.98}          & 90.91          \\
\multicolumn{1}{c|}{}                                 & \multicolumn{1}{l|}{MLS+ReAct}            & \textbf{84.19} & \underline{89.11} & 94.98          & \underline{95.52} & \underline{94.98} & \underline{90.23} & \underline{91.50}                                          & 92.63                                                       & \underline{97.96}                                               & \textbf{96.30}                                               & 79.78                                                           & 91.67          & \underline{91.57} \\
\multicolumn{1}{c|}{}                                 & \multicolumn{1}{l|}{ODIN+ReAct}           & 49.22          & 37.19          & 65.89          & 37.80          & 38.77          & 47.55          & 46.07                                                   & 91.00                                                       & 94.74                                                        & 91.84                                                        & 47.47                                                           & 81.26          & 60.15          \\
\multicolumn{1}{c|}{}                                 & \multicolumn{1}{l|}{Energy+ReAct}         & \textbf{84.13} & \underline{89.13} & 95.10          & \textbf{95.63} & \textbf{95.09} & \textbf{90.32} & \textbf{91.57}                                          & 92.59                                                       & \textbf{98.01}                                               & \textbf{96.33}                                               & 79.90                                                           & 91.71          & \textbf{91.62}    \\\bottomrule    
\end{tabular}
}
\end{table*}

\begin{table*}[h]
\centering
\caption{Results on OOD detection and OSR benchmarks, using ResNet-18 trained with the ARPL+CS loss, adopting the MLS config. The results are averaged from five independent runs.}\label{tab:arpl_cs_mls}
\resizebox{\linewidth}{!}{
\begin{tabular}{c|l|ccccccc|ccccc|c}
\toprule
\multirow{2}{*}{\begin{tabular}[c]{@{}l@{}}Training \\ Method\end{tabular}} & \multicolumn{1}{l|}{\multirow{2}{*}{\begin{tabular}[c]{@{}l@{}}Scoring Rule \end{tabular}}}& \multicolumn{7}{c|}{OOD detection benchmarks}                                                                                                                           & \multicolumn{5}{c|}{OSR benchmarks}                                                                                                                                                                                                                               &         \multirow{2}{*}{Overall}       \\ \cmidrule(rl){3-14} \multicolumn{1}{l|}{}                                 & \multicolumn{1}{l|}{} 
                        & SVHN           & Textures       & LSUN-C           & LSUN-R         & iSUN           & Places365      & \begin{tabular}[c]{@{}c@{}}AVG\\ ID=91.02\end{tabular} & \begin{tabular}[c]{@{}c@{}}CIFAR10\\ ID=96.96\end{tabular} & \begin{tabular}[c]{@{}c@{}}CIFAR+10\\ ID=96.77\end{tabular} & \begin{tabular}[c]{@{}c@{}}CIFAR+50\\ ID=96.69\end{tabular} & \begin{tabular}[c]{@{}c@{}}TinyImageNet\\ ID=86.91\end{tabular} & AVG  &        \\ \midrule
\multicolumn{1}{c|}{\multirow{7}{*}{\begin{tabular}[c]{@{}c@{}}\rotatebox{90}{ARPL+CS}\end{tabular}}}             & \multicolumn{1}{l|}{MLS}                  & \underline{96.36} & 90.20          & \underline{96.59} & \underline{96.95} & \underline{96.88} & \underline{93.29} & 95.05                                                   & \underline{93.16}                                              & 96.58                                                        & 94.67                                                        & \textbf{84.79}                                                   & \textbf{92.30} & \textbf{93.95}          \\
\multicolumn{1}{c|}{}                                 & \multicolumn{1}{l|}{ODIN}                  & 75.92          & 71.64          & 86.25          & 95.14          & 95.19          & 75.97          & 83.35                                                   & 58.04                                                       & 74.80                                                        & 71.52                                                        & 63.13                                                            & 66.87          & 76.76          \\
\multicolumn{1}{c|}{}                                 & \multicolumn{1}{l|}{Energy}               & \textbf{96.52} & 90.11          & \textbf{96.76} & \textbf{97.16} & \textbf{97.07} & \textbf{93.45} & \underline{95.18}                                          & \textbf{93.22}                                              & \textbf{96.74}                                               & \textbf{94.82}                                               & 82.10                                                            & \underline{91.72}          & \underline{93.80}          \\
\multicolumn{1}{c|}{}                                 & \multicolumn{1}{l|}{MLS+ReAct}            & 95.87          & \textbf{92.37} & 96.37          & 96.34          & 96.30          & 92.97          & 95.04                                                   & 92.70                                                       & 96.42                                                        & 94.53                                                        & 82.05                                                            & 91.43          & 93.59 \\
\multicolumn{1}{c|}{}                                 & \multicolumn{1}{l|}{ODIN+ReAct}           & 71.87          & 73.36          & 83.19          & 92.34          & 92.36          & 69.10          & 80.37                                                   & 55.71                                                       & 62.88                                                        & 61.85                                                        & 54.29                                                            & 58.68          & 71.70          \\
\multicolumn{1}{c|}{}                                 & \multicolumn{1}{l|}{Energy+ReAct}         & 96.06          & \underline{92.35}          & \underline{96.59} & 96.58          & 96.53          & 93.17          & \textbf{95.21}                                          & 92.80                                                       & \underline{96.61}                                                        & \underline{94.70}                                                        & \underline{82.14}                                                            & 91.56          & 93.75    \\\bottomrule    
\end{tabular}
}
\end{table*}

\begin{table*}[ht]
\centering
\caption{Results on OOD detection and OSR benchmarks, using ResNet-18 trained with the OE loss, adopting the official configuration of OE. The results are averaged from five independent runs.}\label{tab:oe_other}
\resizebox{\linewidth}{!}{
\begin{tabular}{c|l|ccccccc|ccccc|c}
\toprule
\multirow{2}{*}{\begin{tabular}[c]{@{}l@{}}Training \\ Method\end{tabular}} & \multicolumn{1}{l|}{\multirow{2}{*}{\begin{tabular}[c]{@{}l@{}}Scoring Rule \end{tabular}}}& \multicolumn{7}{c|}{OOD detection benchmarks}                                                                                                                           & \multicolumn{5}{c|}{OSR benchmarks}                                                                                                                                                                                                                               &         \multirow{2}{*}{Overall}       \\ \cmidrule(rl){3-14} \multicolumn{1}{l|}{}                                 & \multicolumn{1}{l|}{} 
                        & SVHN           & Textures       & LSUN-C           & LSUN-R         & iSUN           & Places365       & \begin{tabular}[c]{@{}c@{}}AVG\\ ID=94.77\end{tabular} & \begin{tabular}[c]{@{}c@{}}CIFAR10\\ ID=97.59\end{tabular} & \begin{tabular}[c]{@{}c@{}}CIFAR+10\\ ID=97.38\end{tabular} & \begin{tabular}[c]{@{}c@{}}CIFAR+50\\ ID=97.34\end{tabular} & \begin{tabular}[c]{@{}c@{}}TinyImageNet\\ ID=77.96\end{tabular} & AVG  &        \\ \midrule
\multicolumn{1}{c|}{\multirow{7}{*}{\begin{tabular}[c]{@{}c@{}}\rotatebox{90}{OE}\end{tabular}}}             & \multicolumn{1}{l|}{MLS}                  & \textbf{95.94} & \textbf{94.57} & 76.58          & 85.20          & 87.16          & 88.46          & 87.99                                                   & \underline{96.26}                                              & \textbf{98.95}                                               & \textbf{98.20}                                               & \textbf{77.88}                                                   & \textbf{92.82} & 89.92          \\
\multicolumn{1}{c|}{}                                 & \multicolumn{1}{l|}{ODIN}                  & 93.32          & 92.84          & 65.85          & 84.50          & 87.24          & 82.29          & 84.34                                                   & 93.44                                                       & 96.18                                                        & 93.69                                                        & 77.49                                                            & 90.20          & 86.68          \\
\multicolumn{1}{c|}{}                                 & \multicolumn{1}{l|}{Energy}               & \underline{95.84} & \underline{94.45} & 75.50          & 84.55          & 86.64          & 88.19          & 87.54                                                   & \textbf{96.33}                                              & \textbf{98.95}                                               & 98.04                                                        & \underline{77.73}                                                            & \underline{92.76} & 89.62          \\
\multicolumn{1}{c|}{}                                 & \multicolumn{1}{l|}{MLS+ReAct}            & 95.46          & 94.32          & \textbf{93.57}          & \textbf{90.66}          & \underline{90.75}          & \underline{88.97}          & \textbf{92.29}                                          & 96.20                                              & 98.93                                               & \underline{98.18}                                               & 77.60                                                            & 92.73 & \textbf{92.46} \\
\multicolumn{1}{c|}{}                                 & \multicolumn{1}{l|}{ODIN+ReAct}           & 87.64          & 88.52          & \underline{89.56} & 86.19          & 86.57          & 76.02          & 85.75                                                   & 93.03                                                       & 95.45                                                        & 92.75                                                        & 76.90                                                            & 89.53          & 87.26          \\
\multicolumn{1}{c|}{}                                 & \multicolumn{1}{l|}{Energy+ReAct}         & 87.84          & 89.29          & 88.66          & \underline{90.14} & \textbf{93.69} & \textbf{94.67} & \underline{90.72}                                                   & 91.04                                                       & 98.93                                                        & 98.02                                                        & 77.48                                                            & 91.37          & \underline{90.98} \\  \bottomrule
\end{tabular}
}
\end{table*}

\section{Hyperparameters investigation for ReAct}
\label{supp:react}
\begin{figure*}[h]
    \centering
    \includegraphics[width=0.65\linewidth]{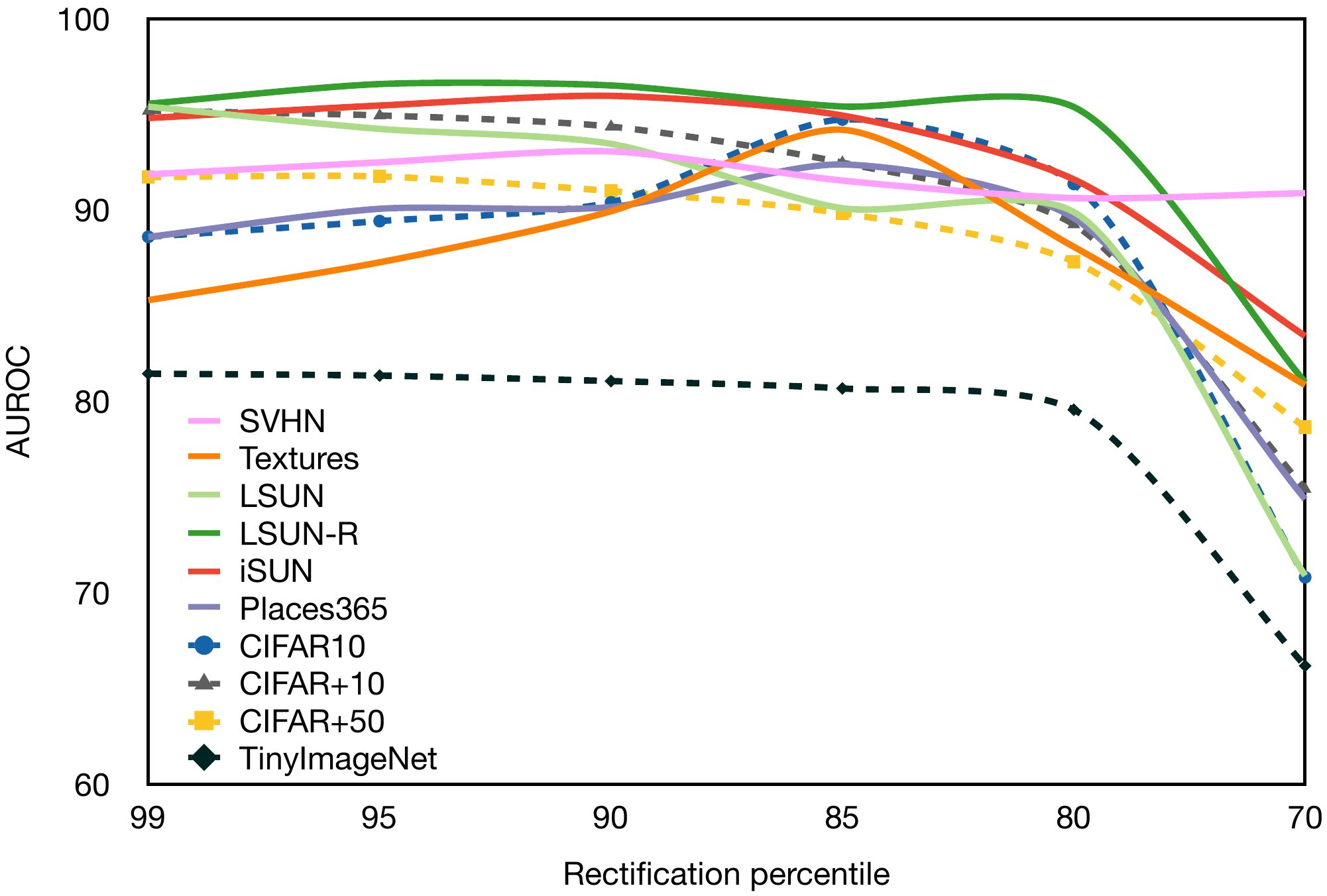}
    \caption{\reb{Investigation on the optimal hyper-parameters of ReAct for different datasets across OOD and OSR. ReAct is sensitive to the choice of the activation pruning percentile, and the optimal value of the percentile differs for each dataset.}}
    \label{fig:reb_react}
\end{figure*}

\reb{To ensure fairness and assess stability in our cross-evaluation, we report the averaged results from five independent trials, carefully selected around the optimal values. We identify the best hyper-parameters of ReAct which truncates activations above threshold to limit the effect of noise. Therefore, we investigate the choice of the percentile of activations for truncation. Our experiments involve varying the percentile of pruning activations. As shown in~\Cref{fig:reb_react}, ReAct's performance exhibits high sensitivity to the choice of the activation pruning percentile, and the optimal value of the percentile differs for each dataset.}

\clearpage

\section{Correlation of different metrics}
\label{supp:aupr}
To demonstrate that our takeaways (in Section 3.2 in the main paper) are metric-agnostic, we further evaluate using the following metrics: (1) the area under the precision-recall curve (AUPR); (2) OSCR \cite{dhamija2018reducing} which summarises the trade-off between closed-set accuracy and out-distribution / open-set performance as the threshold on the open-set score is varied. Results are shown in \Cref{tab:oe_metrics}. The same takeaways key findings remain consistent when utilizing AUROC (see also Table 2 in the main paper).
\begin{table*}[!ht]
\caption{Results of various methods on OOD detection benchmarks under different metrics (AUPR/OSCR), using CIFAR10 as ID. We train ResNet-18 with the OE loss. The results are averaged from five independent runs.}\label{tab:oe_metrics}
\centering
\resizebox{0.95\linewidth}{!}{
\begin{tabular}{l|cccccccccccccccccc}
\toprule
\multirow{3}{*}{Method} & \multicolumn{18}{c}{OOD detection benchmarks}                                                                                                                                                                                                                                                                                                                                                                                                                                                                                                                                                                                                                                                  \\ \cmidrule{2-19} 
                        & \multicolumn{3}{c}{SVHN}                                                                                        & \multicolumn{3}{c}{Textures}                                                                                    & \multicolumn{3}{c}{LSUN}                                                                                        & \multicolumn{3}{c}{LSUN-R}                                                                                      & \multicolumn{3}{c}{iSUN}                                                                                        & \multicolumn{3}{c}{Places365}                                                              \\\cmidrule{3-3}\cmidrule{6-6}
                        \cmidrule{9-9}
                        \cmidrule{12-12}
                        \cmidrule{15-15}
                        \cmidrule{18-18}
                        & AUROC                               & AUPR                                & OSRC                                & AUROC                               & AUPR                                & OSRC                                & AUROC                               & AUPR                                & OSRC                                & AUROC                               & AUPR                                & OSRC                                & AUROC                               & AUPR                                & OSRC                                & AUROC                               & AUPR                                & OSRC           \\ \hline
OE+MLS                  & \multicolumn{1}{c}{\textbf{99.21}} & \multicolumn{1}{c}{\textbf{99.54}} & \multicolumn{1}{c}{\textbf{94.69}} & \multicolumn{1}{c}{\textbf{98.82}} & \multicolumn{1}{c}{\textbf{98.81}} & \multicolumn{1}{c}{\textbf{95.16}} & \multicolumn{1}{c}{\textbf{99.02}} & \multicolumn{1}{c}{\textbf{98.13}} & \multicolumn{1}{c}{\textbf{94.01}} & \multicolumn{1}{c}{\textbf{98.53}} & \multicolumn{1}{c}{91.91}          & \multicolumn{1}{c}{\textbf{90.88}} & \multicolumn{1}{c}{\textbf{98.57}} & \multicolumn{1}{c}{91.94}          & \multicolumn{1}{c}{\textbf{90.94}} & \multicolumn{1}{c}{\textbf{97.32}} & \multicolumn{1}{c}{\textbf{99.42}} & \textbf{95.25} \\
OE+ODIN                 & \multicolumn{1}{c}{\textbf{99.43}} & \multicolumn{1}{c}{\textbf{99.42}} & \multicolumn{1}{c}{\textbf{94.64}} & \multicolumn{1}{c}{\textbf{98.73}} & \multicolumn{1}{c}{94.05}          & \multicolumn{1}{c}{93.21}          & \multicolumn{1}{c}{\textbf{99.14}} & \multicolumn{1}{c}{97.12}          & \multicolumn{1}{c}{92.89}          & \multicolumn{1}{c}{\textbf{98.78}} & \multicolumn{1}{c}{78.48}          & \multicolumn{1}{c}{83.85}          & \multicolumn{1}{c}{\textbf{98.75}} & \multicolumn{1}{c}{79.07}          & \multicolumn{1}{c}{84.42}          & \multicolumn{1}{c}{96.41}          & \multicolumn{1}{c}{95.98}          & 93.01          \\
OE+Energy               & \multicolumn{1}{c}{\textbf{99.20}} & \multicolumn{1}{c}{\textbf{99.52}} & \multicolumn{1}{c}{\textbf{94.65}} & \multicolumn{1}{c}{\textbf{98.78}} & \multicolumn{1}{c}{\textbf{98.76}} & \multicolumn{1}{c}{\textbf{95.14}} & \multicolumn{1}{c}{\textbf{99.02}} & \multicolumn{1}{c}{\textbf{98.01}} & \multicolumn{1}{c}{\textbf{93.94}} & \multicolumn{1}{c}{\textbf{98.55}} & \multicolumn{1}{c}{91.38}          & \multicolumn{1}{c}{90.58}          & \multicolumn{1}{c}{\textbf{98.58}} & \multicolumn{1}{c}{91.52}          & \multicolumn{1}{c}{\textbf{90.70}} & \multicolumn{1}{c}{\textbf{97.31}} & \multicolumn{1}{c}{\textbf{99.44}} & \textbf{95.24} \\
OE+MLS+ReAct            & \multicolumn{1}{c}{95.18}          & \multicolumn{1}{c}{\textbf{99.50}} & \multicolumn{1}{c}{94.45}          & \multicolumn{1}{c}{92.22}          & \multicolumn{1}{c}{98.80}          & \multicolumn{1}{c}{95.01}          & \multicolumn{1}{c}{79.46}          & \multicolumn{1}{c}{97.90}          & \multicolumn{1}{c}{93.49}          & \multicolumn{1}{c}{83.34}          & \multicolumn{1}{c}{\textbf{92.22}} & \multicolumn{1}{c}{\textbf{91.07}} & \multicolumn{1}{c}{83.68}          & \multicolumn{1}{c}{\textbf{92.26}} & \multicolumn{1}{c}{91.13}          & \multicolumn{1}{c}{87.46}          & \multicolumn{1}{c}{\textbf{99.41}} & 95.10          \\
OE+ODIN+ReAct           & \multicolumn{1}{c}{84.16}          & \multicolumn{1}{c}{99.37}          & \multicolumn{1}{c}{94.37}          & \multicolumn{1}{c}{82.92}          & \multicolumn{1}{c}{93.11}          & \multicolumn{1}{c}{92.34}          & \multicolumn{1}{c}{64.00}          & \multicolumn{1}{c}{96.13}          & \multicolumn{1}{c}{91.17}          & \multicolumn{1}{c}{73.90}          & \multicolumn{1}{c}{73.62}          & \multicolumn{1}{c}{79.22}          & \multicolumn{1}{c}{75.45}          & \multicolumn{1}{c}{74.52}          & \multicolumn{1}{c}{80.24}          & \multicolumn{1}{c}{71.65}          & \multicolumn{1}{c}{95.31}          & 92.09          \\
OE+Energy+ReAct         & \multicolumn{1}{c}{94.41}          & \multicolumn{1}{c}{\textbf{99.49}} & \multicolumn{1}{c}{94.41}          & \multicolumn{1}{c}{91.36}          & \multicolumn{1}{c}{\textbf{98.76}} & \multicolumn{1}{c}{94.98}          & \multicolumn{1}{c}{73.88}          & \multicolumn{1}{c}{97.79}          & \multicolumn{1}{c}{93.41}          & \multicolumn{1}{c}{80.03}          & \multicolumn{1}{c}{91.77}          & \multicolumn{1}{c}{90.80}          & \multicolumn{1}{c}{81.16}          & \multicolumn{1}{c}{91.94}          & \multicolumn{1}{c}{90.92}          & \multicolumn{1}{c}{86.19}          & \multicolumn{1}{c}{\textbf{99.43}} & 95.09          \\ \bottomrule 
\end{tabular}
}
\end{table*}

\clearpage

\section{Investigation on applying OE to the strong CLIP model}
\label{apd:vl_yfcc}

Pre-trained vision language models like CLIP~\cite{Radford2021Learning} have recently shown remarkable \textit{robustness} to distribution shifts.
The success of these models suggests that pre-training on diverse and extensive datasets is a promising approach for enhancing robustness. By utilizing pre-trained visual-language models such as CLIP, it is possible to extend OOD detection to various challenging ID datasets. 
\begin{figure}[h]
\centering
\includegraphics[width=\linewidth]{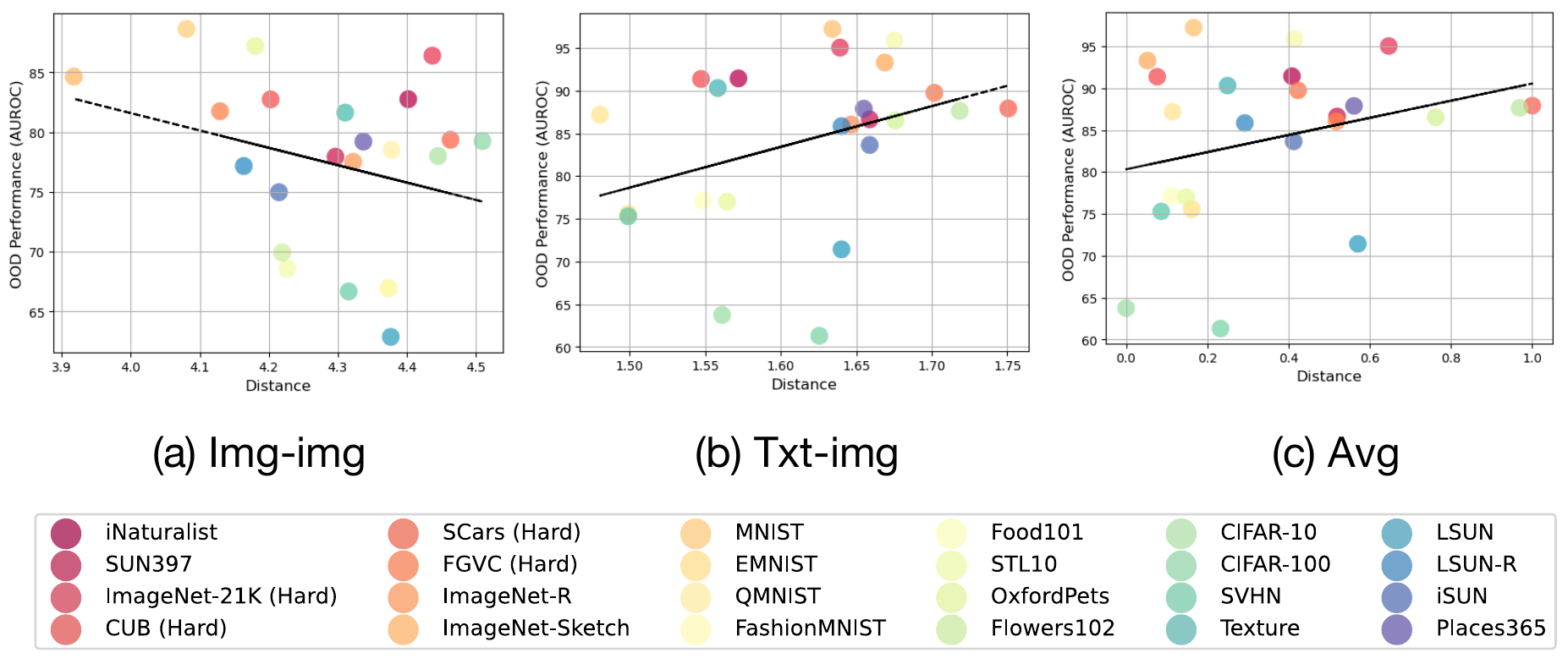}
\caption{The average performance \textit{vs.} distance to YFCC of all datasets when used as leave-one-out ID datasets in turn for the CLIP model. We compute cosine similarity with both image/text embeddings from ID data and image embeddings from auxiliary data.
No clear correlation between the distance to YFCC and OOD detection performance is observed. 
}\label{fig:clip_dist}
\end{figure}

\begin{figure*}[h]
\centering
\includegraphics[width=\linewidth]{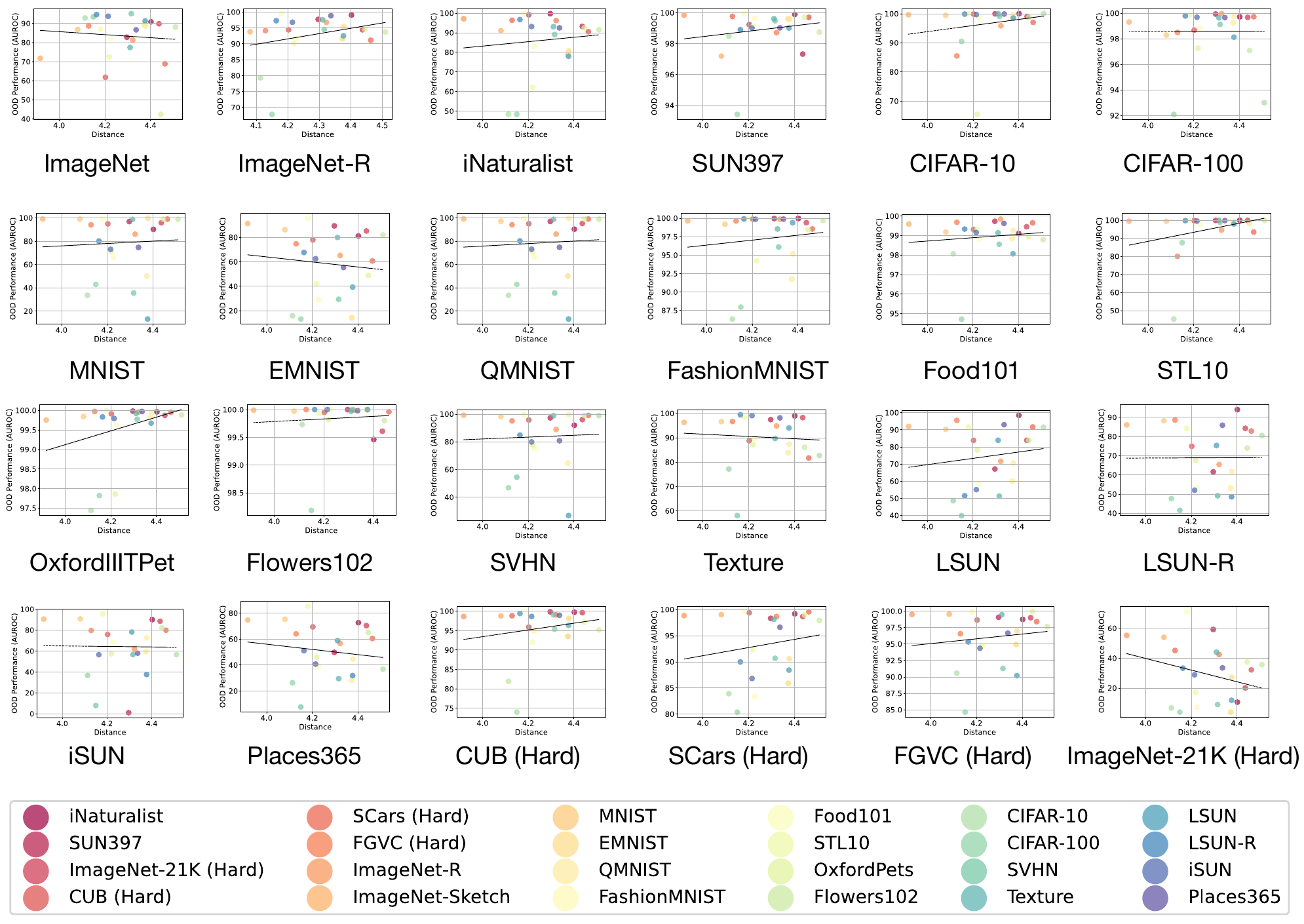}
\caption{OOD detection performance \textit{vs.} OOD-AUX distance to YFCC, when taking each dataset as ID dataset and others as OOD datasets for CLIP model. We calculate the distance between image features of all OOD samples and those of YFCC-15M. }\label{fig:clip_img}
\end{figure*}

\begin{figure*}[h]
\centering
\includegraphics[width=\linewidth]{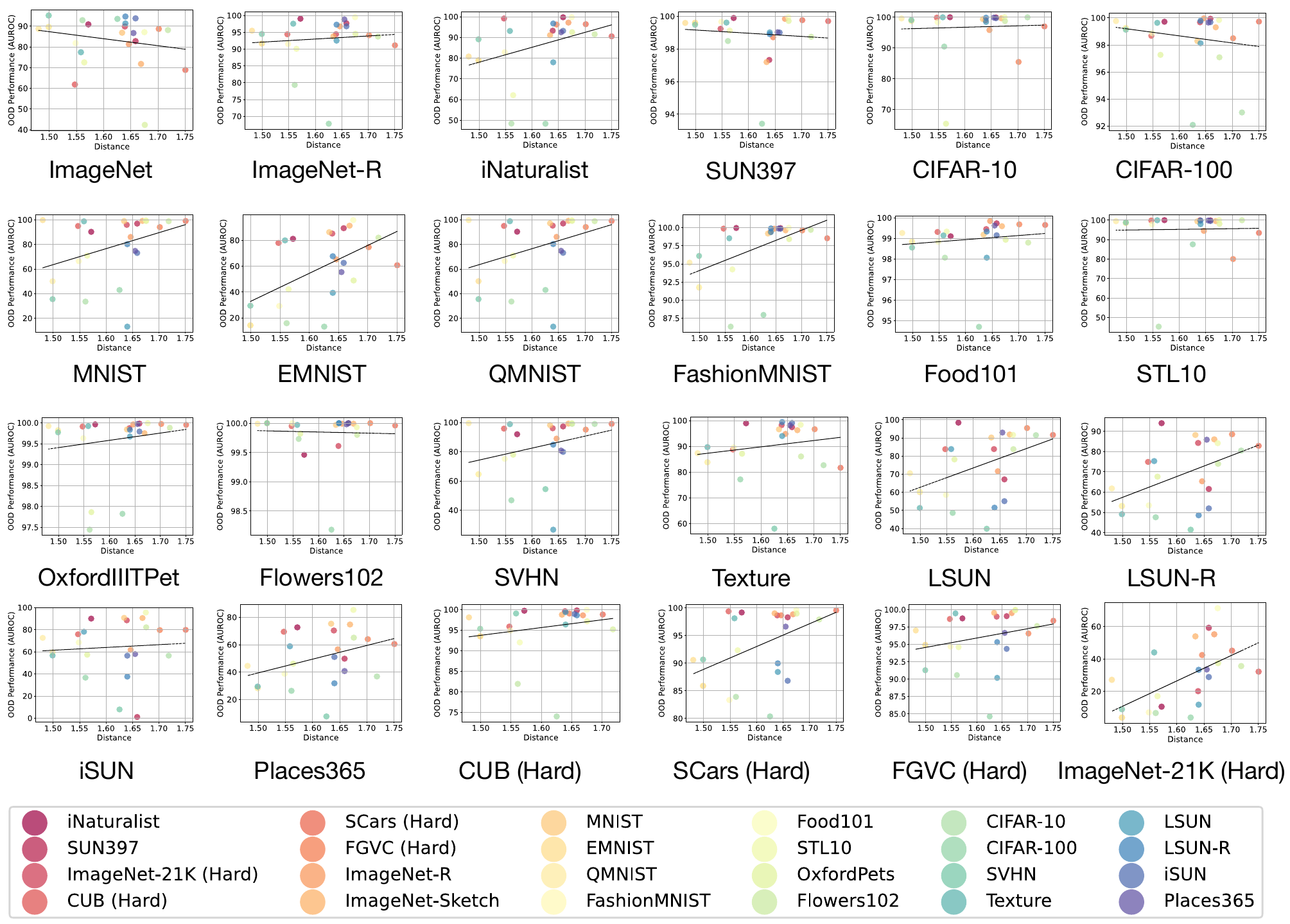}
\caption{OOD detection performance \textit{vs.} OOD-AUX distance to YFCC, when taking each dataset as ID dataset and others as OOD datasets for CLIP model. We calculate the distance between image features of all OOD samples and prompt features of YFCC-15M.}\label{fig:clip_txt}
\end{figure*}

We finetune CLIP using OE with auxiliary data and apply MLS to the similarity scores to separate ID and OOD samples.
For auxiliary data, we use the subset of YFCC-100M~\cite{cheng2021data}, known as YFCC-15M. It consists of the 15 million images which were used in CLIP pretraining. 
Specifically, we consider the following ID datasets: iNaturalist, SUN397, Places, hard splits of SSB according to~\cite{vaze2022openset} (ImageNet-21K, CUB, SCars and FGVC), variants of ImageNet (ImageNet-R and ImageNet-Sketch), variants of MNIST (MNIST, EMNIST, QMNIST and FashionMNIST), Food101, STL10, OxfordPets, Flowers102, CIFAR-10, CIFAR-100, SVHN, Texture, LSUN, LSUN-R, iSUN, Places365. As for the OOD test datasets, we treat one dataset as the hold-out ID dataset in turn to evaluate the performance of others OOD datasets that do not overlap with the ID dataset. \Cref{fig:clip_dist} examines different query types of inputs (\textit{e.g.}, images, prompts). We use a set of pre-defined prompts for each class, which are collected from prior works~\cite{Radford2021Learning,cherti2023reproducible}. We compute cosine similarity with the $L_2$-normalized embeddings of images or the embedding of prompts of each class by averaging over the prompt pool. 
We present the OOD detection performance \textit{vs.} OOD-AUX distance in~\Cref{fig:clip_img} and~\Cref{fig:clip_txt} by comparing different features.
Interestingly, there is no obvious trend shared among all the cases.
This is likely due to the fact that CLIP is pretrained on YFCC-15M, while in OOD detection, YFCC-15M is treated as auxiliary data to mimic OOD data to be pushed away from the ID data, hurting the original pretrained strong embedding space of CLIP.

\clearpage

\bibliographystyle{abbrvnat}
\bibliography{sn-bib}